\documentclass[sigconf]{acmart}
\usepackage{subcaption}
\usepackage{multirow}
\usepackage{pifont}
\usepackage{enumitem}
\usepackage{wrapfig}
\usepackage{graphicx}
\usepackage{natbib}
\usepackage{subcaption}
\theoremstyle{plain}
\newtheorem{theorem}{Theorem}[section]
\newtheorem{proposition}[theorem]{Proposition}

\newtheorem{corollary}[theorem]{Corollary}
\theoremstyle{definition}
\newtheorem{definition}[theorem]{Definition}

\theoremstyle{remark}
\newtheorem{remark}[theorem]{Remark}

\AtBeginDocument{%
  }

\setcopyright{acmlicensed}
\copyrightyear{2018}
\acmYear{2018}
\acmDOI{XXXXXXX.XXXXXXX}
\acmConference[Conference acronym 'XX]{Make sure to enter the correct
  conference title from your rights confirmation email}{June 03--05,
  2018}{Woodstock, NY}
\acmISBN{978-1-4503-XXXX-X/2018/06}

\begin{document}

\title{Unveiling Mode Connectivity in Graph Neural Networks}

\author{Bingheng Li}
\affiliation{%
    \institution{Department of Computer Science and Engineering,}
  \institution{Michigan State University}
  \country{}
}
\email{libinghe@msu.edu}

\author{Zhikai Chen}
\affiliation{%
    \institution{Department of Computer Science and Engineering,}
  \institution{Michigan State University}
  \country{}
}
\email{chenzh85@msu.edu}

\author{Haoyu Han}
\affiliation{%
    \institution{Department of Computer Science and Engineering,}
  \institution{Michigan State University}
  \country{}
}
\email{hanhaoy1@msu.edu}

\author{Shenglai Zeng}
\affiliation{%
    \institution{Department of Computer Science and Engineering,}
  \institution{Michigan State University}
  \country{}
}
\email{zengshe1@msu.edu}

\author{Jingzhe Liu}
\affiliation{%
    \institution{Department of Computer Science and Engineering,}
  \institution{Michigan State University}
  \country{}
}
\email{liujin33@msu.edu}

\author{Jiliang Tang}
\affiliation{
\institution{Department of Computer Science and Engineering,}
  \institution{Michigan State University}
  \country{}
}
\email{tangjili@msu.edu}

\renewcommand{\shortauthors}{Trovato et al.}
\newcommand\liu[1]{\textcolor{blue}{liu: #1}}
\begin{abstract}
A fundamental challenge in understanding graph neural networks (GNNs) lies in characterizing their optimization dynamics and loss landscape geometry, critical for improving interpretability and robustness. While mode connectivity—a lens for analyzing geometric properties of loss landscapes—has proven insightful for other deep learning architectures, its implications for GNNs remain unexplored. This work presents the first investigation of mode connectivity in GNNs. We uncover that GNNs exhibit distinct non-linear mode connectivity, diverging from patterns observed in fully-connected networks or CNNs. Crucially, we demonstrate that graph structure, rather than model architecture, dominates this behavior, with graph properties like homophily correlating with mode connectivity patterns. We further establish a link between mode connectivity and generalization, proposing a generalization bound based on loss barriers and revealing its utility as a diagnostic tool. Our findings further bridge theoretical insights with practical implications: they rationalize domain alignment strategies in graph learning and provide a foundation for refining GNN training paradigms. 
\end{abstract}

\ccsdesc[500]{Theory of computation~Graph algorithms analysis}
\ccsdesc[500]{Computing methodologies~Neural networks}



\maketitle

\section{Introduction}



Graph Neural Networks (GNNs) have emerged as a dominant paradigm for processing graph-structured data, achieving state-of-the-art results across diverse applications, from social network analysis to bioinformatics and traffic prediction~\citep{ma2021deep,wang2020traffic}. Despite their empirical success, the fundamental understanding of GNN optimization dynamics and the intricate geometry of their loss landscapes remains limited, hindering progress in model interpretability and robustness~\cite{jin2020graph}. Meanwhile, the concept of mode connectivity~\citep{garipov2018loss, draxler2018essentially}, which refers to the relationships between disparate local minima obtained from varied training runs, offers a powerful lens for examining the loss landscape. Mode connectivity has been shown closely related to model robustness~\citep{zhao2020bridging} and generalization capabilities~\citep{vrabel2024input}. Standing apart from other theoretical tools like Neural Tangent Kernels~\citep{yanggraph} or Mean Field Theory~\citep{aminian2024generalization} that often rely on assumptions not readily applicable to real-world scenarios, mode connectivity allows for the study of practical models under realistic conditions.

While mode connectivity has been extensively explored in fully-connected networks (FCNs), convolutional neural networks (CNNs), and transformers \citep{garipov2018loss, draxler2018essentially, qin2022exploring, entezari2021role}, a systematic investigation into GNN mode connectivity is, to the best of our knowledge, absent.  GNNs, operating on non-Euclidean graph data, introduce unique data-structure interactions that may fundamentally alter training dynamics compared to models processing independent and identically distributed (i.i.d.) data. Understanding whether GNN mode connectivity deviates from that of other architectures is crucial. Dissimilar behavior could unveil unique properties of the GNN loss landscape, while similarities would allow us to leverage existing theoretical frameworks from other domains to explain the GNN optimization.  Therefore, exploring GNN mode connectivity is a critical step towards demystifying these powerful models, motivating our central inquiry: \textit{How does mode connectivity manifest in GNNs, and how is it influenced by the inherent structure of the input graph?}

This work provides an initial, systematic answer to these questions through a comprehensive controlled study encompassing 12 diverse graphs from various domains.  Our key observation is that, in contrast to FCNs and CNNs, \textbf{GNNs exhibit a distinct non-linear mode connectivity, accurately characterized by a polynomial curve such as a quadratic Bézier curve}. This phenomenon strongly suggests that the non-i.i.d. nature of graph data profoundly impacts GNN training dynamics.  Furthermore, dissecting the influence of model architecture versus data characteristics, we find that \textbf{GNN architectures have minimal impact on mode connectivity, whereas graph structure emerges as the dominant factor}. This relationship is further corroborated by a strong correlation between specific graph properties and the observed mode connectivity. We complement these empirical findings with a theoretical analysis providing rigorous justification for our observations.

Beyond characterizing GNN mode connectivity, our empirical evidence reveals a compelling link between mode connectivity and generalization performance. For instance, models trained on graphs exhibiting enhanced feature separability and higher homophily levels demonstrate simultaneously superior predictive accuracy and more robust mode connectivity.  This motivates the use of mode connectivity as a diagnostic tool for GNN generalization.  We introduce a generalization bound based on loss barriers, a key metric derived from mode connectivity analysis. Extending our investigation across diverse graph domains, we further construct a Wasserstein distance based on mode connectivity, offering a novel geometric perspective for domain alignment and providing a potential explanation for the efficacy of existing model-based knowledge transfer. In summary, we present a systematic study on the mode connectivity of GNNs and uncover novel insights into their optimization behavior.

\section{Preliminaries}
\label{sec: pre}

In this section, we introduce the graph neural network (GNN) model used for node classification (Section~\ref{sec: gnn}) and the Contextual Stochastic Block Model (CSBM) as the underlying data model for theoretical analysis (Section~\ref{sec: csbm}). We then formally define mode connectivity in GNNs (Section~\ref{sec: mc}) and discuss different experimental evaluation methods used to study it.

\subsection{Graph Neural Networks}
\label{sec: gnn}

In this work, we consider a graph neural network (GNN) trained for node classification task on a graph represented as $\mathcal{G} = (\mathbf{A}, \mathbf{X}, Y)$. The adjacency matrix $\mathbf{A} \in \mathbb{R}^{n \times n}$ encodes the graph structure, where $n$ is the number of nodes. The node feature matrix $\mathbf{X} \in \mathbb{R}^{n \times d}$ contains $d$-dimensional features for each node. The label matrix $Y \in \mathbb{R}^{n \times C}$ represents the ground-truth labels for $C$ classes.

\noindent\textbf{GNN Forward Propagation:}  
In this work, we focus on Graph Convolutional Networks (GCNs)\cite{kipf2016semi}, the classic GNN architecture, which propagates node representations iteratively as:
A GNN with $L$ layers propagates node representations iteratively as:
\begin{equation}
    \mathbf{H}^{(l)} = \sigma\Bigl(\hat{\mathbf{A}} \mathbf{H}^{(l-1)} \mathbf{W}^{(l)}\Bigr), \quad l = 1, \dots, L,
\end{equation}
where $\mathbf{H}^{(l)} \in \mathbb{R}^{n \times d_l}$ denotes node embeddings at layer $l$, with $d_l$ the embedding dimension of that layer. $\hat{\mathbf{A}}$ is the normalized adjacency matrix, $\mathbf{W}^{(l)}$ is the trainable weight matrix, and $\sigma(\cdot)$ is a non-linear activation function. The initial representation is $\mathbf{H}^{(0)} = \mathbf{X}$.

\noindent\textbf{Final Node Embeddings:}  
After $L$ layers of message passing, the final node representations are: $\mathbf{Z} = \mathbf{H}^{(L)} \mathbf{W}^{(L)}$,
where $\mathbf{Z} \in \mathbb{R}^{n \times C}$ represents the node embeddings before classification.

\noindent\textbf{Output Layer (Softmax Probability Distribution):}  
The predicted class probabilities are obtained using the softmax function:$\mathcal{P}^{\mathbf{Z}} = \text{softmax}(\mathbf{Z})$, where $\mathcal{P}^{\mathbf{Z}} \in \mathbb{R}^{n \times C}$ is the probability distribution over $C$ classes.


\noindent\textbf{Loss Function:}  
The model is trained by minimizing the cross-entropy loss over the labeled nodes in the training set $\mathcal{V}_{\text{train}}$:  
\begin{equation}
    \mathcal{L}_{\text{Tr}} = - \sum_{i \in \mathcal{V}_{\text{train}}} \sum_{c=1}^{C} Y_{i,c} \log \mathcal{P}_{i,c}^{\mathbf{Z}},
\end{equation}
where $Y_{i,c}$ denotes the one-hot ground-truth label. For evaluation purposes, we also define the test loss on $\mathcal{V}_{\text{test}}$ as:  
\begin{equation}
    \mathcal{L}_{\text{Te}} = - \sum_{i \in \mathcal{V}_{\text{test}}} \sum_{c=1}^{C} Y_{i,c} \log \mathcal{P}_{i,c}^{\mathbf{Z}}.
\end{equation}
\noindent\textbf{Accuracy Metric:}  
Classification accuracy is measured as the proportion of correctly classified nodes: 

\begin{equation}
    \text{Acc}_{\text{Tr}} = \frac{1}{|\mathcal{V}_{\text{train}}|} \sum_{i \in \mathcal{V}_{\text{train}}} \mathbb{I}(\hat{y}_i = \arg\max_c Y_{i,c}),
\end{equation}

\begin{equation}
    \text{Acc}_{\text{Te}} = \frac{1}{|\mathcal{V}_{\text{test}}|} \sum_{i \in \mathcal{V}_{\text{test}}} \mathbb{I}(\hat{y}_i = \arg\max_c Y_{i,c}),
\end{equation}
where $\hat{y}_i = \arg\max_c \mathcal{P}_{i,c}^{\mathbf{Z}}$ is the predicted label, and $\mathbb{I}(\cdot)$ is the indicator function.

\noindent\textbf{Generalization Gap~\cite{neyshabur2017exploring}:}  
 The generalization gap is defined as the difference between the loss on test set and that on training set:
\[
\Delta_{\text{gen}} = \mathcal{L}_{\text{Te}} - \mathcal{L}_{\text{Tr}}.
\]

\noindent\textbf{Domain Performance Gap~\cite{ben2010theory}:}  
Assume that the model is evaluated on two different domains, a source domain $\mathcal{G}_S $ and a target domain$\mathcal{G}_T$, with corresponding losses \(\mathcal{L}_S\) and \(\mathcal{L}_T\). The domain adaptation gap is defined as:
\[
\Delta_{\text{da}} = \mathcal{L}_T - \mathcal{L}_S.
\]
\subsection{Contextual Stochastic Block Model (CSBM)}
\label{sec: csbm}

To theoretically understand the behavior of GNNs, we employ Contextual Stochastic Block Model (CSBM)~\cite{deshpande2018contextual} as the underlying data model for graphs, which has been widely adopted for node-level task~\cite{Ma2021IsHA, baranwal2021graph, wang2022augmentation, baranwal2023effects, wang2022can, fountoulakis2022graph} due to its ability to simultaneously capture both node features and graph structures.

\begin{definition}[Contextual Stochastic Block Model (CSBM)]
A CSBM with two classes defines a graph \( \mathcal{G} = (\mathcal{V}, \mathcal{E}, X, Y) \), where  
\(|V|=n\) denotes the number of nodes nodes and \( \mathcal{V} \) belong to two disjoint classes \( \mathcal{C}_1 \) and \( \mathcal{C}_2 \), with labels \( Y \in \{c_1, c_2\} \). Each node \( v \) with label \( c_v \) has a feature vector \( x_v \sim \mathcal{N}(\mu_i, \sigma I) \), where \( \mu_i \in \mathbb{R}^d \) is the class-wise mean, and $\sigma$ quantifies the separability of node features. Edges form independently with probability  
\[
P((u,v) \in \mathcal{E}) =
\begin{cases}
    p_{in}, & \text{if } Y_u = Y_v, \\
    p_{out}, & \text{if } Y_u \neq Y_v,
\end{cases}
\]
where \( p_{in} > p_{out} > 0 \) denote intra-class and inter-class connection probabilities, respectively.
\end{definition}
\paragraph{Remark.}  
The CSBM model describes a graph where edges form based on class memberships, and node features are drawn from class-specific distributions. The probability sum \( p_{in} + p_{out} \) reflects the overall edge density, influencing how well information propagates in the network. The ratio \( \frac{p_{in}}{p_{in} + p_{out}} \) measures the tendency of nodes to connect within their class, affecting the homophily of the graph. The feature variance \( \sigma \) controls how tightly node features cluster around their class means, impacting the difficulty of classification. Additionally, the number of nodes \( n \) determines the scale of the graph, which can influence statistical properties and computational complexity.

    

\subsection{Mode Connectivity}
\label{sec: mc}

Mode connectivity~\cite{garipov2018loss,draxler2018essentially} studies the relationship between different parameter space modes in neural networks. Despite the abundant research about mode connectivity in FCN, CNN, and transformer~\citep{draxler2018essentially, garipov2018loss, entezari2021role, qin2022exploring}, there's yet no existing research on mode connectivity in GNN. So, we aim to extend existing definitions of mode connectivity to GNNs. 
\noindent\textbf{Parameter Space Mode.}  
A \textit{mode} in the parameter space refers to a set of parameters $\theta$ that minimizes the loss function $\mathcal{L}$ for a given graph $\mathcal{G}$. Formally, an optimal mode is defined as:
\begin{equation}
\theta^* = \operatorname*{argmin}_\theta \mathcal{L}(f(\mathcal{G};\theta), Y),
\end{equation}
where $f(\mathcal{G}; \theta) = \mathcal{P}^{\mathbf{Z}}$ represents the probability distribution predicted by the GNN model, and $Y$ is the ground-truth label matrix. For an \( L \)-layer GNN, the parameters are given by \( \theta = \{W^{l}\} \) for \( l = 0, \dots, L \).
Different modes arise due to variations in training conditions such as random initialization and data sampling.

Given two independently trained GNN models under different training configurations $M_a$ and $M_b$, we obtain two sets of parameters $\theta_a$ and $\theta_b$, each corresponding to a local minimum of $\mathcal{L}$. 
A fundamental property of mode connectivity is the ability to connect these minima with a low-loss path~\citep{draxler2018essentially, garipov2018loss}. Two commonly considered paths are linear interpolation and Bézier curve interpolation, both of which are used to fit the minima of the trained model. A good fit suggests that the modes lie along a simple, low-loss manifold, providing further insights into the structure of the loss landscape.

\noindent\textbf{Linear Interpolation Path.}  
\citet{frankle2020linear,qin2022exploring} demonstrate that neural networks initialized similarly often exhibit a linear low-loss connection between minima. A linear interpolation path can be used to examine this property, which can be defined as continuous curve $\phi(\alpha): [0, 1] \to \mathbb{R}^{|\theta|}$ connecting $\theta_a$ and $\theta_b$:
\begin{equation}
    \phi(\alpha) = (1 - \alpha) \theta_a + \alpha \theta_b, \quad \alpha \in [0,1].
\end{equation}

Even if a fully linear path between two minima results in a high loss, a low-loss non-linear path may still exist. A quadratic Bézier curve is often adopted~\cite{garipov2018loss,gotmare2018using,draxler2018essentially} to examine such property.

\noindent\textbf{Bezier Curve Interpolation Path.} A bezier curve with  two modes $\theta_{a}$ and $\theta_{b}$ as endpoints can be given as 
\begin{equation}
\phi(\alpha) = (1-\alpha)^2 \theta_{a} + 2\alpha(1-\alpha) \theta + \alpha^2 \theta_{b},
\end{equation}
where $\theta$ is a learnable parameter. Specifically, mode connectivity can be analyzed through  
(i) the smoothness of the loss and accuracy along $\phi(\alpha)$, and  
(ii) the \textit{loss barrier}, which captures the maximum increase in loss along the interpolation.

\begin{definition}[Loss Barrier]
For two given models $\theta_a$ and $\theta_b$, the loss barrier is defined as:
\begin{equation}
    B(\theta_a, \theta_b) := \max_{\alpha \in [0,1]} 
    \Big( \mathcal{L}(\phi(\alpha)) - \mathcal{L}_{\text{lin}}(\alpha) \Big),
\end{equation}
where $\mathcal{L}_{\text{lin}}(\alpha)$ is the linearly interpolated loss: $\mathcal{L}_{\text{lin}}(\alpha) = (1-\alpha) \mathcal{L}(\theta_a) + \alpha \mathcal{L}(\theta_b). $
\end{definition}

The loss barrier measures the deviation from a perfect linear interpolation in the loss landscape. Taking linear interpolation path as an example, a higher $B(\theta_a, \theta_b)$ suggests weaker connectivity, whereas $B(\theta_a, \theta_b) \approx 0$ indicates linear mode connectivity.  

\begin{figure*}[!ht]\label{fig:mc1}
    \centering
    \begin{subfigure}[b]{0.24\textwidth}
        \centering
        \includegraphics[width=\textwidth]{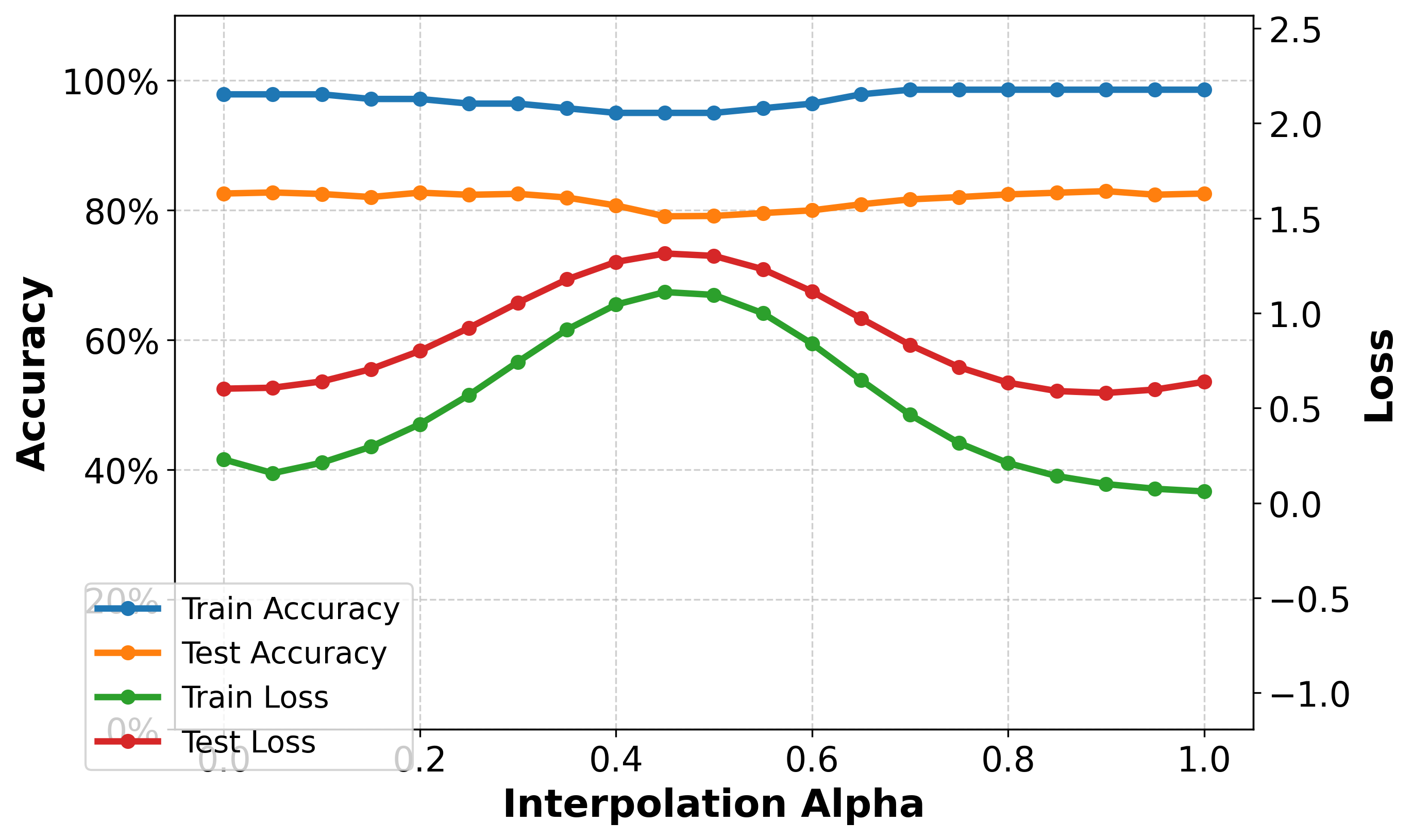}
        \caption{Cora}
    \end{subfigure}
    \begin{subfigure}[b]{0.24\textwidth}
        \centering
        \includegraphics[width=\textwidth]{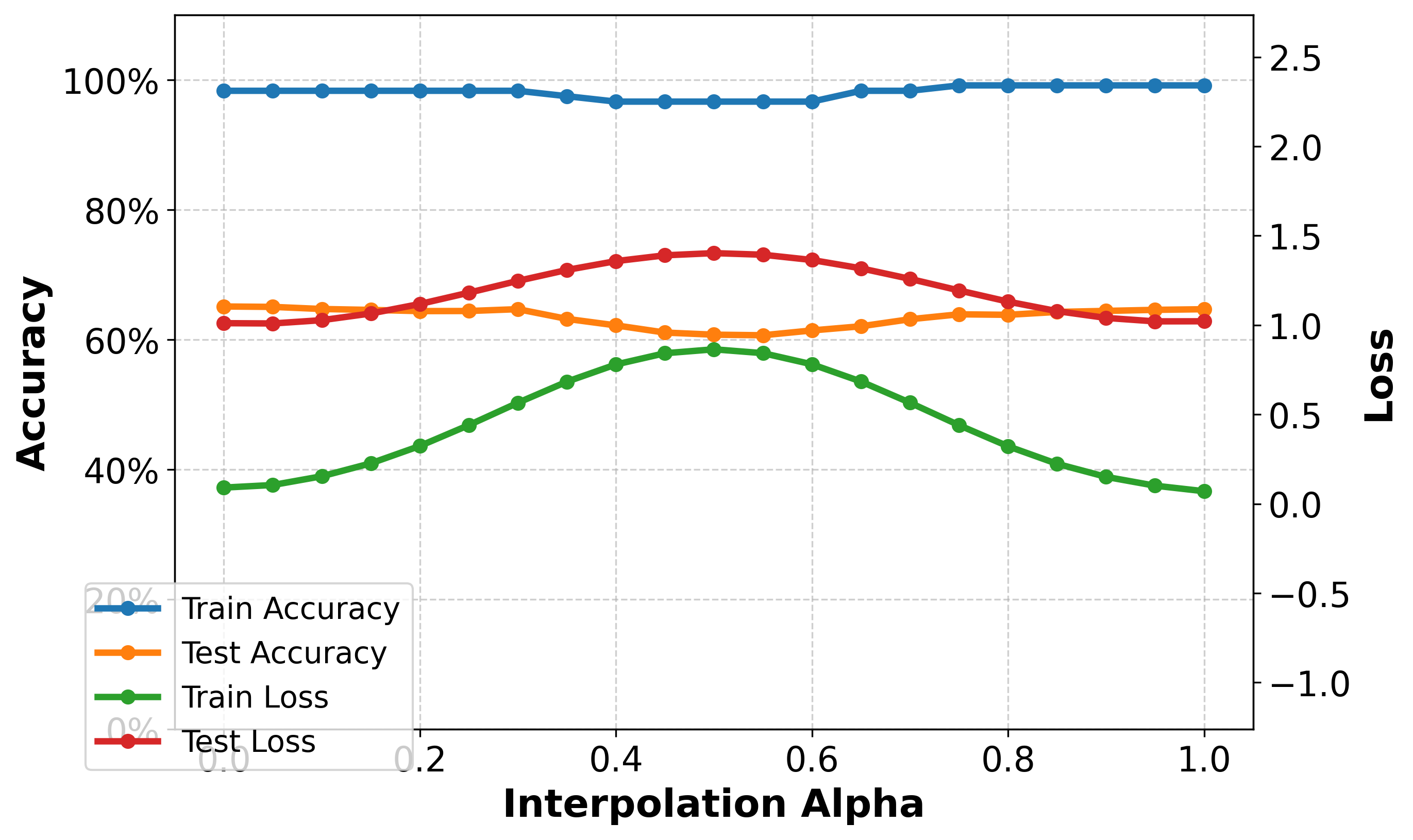}
        \caption{Citeseer}
    \end{subfigure}
    \begin{subfigure}[b]{0.24\textwidth}
        \centering
        \includegraphics[width=\textwidth]{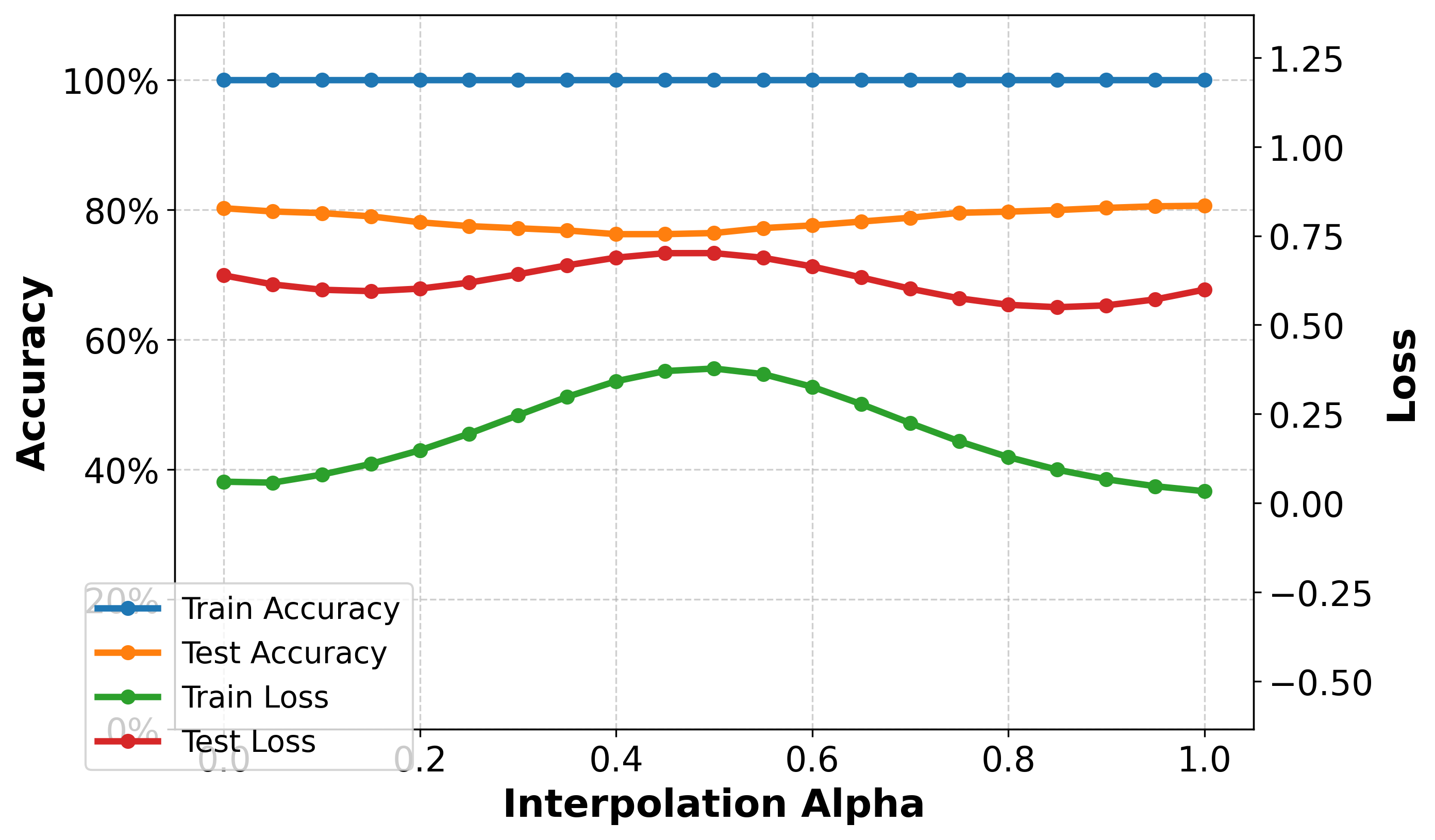}
        \caption{Pubmed}
    \end{subfigure}
    \begin{subfigure}[b]{0.24\textwidth}
        \centering
        \includegraphics[width=\textwidth]{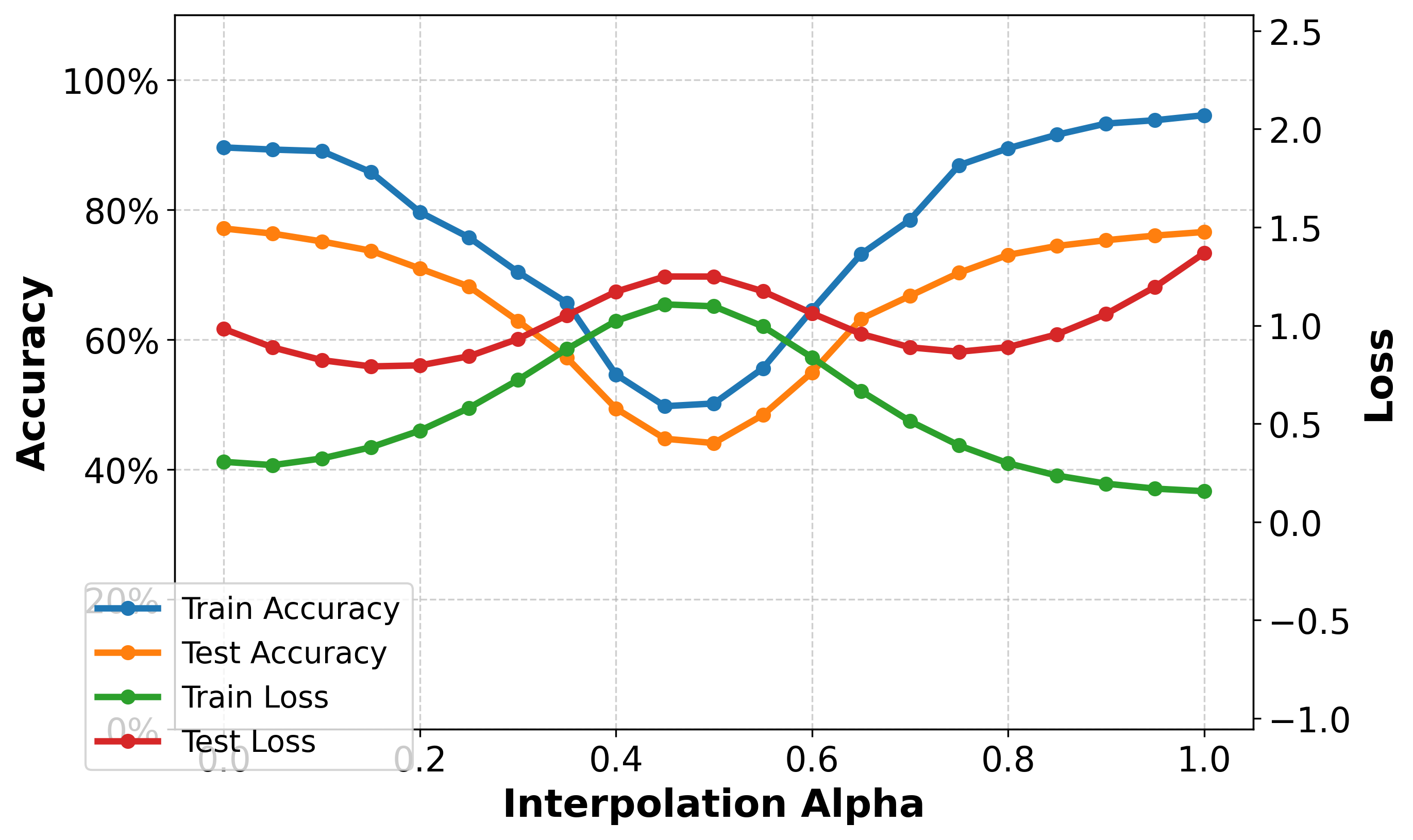}
        \caption{Wiki-CS}
    \end{subfigure}
    \vspace{5pt}
    \label{fig: figure1}
    \newline
    \begin{subfigure}[b]{0.24\textwidth}
        \centering
        \includegraphics[width=\textwidth]{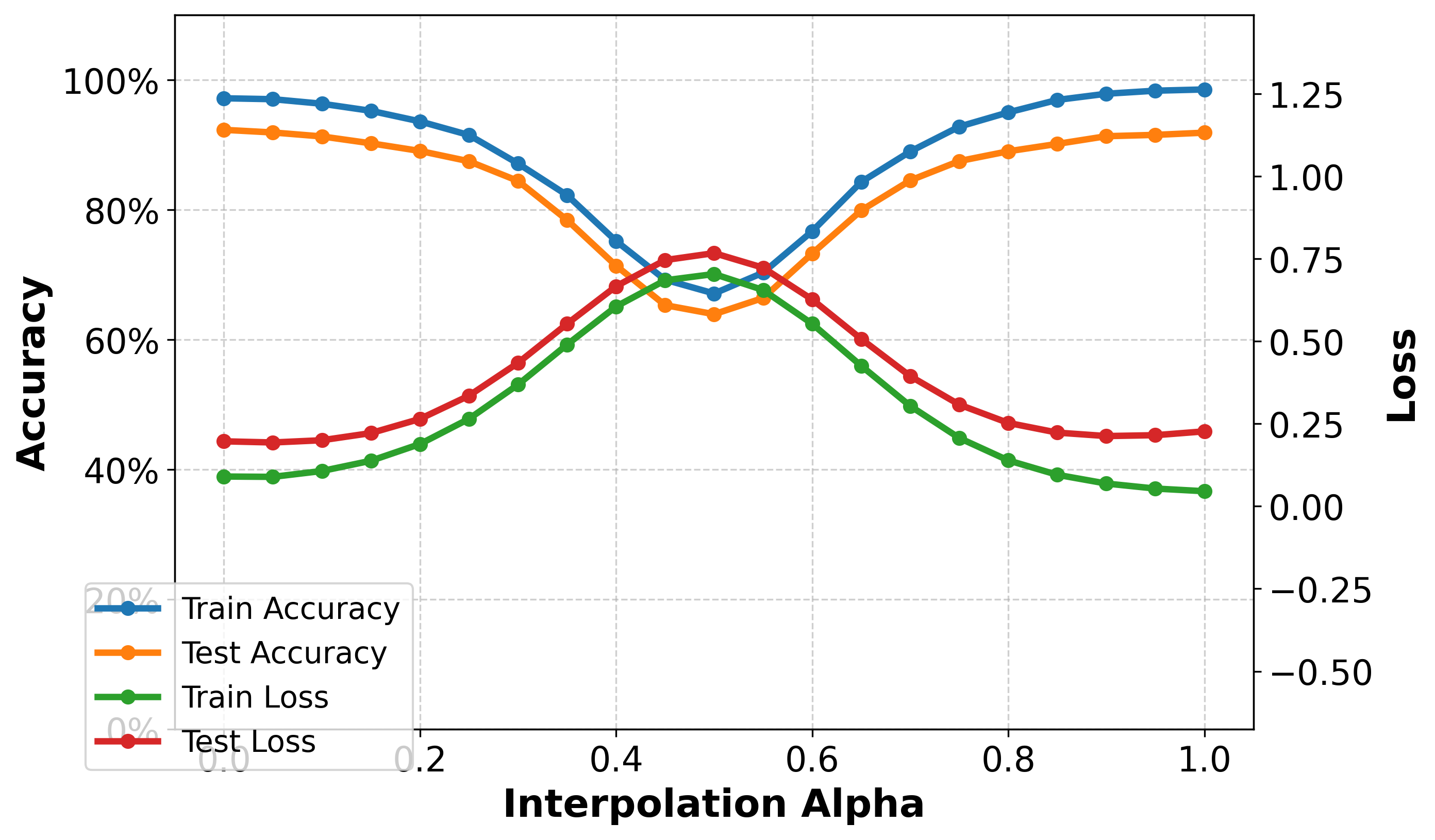}
        \caption{Coauthor-CS}
    \end{subfigure}
    \begin{subfigure}[b]{0.24\textwidth}
        \centering
        \includegraphics[width=\textwidth]{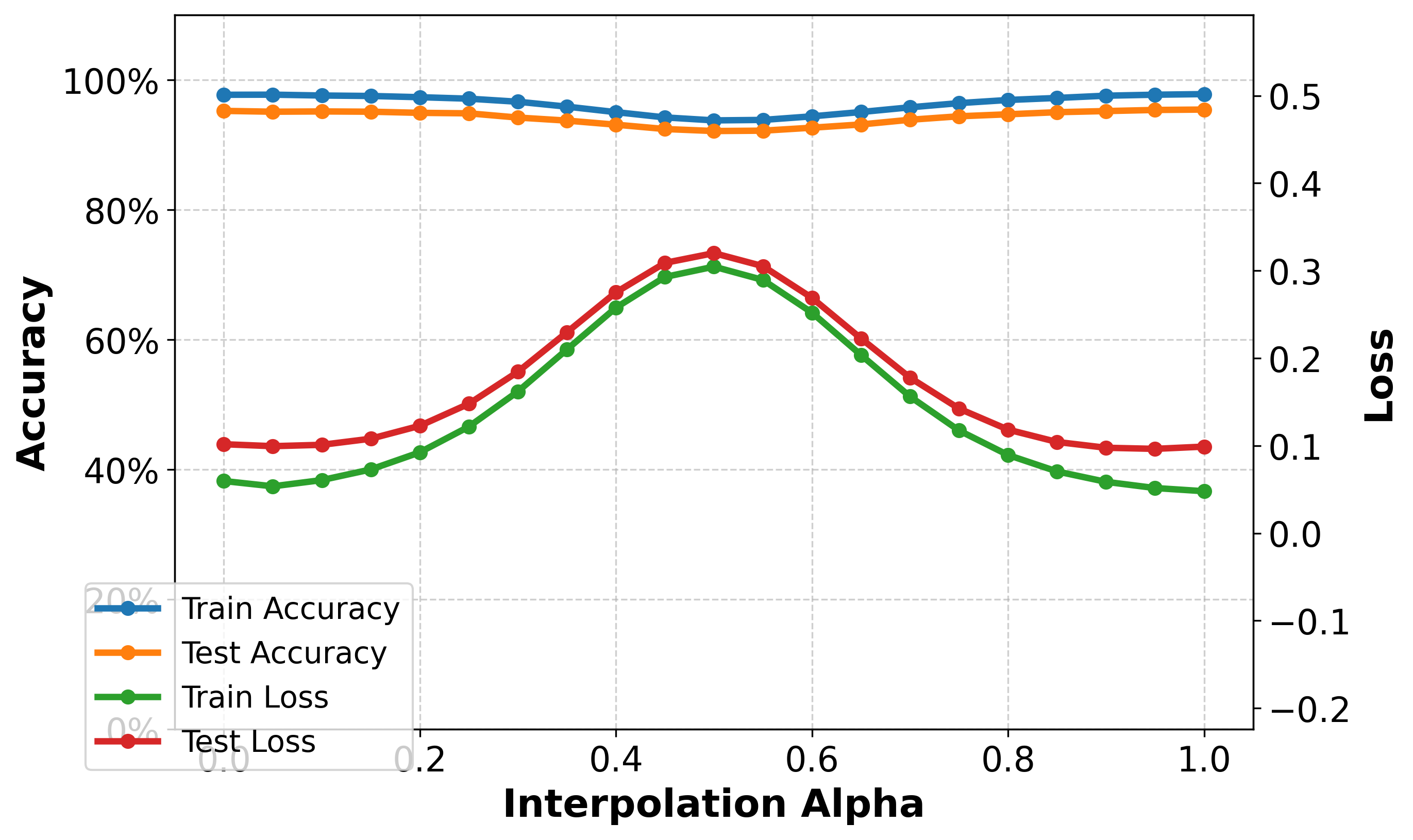}
        \caption{Coauthor-Physics}
    \end{subfigure}
    \begin{subfigure}[b]{0.24\textwidth}
        \centering
        \includegraphics[width=\textwidth]{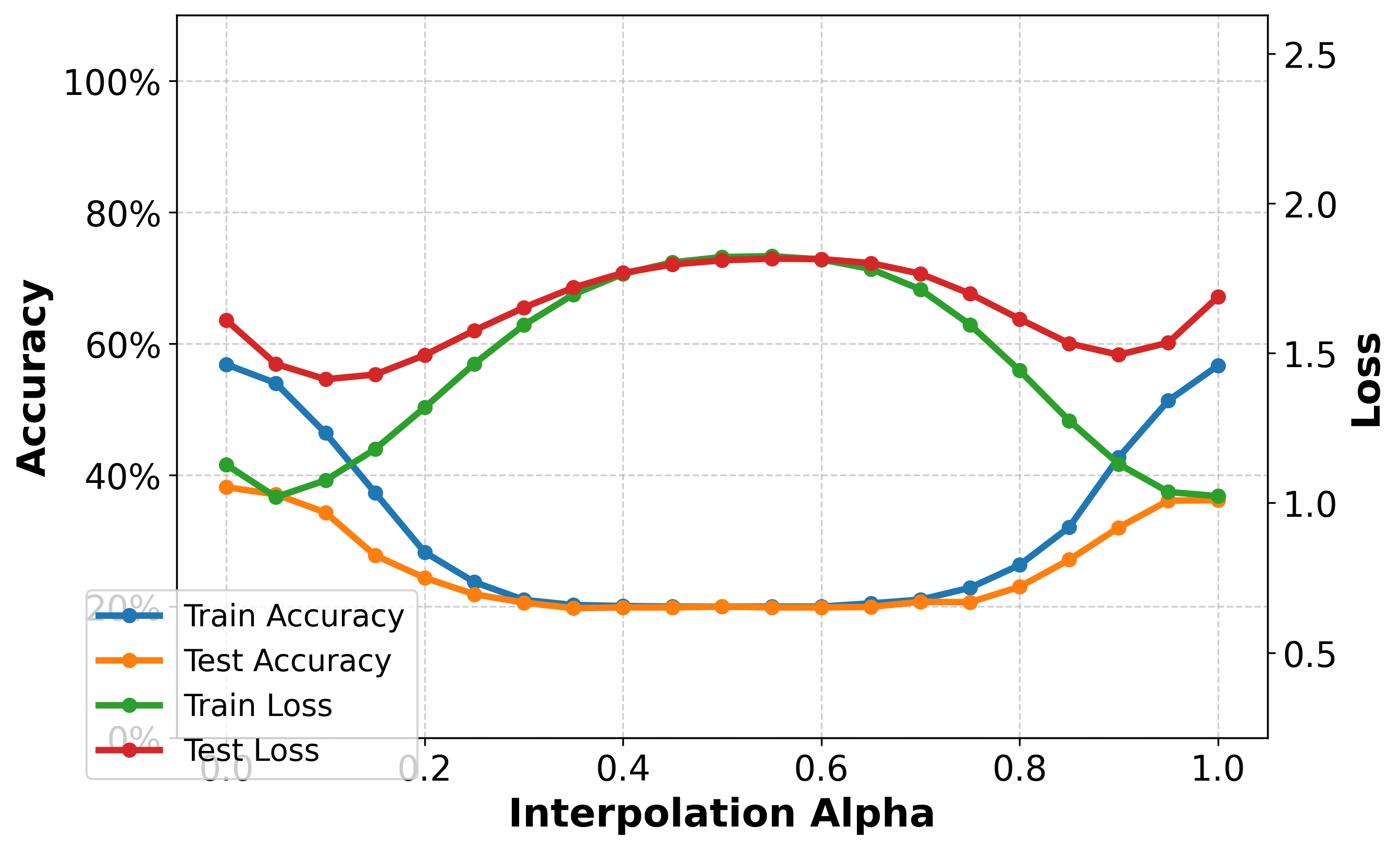}
        \caption{Squirrel}
    \end{subfigure}
    \begin{subfigure}[b]{0.24\textwidth}
        \centering
        \includegraphics[width=\textwidth]{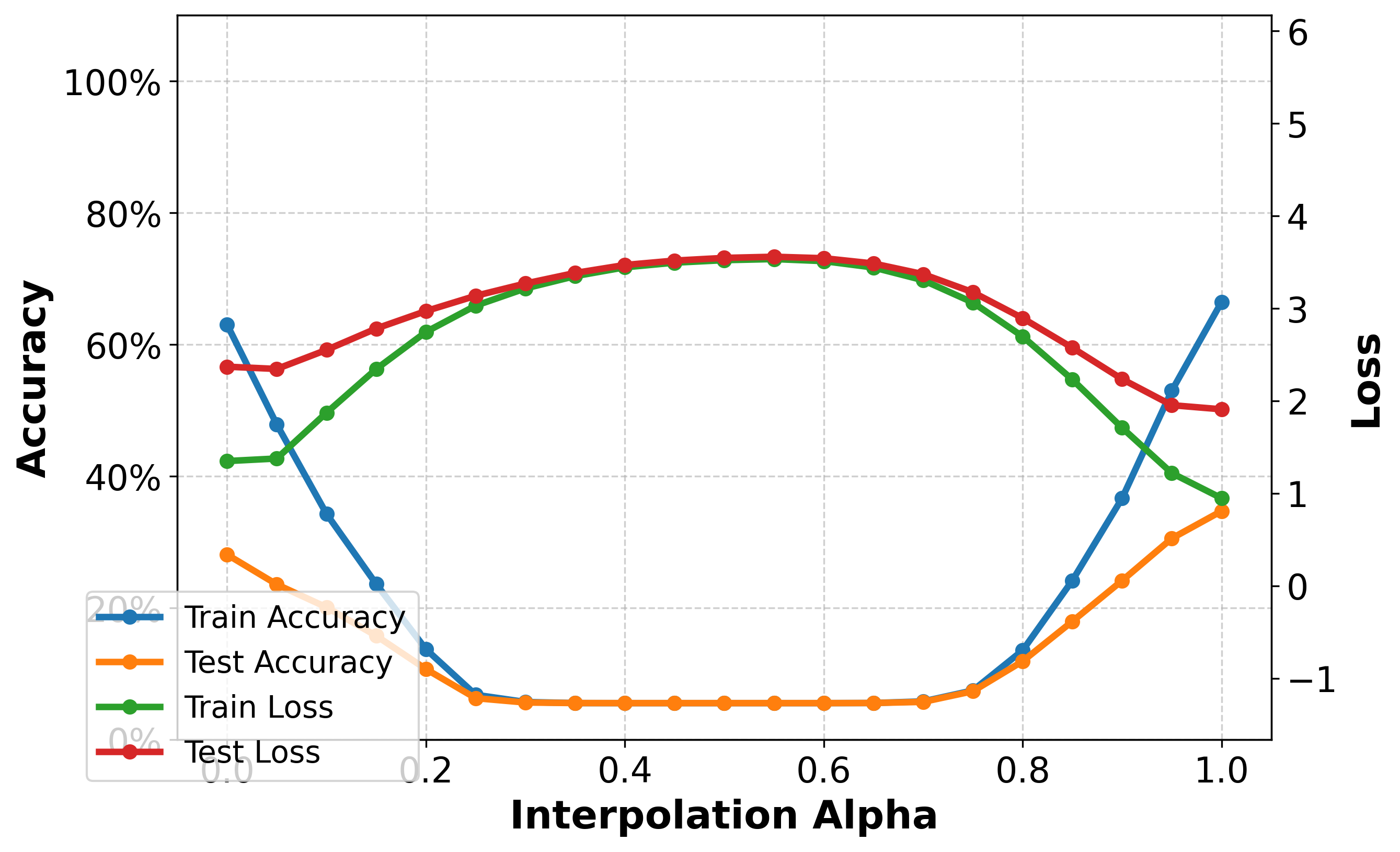}
        \caption{Chameleon}
    \end{subfigure}

    \caption{Performance of linear interpolations between two minima on eight real-world datasets. The x-axis represents the interpolation coefficient $\alpha$, and the y-axis shows train accuracy, test accuracy, train loss, and test loss.}
    \vspace{-0.2in}
    \label{fig: figure1}
\end{figure*}
\section{Main Analysis}
\label{sec: main}

In this section, we systematically study mode connectivity of GNNs. We begin by exploring the manifestation of mode connectivity in GNNs~(Section~\ref{sec:lmc}). Considering the non-Euclidean nature of graphs and the intertwined roles of structure and features, we proceed to investigate how these graph properties impact the mode connectivity of GNNs~(Section~\ref{sec:gp}). Finally, we develop a theoretical framework to explain the empirically observed phenomena (Section~\ref{sec:theory}) and provide further insights.

\begin{figure*}[!ht]\label{fig:Bezier}
    \centering
    \begin{subfigure}[b]{0.24\textwidth}
        \centering
        \includegraphics[width=\textwidth]{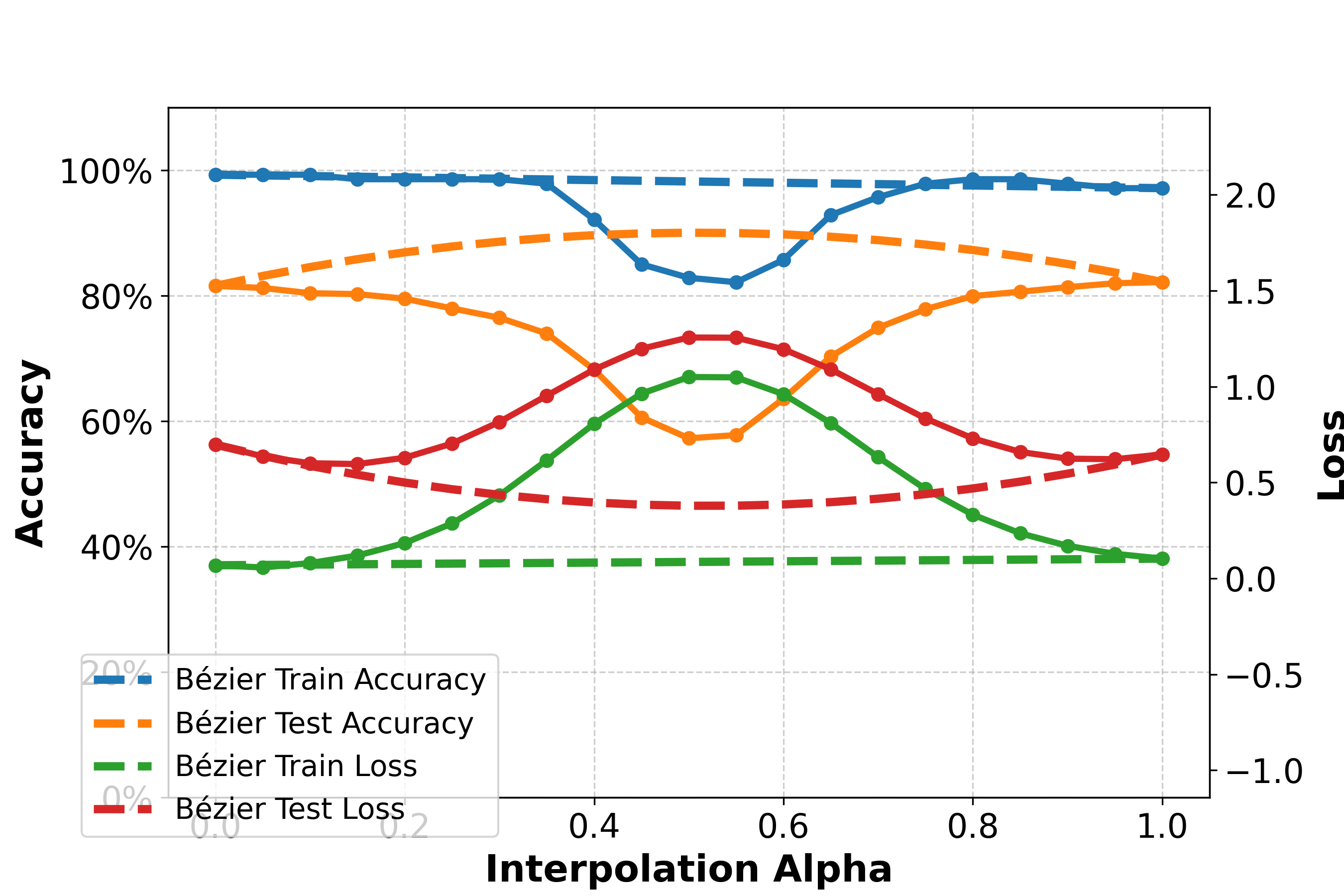}
        \caption{Cora}
    \end{subfigure}
    \begin{subfigure}[b]{0.24\textwidth}
        \centering
        \includegraphics[width=\textwidth]{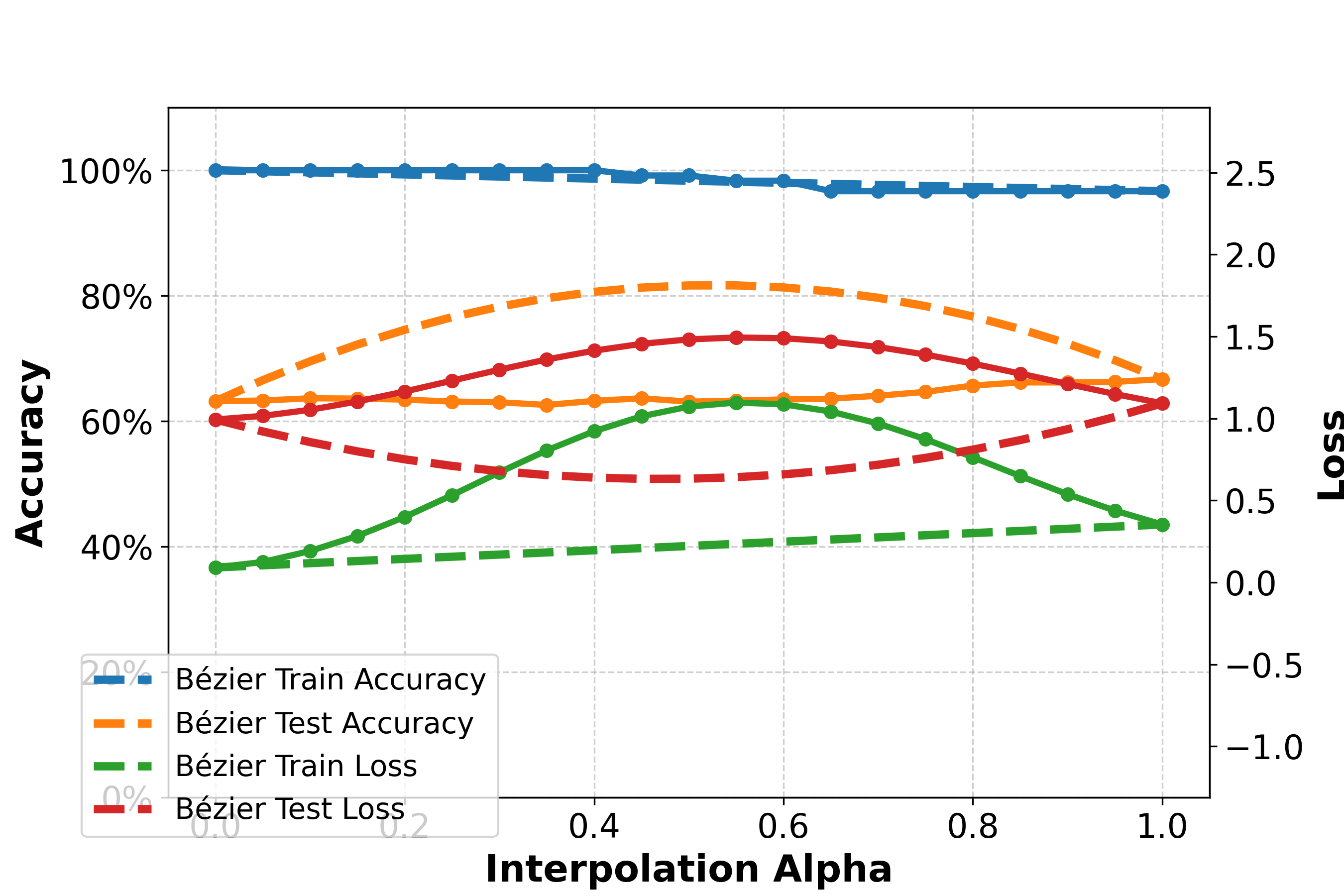}
        \caption{Citeseer}
    \end{subfigure}
    \begin{subfigure}[b]{0.24\textwidth}
        \centering
        \includegraphics[width=\textwidth]{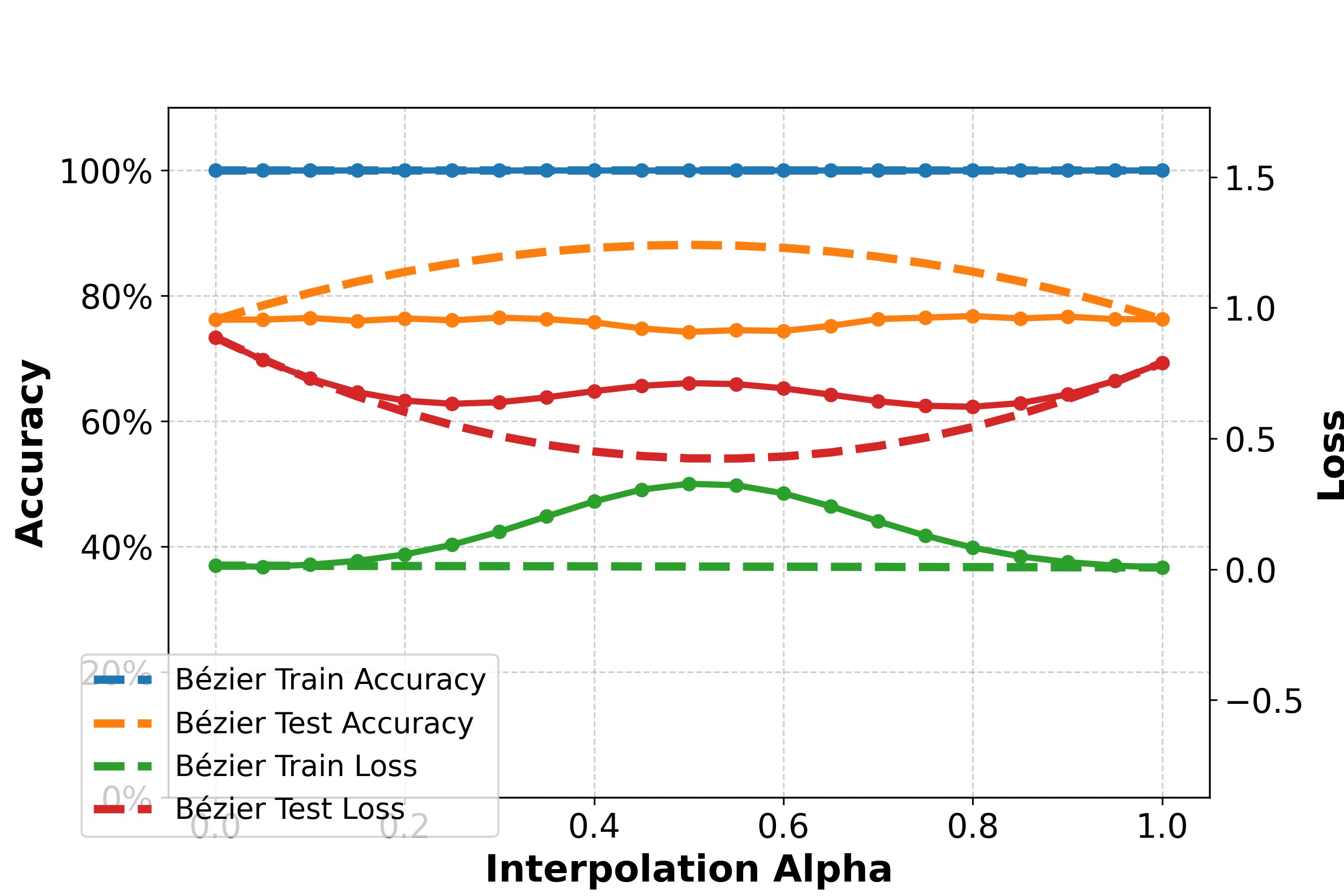}
        \caption{Pubmed}
    \end{subfigure}
    \begin{subfigure}[b]{0.24\textwidth}
        \centering
        \includegraphics[width=\textwidth]{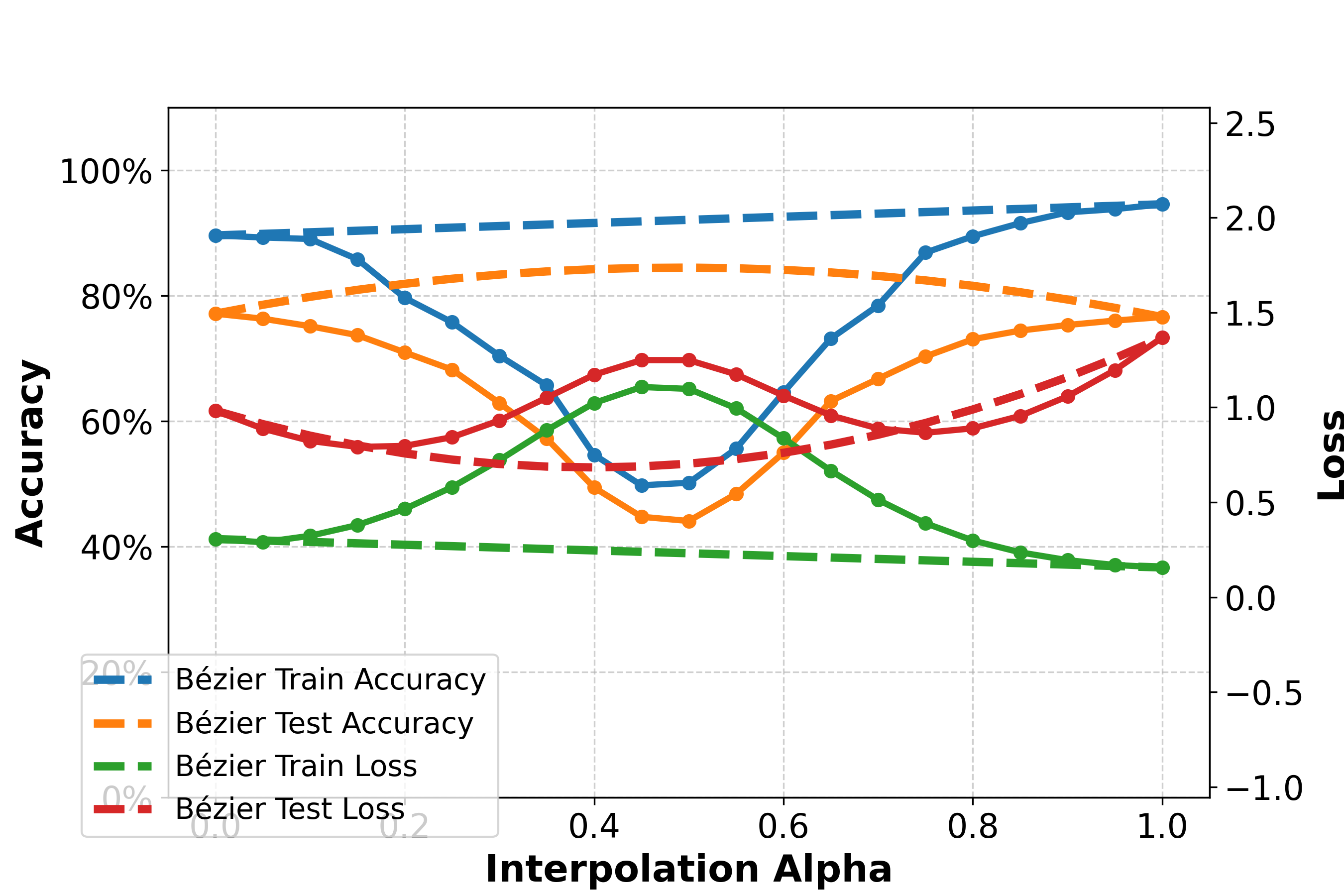}
        \caption{Wiki-CS}
    \end{subfigure}
    \vspace{5pt}
    \newline
    \begin{subfigure}[b]{0.24\textwidth}
        \centering
        \includegraphics[width=\textwidth]{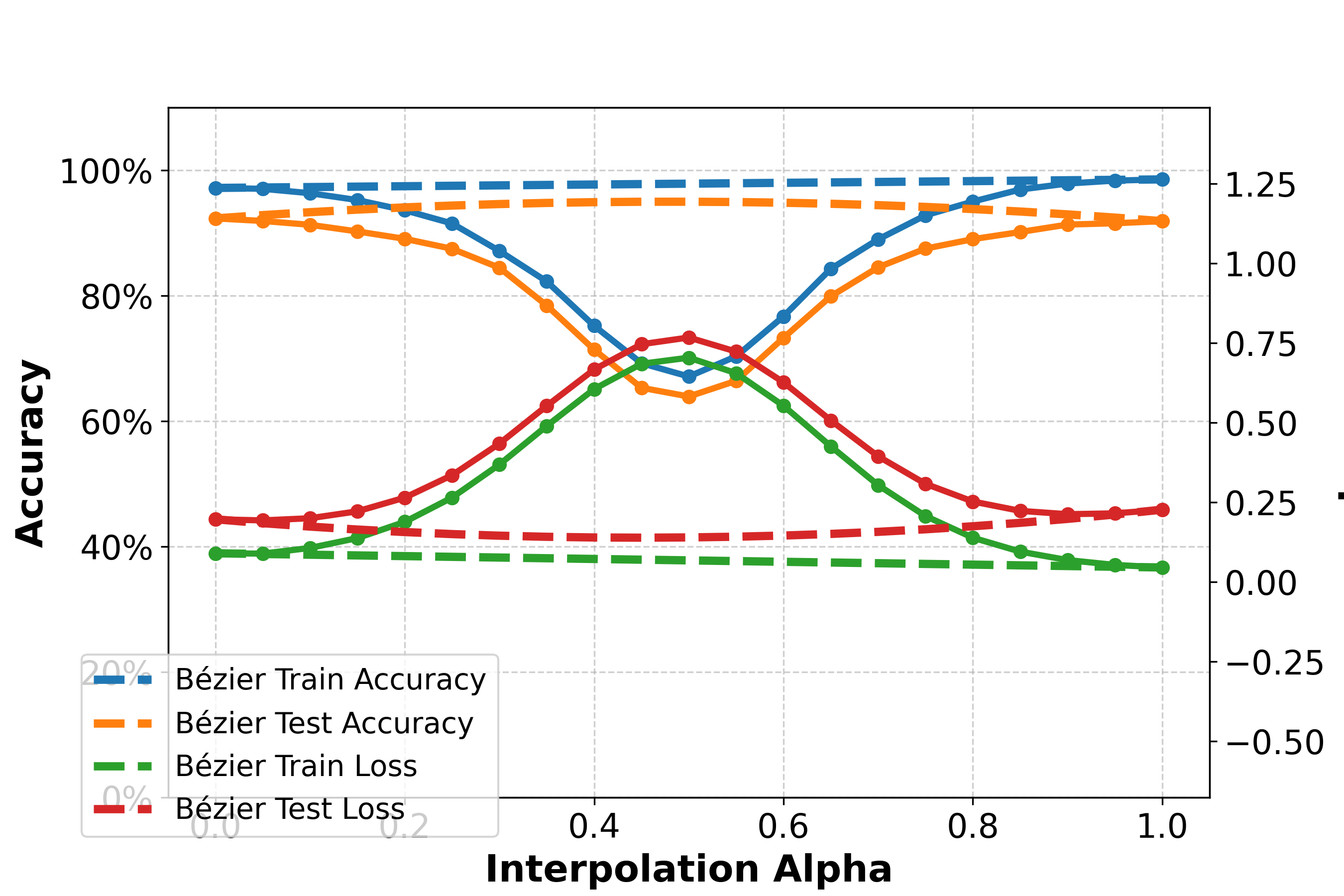}
        \caption{Coauthor-CS}
    \end{subfigure}
    \begin{subfigure}[b]{0.24\textwidth}
        \centering
        \includegraphics[width=\textwidth]{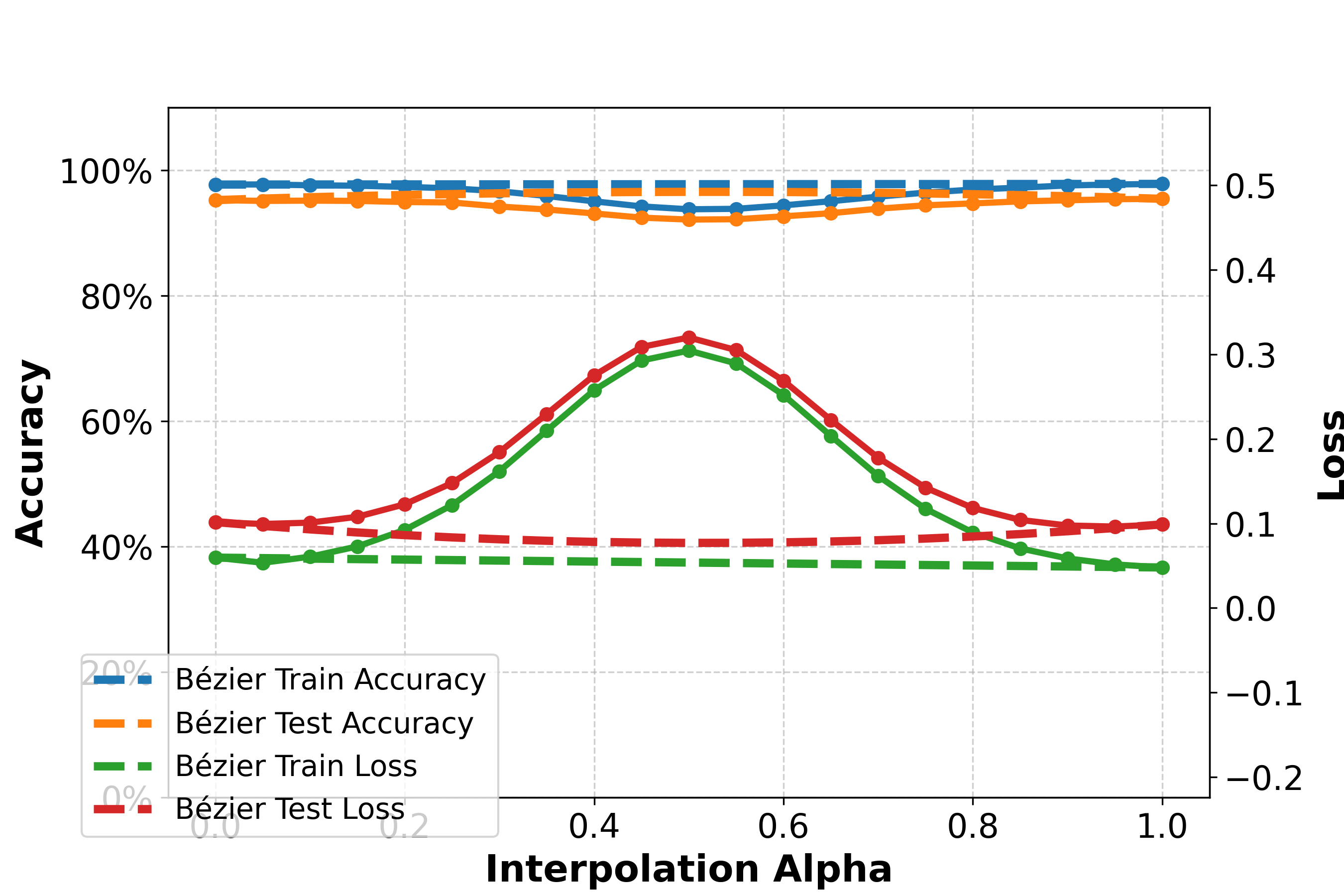}
        \caption{Coauthor-Physics}
    \end{subfigure}
    \begin{subfigure}[b]{0.24\textwidth}
        \centering
        \includegraphics[width=\textwidth]{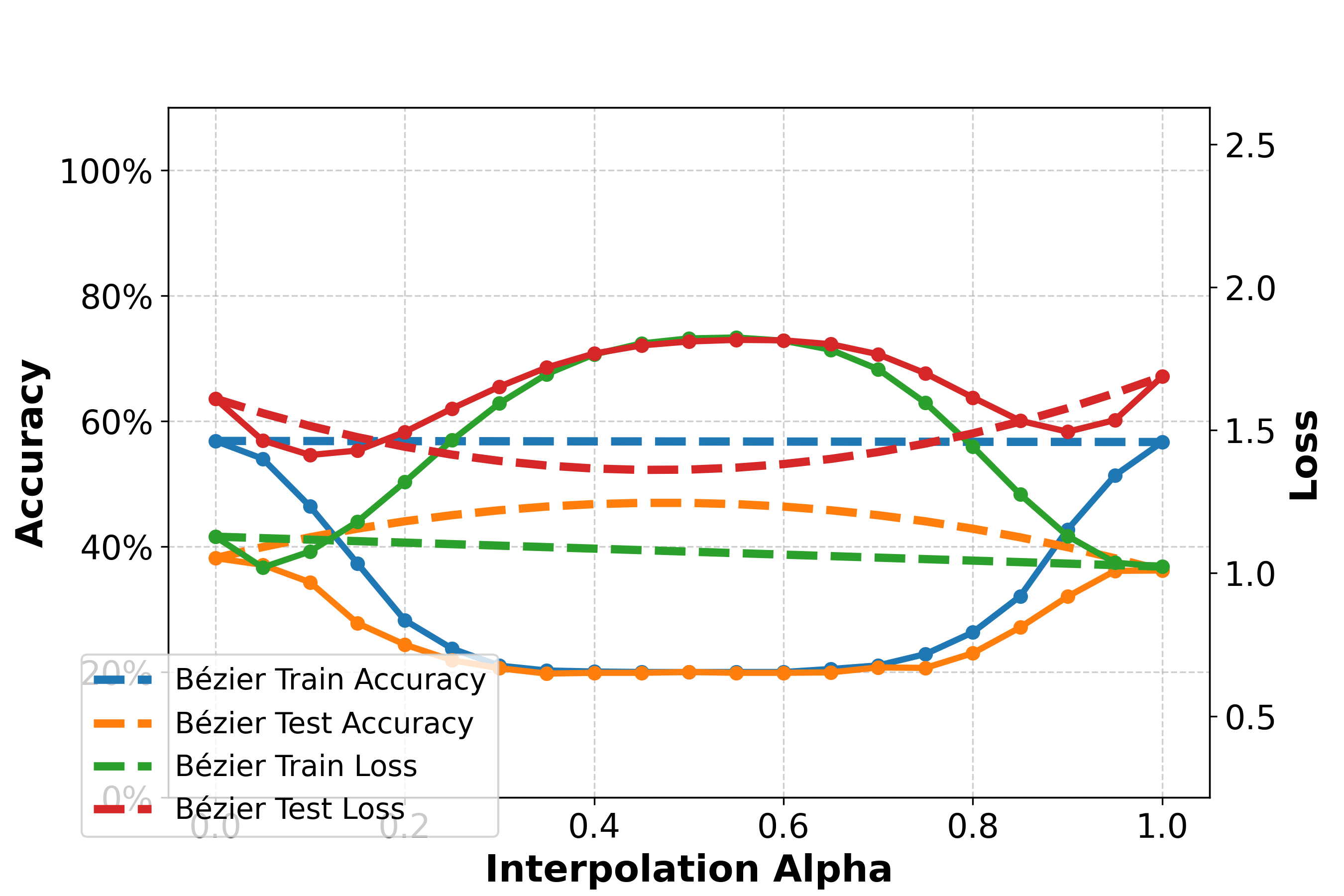}
        \caption{Squirrel}
    \end{subfigure}
    \begin{subfigure}[b]{0.24\textwidth}
        \centering
        \includegraphics[width=\textwidth]{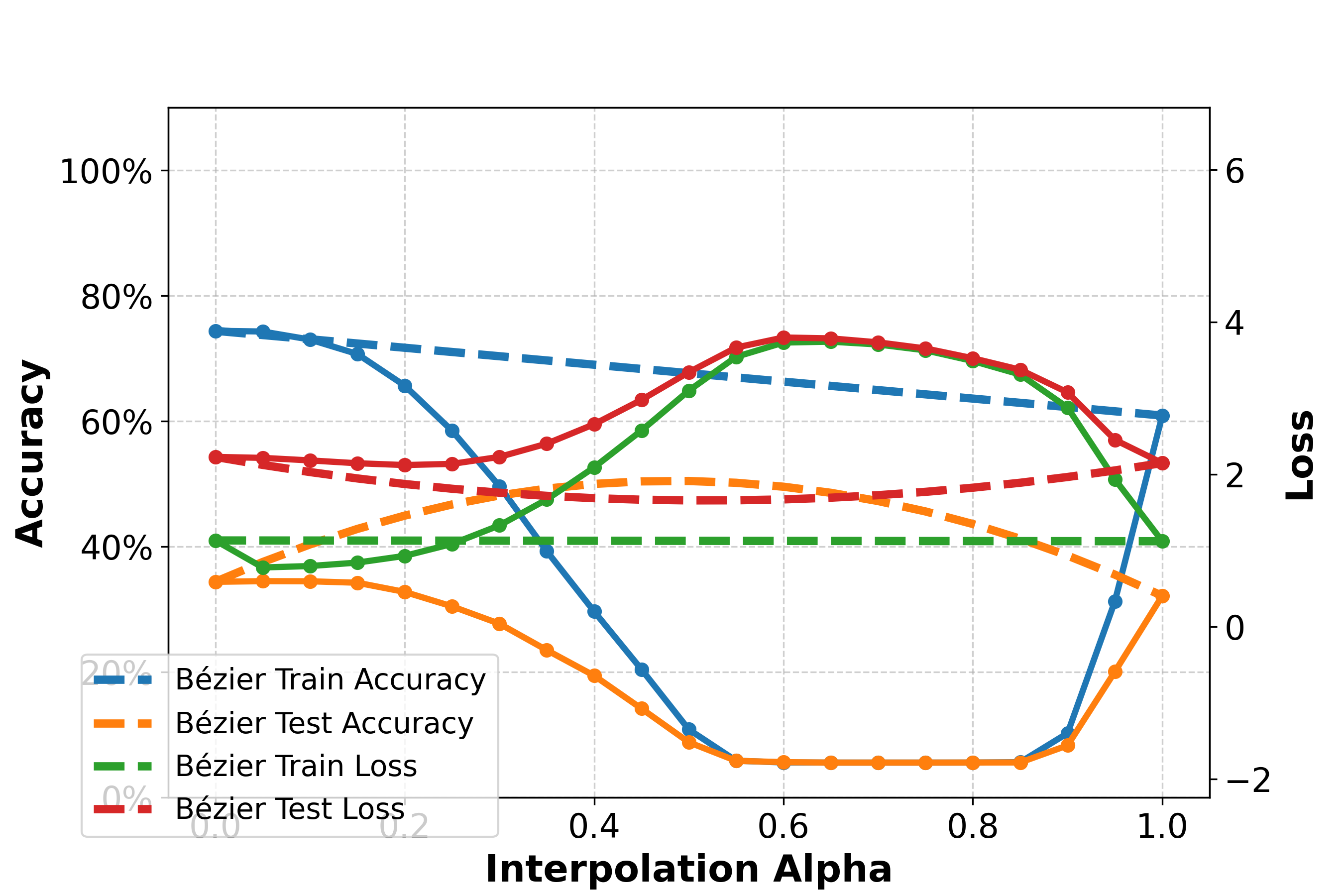}
        \caption{Chameleon}
    \end{subfigure}
    \caption{Performance of Quadratic Bézier Curve Interpolations Between Two Minima. This figure shows train/test loss and accuracy curves for quadratic Bézier curve interpolations across eight datasets. Compared to linear interpolation (Figure~\ref{fig: figure1}), Bézier curves better bypass loss barriers, indicating that GNN minima, while not always linearly connected, often lie on a smooth low-loss manifold.}
    \label{fig: figure2}
\vspace{-0.2in}
\end{figure*}

\subsection{Does Mode Connectivity Exist in Graph Neural Networks?}\label{sec:lmc}

While prior works have studied the mode connectivity for fully connected and convolutional neural network~\cite{garipov2018loss, draxler2018essentially}, the graph-dependent message-passing mechanisms of GNNs may influence the loss landscape and thus mode connectivity, which motivates us to first study the the existence and characteristics of mode connectivity in GNNs.


\noindent\textbf{Investigation protocol.} We adopt node classification as our research subject and utilize one of the most commonly adopted GNNs, GCN~\cite{Kipf2016SemiSupervisedCW}, as the backbone models, while also studying the influence of different model architectures. In terms of data selection, we consider $12$ graphs with different properties, including those exhibiting homophilic and heterophilic~\cite{Ma2021IsHA} characteristics.  Following the settings in ~\cite{luo2024classic}, we fix all hyperparameters except one, which we vary to generate different modes. Specifically, we investigate the impact of different initialization strategies, data orders (Appendix~\ref{app: A}), and model selections, as explored in \citet{garipov2018loss,qin2022exploring}. 
After generating these modes, we follow \citet{garipov2018loss} to fit the corresponding performance data points using linear interpolation and Bézier curve interpolation, where train accuracy, test accuracy, train loss, and test loss are measured. To ensure statistical reliability, all experiments are repeated three times with different random seeds, and we report the average performance. Additional details and supplementary results are provided in Appendix~\ref{app: A}.

\noindent\textbf{Observation 1: The GNN minima are usually connected but not always linearly connected.} 
After fitting the performance data points of different modes using linear interpolation, we observe a significant increase in both the training loss and the test loss (see Figure\ref{fig: figure1}). This indicates that although linear mode connectivity is a common phenomenon in fully connected networks (such as MLPs and CNNs)~\citep{frankle2020linear, yunis2022convexity}, it does not always hold true for GNNs. \textbf{First}, as shown in Figure\ref{fig: figure1}, 
linear interpolation often results in noticeable loss barriers. For example, the interpolation curves for datasets such as Cora and Citeseer are relatively smooth with only slight increases in loss, which suggests that the minima obtained during training are relatively well-connected in the parameter space. However, in datasets such as Coauthor-CS and Squirrel, we observe significant loss barriers - linear interpolation leads to a dramatic drop in model performance. This phenomenon indicates that for some datasets, the solutions obtained from training GNNs are more isolated and cannot be connected by a simple linear path. \textbf{Second}, although there is a lack of a linear path, most modes can still be connected by a simple polynomial curve, such as a quadratic Bézier curve. We use a quadratic Bézier curve to fit the performance data of different modes and observe that the loss does not increase significantly (see Figure \ref{fig: figure2}). This indicates that although GNNs may lack linear mode connectivity (LMC), in most cases their minima are not completely isolated but are distributed on a well-structured low-loss manifold and can be connected via learnable nonlinear paths.

\begin{figure}[ht]
    \centering
    \begin{subfigure}[b]{0.23\textwidth}
        \centering
        \includegraphics[width=\textwidth]{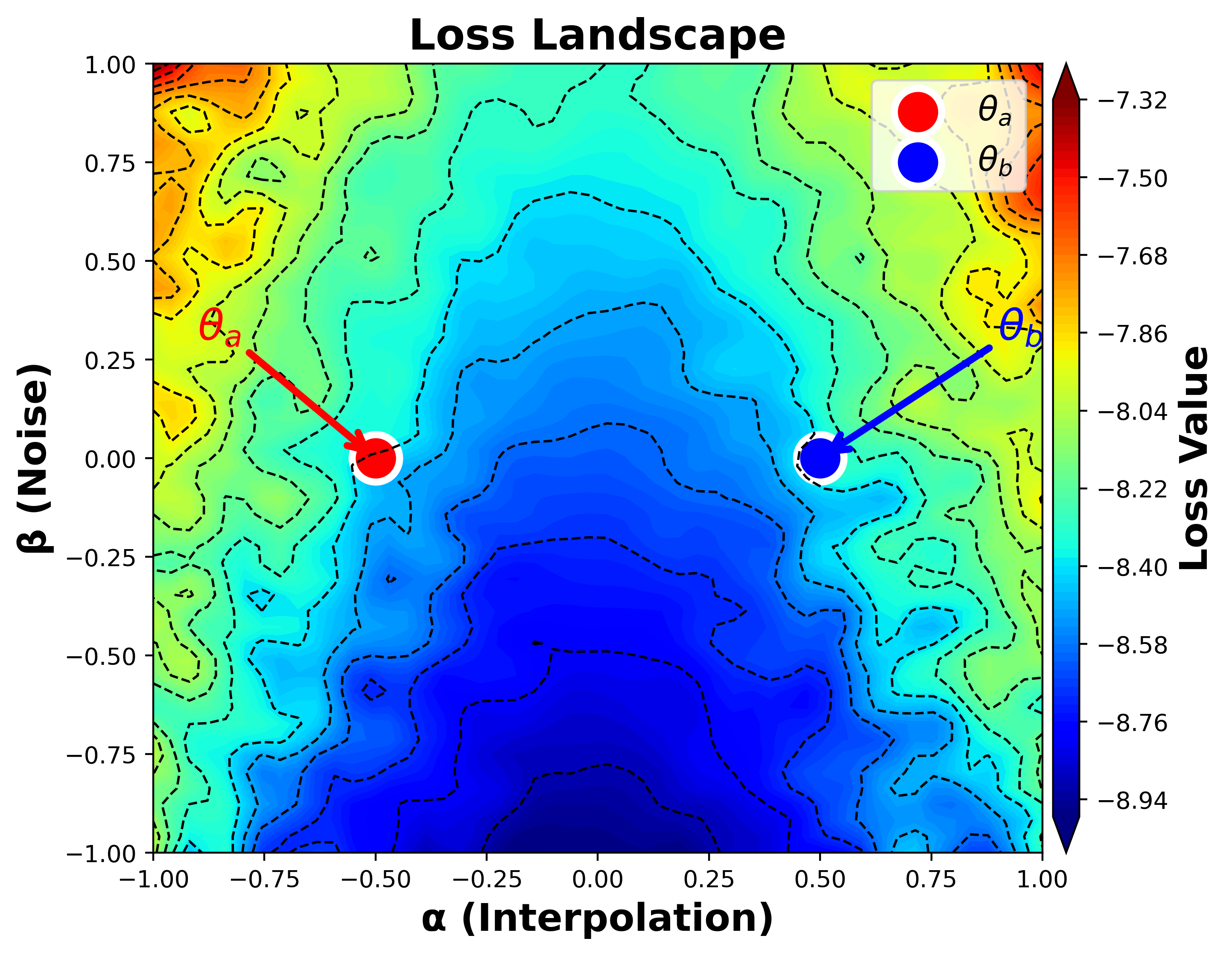}
        \caption{Pubmed}
    \end{subfigure}
    \begin{subfigure}[b]{0.23\textwidth}
        \centering
        \includegraphics[width=\textwidth]{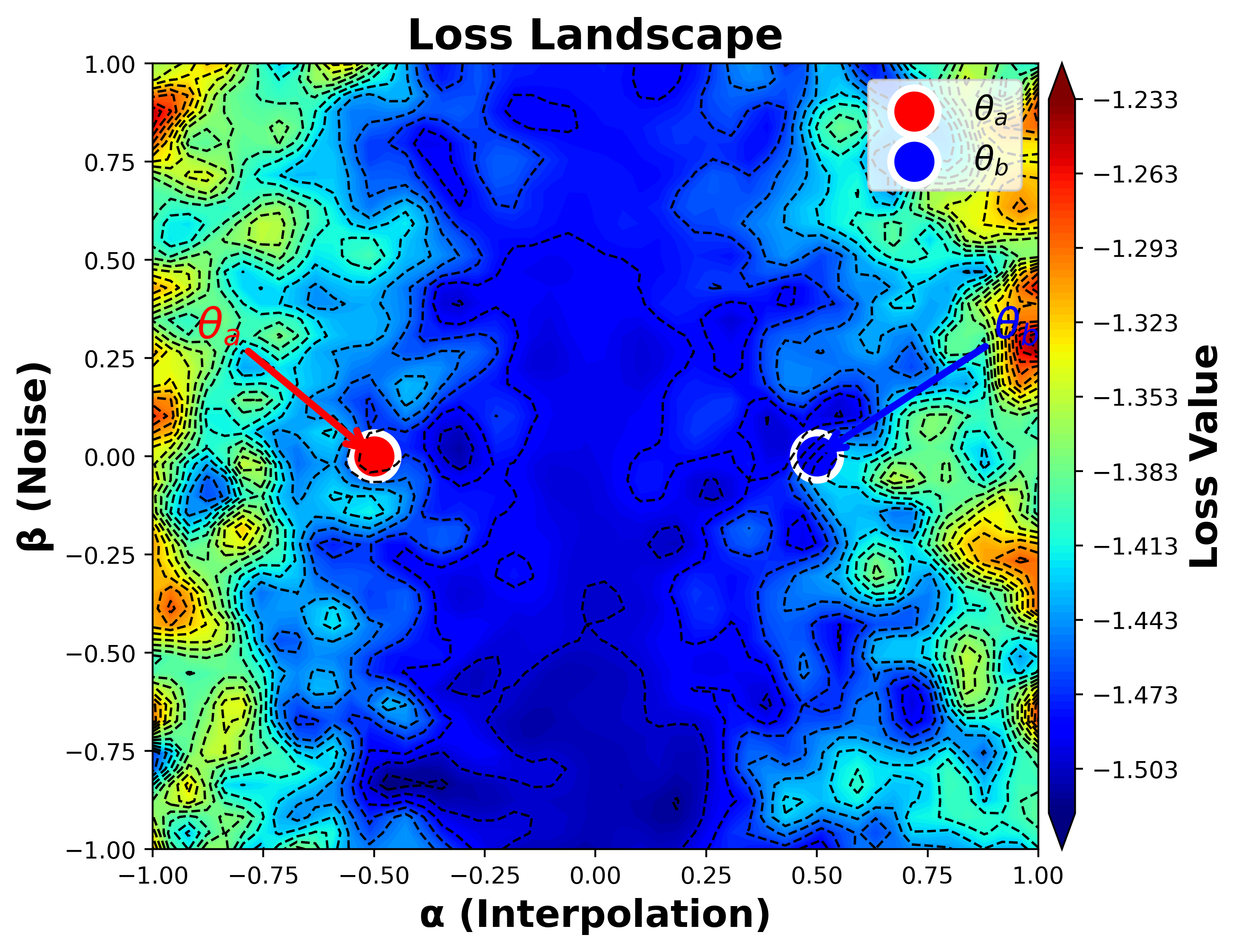}
        \caption{Squirrel}
    \end{subfigure}
    \vspace{-0.1in}
    \caption{Contour visualization of loss basin} 
    \vspace{-0.2in}
    \label{fig:contour}
    \end{figure}

\textbf{Further probing.} To further understand this phenomenon, we visualize the loss landscapes on different datasets (Figure \ref{fig:contour}). The loss contour map for the Pubmed dataset shows that its minima typically lie within the same smooth local loss basin, implying that the training process tends to converge to a connected low-loss region, allowing different minima to be connected via a low-loss path. In contrast, the Squirrel dataset exhibits a significantly more rugged loss landscape, with steeper boundaries for the local loss basins and minima that may be distributed across different disjoint basins. This explains why, in some datasets, the mode connectivity of GNNs is poor and why they cannot be effectively connected via linear interpolation.

\textbf{Implications for lottery ticket hypothesis and pruning.} It is noteworthy that the observed phenomenon is related to studies on linear mode connectivity (LMC), the Lottery Ticket Hypothesis, and pruning~\citep{frankle2020linear}. When studying the LMC of fully connected neural networks, pruning is shown to generally improve mode connectivity, making network weights easier to interpolate along a low-loss path~\citep{frankle2020linear}. However, our observations suggest that the pruning capability of GNNs may not be universal: on some datasets (e.g., Cora), the model weights are relatively smooth in the loss landscape and are easy to connect, whereas on other datasets (e.g., Squirrel) the loss landscape is more complex, and the minima may reside in different local basins, leading to pruned modes that are still difficult to connect. This finding emphasizes the role of the GNN model structure and its optimization strategy in mode connectivity and suggests that the existing LMC theory might not be directly applicable to all GNN architectures. This observation can potentially explain why the heterophilic graph is seldom considered in model-based graph condensation~\citep{jin2021graph, zheng2024structure}. 



\noindent\textbf{Observation 2.} \textbf{The interpolation curves present similar patterns for graphs coming from similar domains.} 
We then look into the detailed pattern difference among interpolation curves of different graphs. Surprisingly, we observe that datasets from different domains exhibit distinct patterns, and those from similar domains tend to have more similar curves. In particular, citation networks (such as Cora, CiteSeer, and PubMed) display a smoother interpolation process compared to co-authorship networks (such as Coauthor-CS and Coauthor-Physics), which may relate to the nature of each graph. In citation networks, papers typically cite other papers within the same or similar research fields, which results in a stronger clustering effect among nodes of the same class. In contrast, due to cross-domain collaborations, the local structure in co-authorship networks is more complex; certain nodes (researchers) might belong to multiple communities simultaneously, leading to blurred class boundaries. 


\textbf{Implications for measuring domain discrepancy of different graphs.} This observation suggests that mode connectivity can be effectively employed to gauge the domain discrepancy among different graphs, which is important for applications like graph foundation models~\citep{chen2024textspace, maoposition}. While traditional measures such as Maximum Mean Discrepancy (MMD)~\citep{JMLR:v13:gretton12a} or degree distribution capture only partial aspects of a graph's feature distribution or structural properties, model-based mode connectivity offers a far more nuanced and detailed assessment. The advantages of this approach and its applications are discussed further in Section~\ref{subsec:graph_similarity}. 



\begin{figure}
    \centering
    \includegraphics[width=1.0\linewidth]{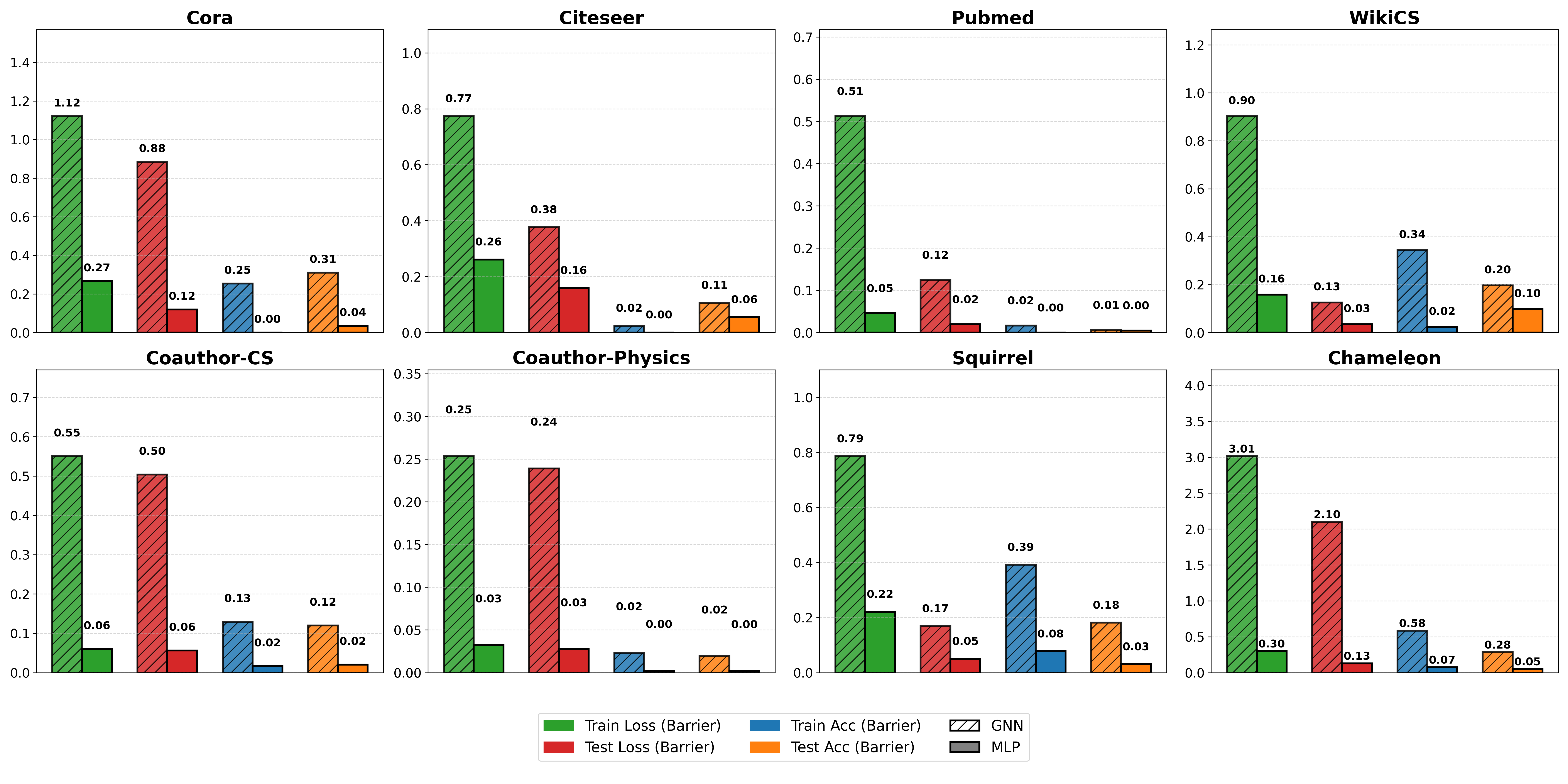}
    \vspace{-0.1in}
    \caption{Comparison between the loss barrier of GNN and MLP.}
    \label{fig:obs3}
    \vspace{-0.2in}
\end{figure}

In the previous study, we restrict the model backbone to GCN and explore the mode connectivity of the same model across different graphs. To further investigate the impact of GNN model architectures on mode connectivity, we replace GCN with MLP, GraphSAGE~\citep{hamilton2017inductive} and GAT~\citep{velickovic2017graph} models and study their mode connectivity. We observe
\noindent\textbf{Observation 3: Message passing significantly affects the mode connectivity while different aggregation design puts little influence.} 
As shown in Figure~\ref{fig:obs3}, when using the loss barrier to assess the mode connectivity of each model, there is a significant gap between the mode connectivity of the MLP and the GCN. In contrast, for GNNs with varying architectural designs, the loss barrier values remain similar (see Appendix~\ref{app: conv}). This phenomenon suggests that while message passing has a substantial impact on mode connectivity, the effect of the specific model design is minimal. Consequently, we are motivated to focus on investigating the influence of the graph data itself on mode connectivity, and in the next section, we will examine how graph properties affect mode connectivity.

\subsection{How Do Graph Properties Influence Mode Connectivity?}\label{sec:gp}
\label{sec: gp}
\begin{figure*}
    \centering
    \includegraphics[width=0.9\linewidth]{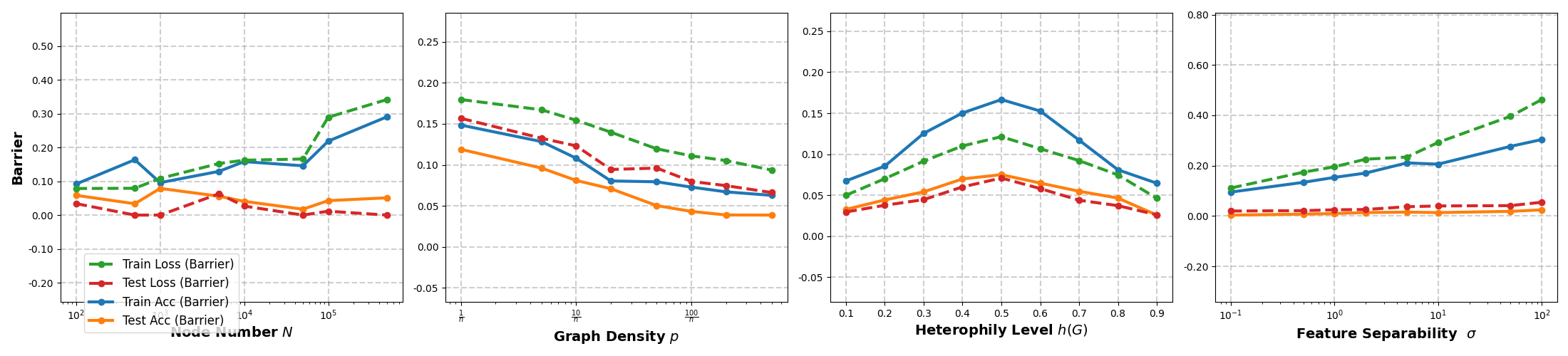}
    \vspace{-0.1in}
    \caption{The trend of mode connectivity, as measured by the barrier, with changes in graph properties.}
    \label{fig:gp}
    \vspace{-0.2in}
\end{figure*}

In this section, we further study how relational dependencies inherent in graphs affect mode connectivity of GNNs through the lens of graph properties. 

\noindent\textbf{Study protocol.} To come up with a controlled study in terms of graph properties, we first adopt a synthetic graph generative model CSBM. [add more experimental details here] We further consider the following key graph properties reflecting the graph structures and feature relationships 
\begin{itemize}[leftmargin=0pt, nosep]
    \item \textbf{Density of the graph} \textbf{{\(p = (p_{\mathrm{in}} + p_{\mathrm{out}})\)}},  which affects the connectivity of a graph. A denser graph implies a stronger structural relationship across different nodes. 
    \item \textbf{Homophily} \textbf{$h(G)= \frac{p_{in}}{p_{in} + p_{out}} $}, which
    refers to the tendency of nodes (vertices) with the same or similar labels to be more likely to connect to each other than to nodes with different labels.
    \item \textbf{Feature separability} \textbf{{\(\sigma\)}} reflects the difficulty of GNNs distinguishing the two different classes, and it can imply the influence of feature distribution on mode connectivity.
\end{itemize}

After fitting different GNNs on generated graphs, we measure the correlation between mode connectivity and corresponding graph properties, where we use loss barriers as a metric for the mode connectivity. As shown in Figure~\ref{fig:gp}, we observe:

\noindent\textbf{Observation 4.} \textbf{Both structural and feature properties of graphs affect the mode connectivity of GNNs.} As shown in Figure~\ref{fig:gp}, we find that 
\begin{enumerate}[nosep, leftmargin=20pt]
    \item \textit{Models trained on denser graphs present lower loss barriers, which implies better mode connectivity}.  Our experiments show that denser graphs—with higher intra-class connection probability \( p \)—exhibit lower loss barriers, indicating smoother transitions between minima. This supports the intuition that well-connected graphs enable stable feature propagation and reduce optimization variance. Conversely, sparse graphs lead to fragmented loss landscapes with higher barriers, particularly when \( p \) is very low, resulting in isolated subgraphs and less alignment across training runs.
    \item \textit{models trained on graphs with either high homophily or high heterophily exhibit better mode connectivity compared to graphs with medium homophily level.} This phenomenon arises from how class information propagates in different structural regimes. In strongly homophilic graphs, message passing reinforces class boundaries, yielding stable and aligned representations. In highly heterophilic graphs, despite inter-class edges, distinct feature distributions keep embeddings separable. However, intermediate homophily introduces label ambiguity, increasing optimization instability and resulting in rugged loss landscapes—a trend linked to the "mid homophily pitfalls" noted in \citet{luan2023graph}.
    \item \textit{Models trained on graphs with greater feature separability demonstrate improved mode connectivity.} Increasing \( \rho \) enhances mode connectivity by making same-class nodes more distinguishable and yielding consistent local optima. However, high intra-class variance or outliers raise loss barriers, indicating that unstable feature aggregation undermines training alignment. This underscores the interplay between graph topology and feature smoothness in optimization stability.
\end{enumerate}

\noindent\textbf{Relationship to real-world graphs.} The phenomenon observed in CSBM graphs provides valuable insights into the mode connectivity of GNNs trained on real-world graphs. Specifically, we observe that graphs characterized by stronger feature separability and higher homophily levels, such as Cora, Citeseer, and Pubmed, are associated with GNNs displaying lower loss barriers, further indicating enhanced mode connectivity. These findings suggest that graph density, feature separability, and homophily play significant roles in shaping the mode connectivity landscape of GNNs. In general, we find the phenomenons observed in synthetic graphs match ones observed in real-world graphs.

\subsection{Why Does Mode Connectivity Occur? A General Bound for GNNs}
\label{sec:theory}

While previous sections empirically investigated mode connectivity in GNNs, a rigorous theoretical understanding of why this phenomenon occurs remains elusive. In this section, we bridge this gap by establishing a theoretical framework that quantifies mode connectivity through a general bound on the loss barrier. 
Our analysis effectively validates previous experimental observations based on our understanding. We begin by formalizing the role of the graph structure in mode connectivity. One key factor is the spectral gap of the expected adjacency matrix, which controls the effective propagation of information in GNNs.
\begin{definition}[Spectral Gap~\cite{chung1997spectral}]
\label{def:spectral-gap}
Let \(\mathbb{E}[\hat{\mathbf{A}}]\) be the expected normalized adjacency matrix with eigenvalues
\[
\lambda_1(\mathbb{E}[\hat{\mathbf{A}}]) \ge \lambda_2(\mathbb{E}[\hat{\mathbf{A}}]) \ge \cdots \ge \lambda_n(\mathbb{E}[\hat{\mathbf{A}}]).
\]
The \emph{spectral gap} is defined as:
\[
\Delta \triangleq 1 - \max_{i\ge 2} |\lambda_i(\mathbb{E}[\hat{\mathbf{A}}])|.
\]
\end{definition}
 A larger spectral gap \(\Delta\) implies faster information propagation across the graph and better separability of node representations, which is crucial for stable optimization~\cite{chung1997spectral,abbe2018community}.

\begin{theorem}[General Bound of the Loss Barrier in \(L\)-Layer GNNs]
\label{thm:loss-barrier-GNN}
Let \(\theta_a\) and \(\theta_b\) be two sets of GNN parameters obtained under different initializations. Suppose the graph aggregation operator \(\hat{\mathbf{A}}\) has an effective propagation factor \(\lambda_{\mathrm{eff}}\) (related to the spectral gap) and let:
\[
N_l \triangleq \min\Bigl\{ \prod_{j=l+1}^{L} \|W_a^{(j)}\|,\,\prod_{j=l+1}^{L} \|W_b^{(j)}\| \Bigr\}.
\]
Then, the loss barrier satisfies:
\begin{equation}
\label{eq:final-bound}
\begin{aligned}
    B(\theta_a,\theta_b) &\le \max_{\lambda \in [0,1]} \Bigg\{ 
    (1-\lambda) C_L + \lambda\,L_\ell\,\lambda_{\mathrm{eff}}\,\|X\|\,\sum_{l=0}^{L} N_l  \|W_a^{(l)}-W_b^{(l)}\| 
    \Bigg\}
\end{aligned}
\end{equation}
where \(C_L\) captures higher-order curvature effects in the loss landscape, and \(L_\ell\) is the Lipschitz constant of the loss function.(Proof in Appendix \ref{the:3.2})
\end{theorem}

\begin{remark}[Implications for Mode Connectivity]
Theorem~\ref{thm:loss-barrier-GNN} formally establishes that the loss barrier is influenced by three fundamental factors:
\begin{itemize}
    \item \textbf{Initialization Variability:} The term 
    \(
    \|W_a^{(l)}-W_b^{(l)}\|
    \)
    quantifies differences in learned parameters at layer \(l\), arising from different initializations. A lower initialization variability leads to smaller barriers, facilitating smoother connectivity.
    \item \textbf{Weight Norm and Propagation:} The term \( N_l \) reflects how gradients propagate through layers. Larger \( N_l \) increases the loss barrier \( B(\theta_a, \theta_b) \), making the optimization landscape sharper. Techniques such as BatchNorm and LayerNorm help regulate \( N_l \), improving stability.~\cite{luo2024classic}
    \item \textbf{Graph Structural Influence:} The effective propagation factor \(\lambda_{\mathrm{eff}}\), which depends on the spectral gap \(\Delta\), controls how node features spread. A smaller \(\lambda_{\mathrm{eff}}\) (caused by a small spectral gap) results in less smooth minima transitions, increasing the loss barrier.
\end{itemize}
\end{remark}

By leveraging Theorem~\ref{thm:loss-barrier-GNN}, we establish a direct connection between the graph structure and mode connectivity.

\begin{corollary}[Graph Property Perspective of the Loss Barrier]
\label{cor:graph-property}
Let \(\theta_a\) and \(\theta_b\) be two sets of GNN parameters obtained under different initializations and/or mini-batch orders. Suppose the graph aggregation operator \(\hat{\mathbf{A}}\) satisfies:
\(
\lambda_{\mathrm{eff}} = 1 - \Delta + C_2 \sqrt{\frac{\log n}{d_{\min}}},
\)
where the spectral gap is:
\(
\Delta \triangleq 1 - \max_{i\ge 2} |\lambda_i(\mathbb{E}[\hat{\mathbf{A}}])|.
\)
Then, the loss barrier satisfies:
\[
B(\theta_a,\theta_b) \le O\Bigl( L_\ell\, \lambda_{\mathrm{eff}}\, \|X\| \Bigr)
= O\Biggl( L_\ell\|X\|\cdot \left[1 - \Delta + C_2 \sqrt{\frac{\log n}{d_{\min}}}\right]  \Biggr).
\]
\end{corollary}

This result suggests that a larger spectral gap \(\Delta\) leads to a smaller loss barrier, facilitating better mode connectivity and generalization.

\begin{proposition}[Loss Barrier in Node Classification on CSBM Datasets]
\label{prop:loss-barrier-CSBM1}
Let \(\theta_a\) and \(\theta_b\) be two sets of GNN parameters trained under different conditions. Suppose the graph follows the Community Stochastic Block Model (CSBM) with intra-class and inter-class connection probabilities \(p_{\mathrm{in}}\) and \(p_{\mathrm{out}}\), respectively. The loss barrier then satisfies the upper bound:
\begin{equation}
    \begin{aligned}
\label{eq:loss-barrier-community}
B(\theta_a,\theta_b) &\le O\Biggl( \sigma \sqrt{d \log n}\cdot\Bigl[ C_2 \sqrt{\frac{\log n}{d_{min}}} - \frac{(h(G)-\frac{1}{2})^2}{C_1}\cdot(p_{\mathrm{in}} + p_{\mathrm{out}})  \Bigr] \Biggr).
    \end{aligned}
\end{equation}
Here, \(C_1,C_2\) are constants.
\end{proposition}

Detailed proof is in Appendix \ref{pro:3.5}.
The theoretical results align well with empirical observations. Experiments show that graphs with strong homophily or heterophily (higher \(h(G)\)) exhibit lower loss barriers, supporting the hypothesis that mode connectivity is enhanced under these conditions. Corollary~\ref{cor:graph-property} provides a quantitative explanation for this by linking the spectral gap to loss landscape smoothness. Furthermore, Proposition~\ref{prop:loss-barrier-CSBM1} suggests that increasing graph density (higher \( p_{\mathrm{in}} + p_{\mathrm{out}} \)) reduces the loss barrier, leading to smoother optimization landscapes. This is empirically validated by observing that denser graphs result in lower barriers and more stable training. These findings establish a principled connection between graph structure and GNN optimization, offering valuable insights into how structural properties influence generalization and mode connectivity.

\section{Implications for the generalization of GNNs}
\label{sec: general}

After thoroughly investigating GNN mode connectivity, and with Observations 2 and 4 highlighting its relationship to GNN generalization behavior across both single and diverse graphs, we now leverage mode connectivity to derive a generalization bound for GNNs and, subsequently, to propose a mode connectivity-based metric for quantifying graph domain discrepancy.

\subsection{How Does Mode Connectivity Reflect Generalization Performance?}
Understanding generalization in GNNs remains a fundamental challenge, as their performance depends on both graph topology and node features. Specifically, while previous studies~\cite{keskar2016large,tatro2020optimizing,juneja2022linear} in deep learning suggest that flatter minima lead to better generalization, it remains unclear whether this holds for GNNs due to their reliance on relational structures. To address this, we establish a formal connection between mode connectivity and generalization by analyzing the role of mode connectivity geometry in GNN training. The barrier can quantify the degree of sharpness between minima, where a higher barrier suggests a more isolated minimum with a steeper surrounding manifold.

\begin{theorem}[Generalization Bound via Loss Barrier]  
\label{thm:generalization-barrier}  
 Let \(\theta^T_a,\theta^T_b\) be the model parameter obtained after \(T\) training iterations under different initialization, with \(m\) labeled and \(n-m\) unlabeled data samples . Then, for any \(\delta \in (0,1)\), with probability at least \(1 - \delta\), the generalization gap satisfies  
\begin{equation}
\Delta_{\text{gen}} \leq O \bigg( 8 B(\theta_a, \theta_b)\cdot \frac{n^{\frac{3}{2}}}{m(n-m)} \cdot \log(c(T)) T^\rho \log \frac{1}{\delta} \bigg).
\end{equation}
\end{theorem}
Detailed proof is in Appendix \ref{the:4.1}.
This bound formally establishes that the generalization gap is controlled by the mode connectivity barrier \( B(\theta_a, \theta_b) \), along with dataset-dependent terms and the training dynamics. Several key insights are suggested from this result:
1. The bound directly links the loss barrier to generalization, indicating that a smaller barrier leads to stronger generalization. 2. The dependence on the labeling ratio \( (m+u)^{3/2} / (mu) \) highlights the importance of labeled data in GNN generalization.

\begin{figure}
    \centering
    \includegraphics[width=0.7\linewidth]{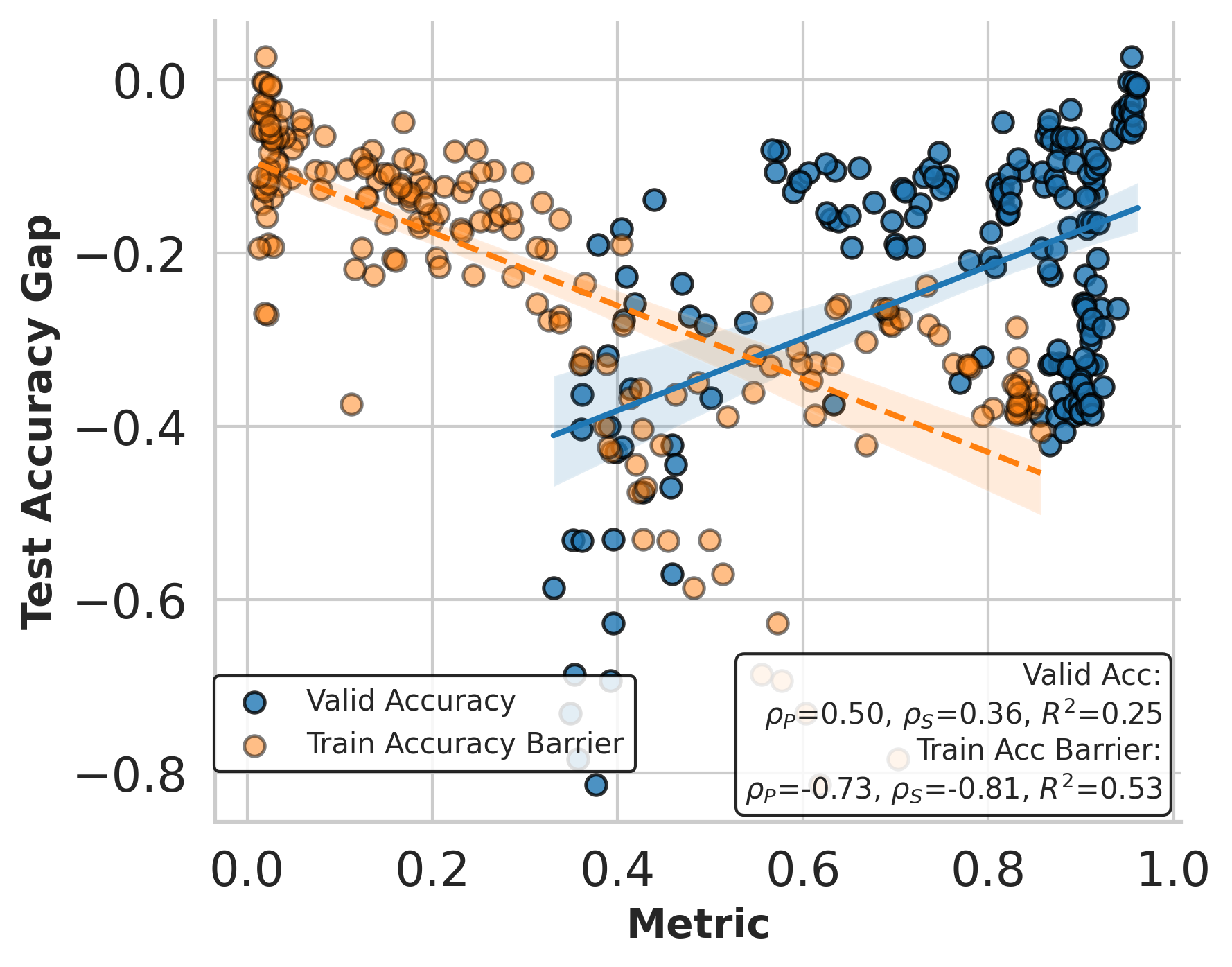}
    \vspace{-0.1in}
    \caption{Train accuracy barrier presents a stronger correlation to generalization gap compared to the commonly used validation accuracy. $\rho_p$ means Pearson correlation coefficient, $\rho_s$ means Spearman's rank correlation, while $R^2$ means Coefficient of determination. }
    \label{fig:compare}
     \vspace{-0.3in}
\end{figure}

To validate the theorem, we measure the correlation between the barrier value and the generalization gap. Also, we compare the correlation between the training accuracy barrier and the generalization gap to that between the validation accuracy and the generalization gap, where the latter is one of the most commonly used metrics for measuring overfitting and selecting model checkpoints. As illustrated in Figure~\ref{fig:compare}, where Each data point represents a trained model, we find that the training accuracy barrier demonstrates a stronger correlation with the generalization gap compared to the widely used validation accuracy. This suggests its potential utility as an indicator for selecting model checkpoints and mitigating overfitting. These findings collectively highlight that mode connectivity serves as a fundamental characteristic of GNN generalization, independent of specific architectures or hyperparameters.



\subsection{Inter-Domain Graph Similarity via Mode Connectivity}
\label{subsec:graph_similarity}

Inspired from {Observation 3}, where the way trained minima are connected in the {parameter space} may implicitly capture {domain-specific properties}, we study how to come up with a mode connectivity-based metric to measure domain discrepancy. Measuring similarity between different {graph domains} is a fundamental problem in transfer learning, where an appropriate similarity metric helps guide domain adaptation. Prior works~\cite{JMLR:v13:gretton12a, lee2019domain} have proposed various {structure-based} and {feature-based} metrics, yet these are often {heuristic and predefined}, lacking validation from the model’s training dynamics. In contrast, mode connectivity can reflect complex training dynamics and simultaneously consider both features and graph structures. We first introduce the metrics we design and then validate its effectiveness. 

{\bf Mode Connectivity Distance.}
\label{subsubsec:mc_distance} Consider two graphs, $G^1$ and $G^2$, trained under the same GNN, leading to minima $\theta^1_a, \theta^1_b$ for $G^1$ and $\theta^2_a, \theta^2_b$ for $G^2$. The {mode connectivity curve} between two minima is 
\(
    \mathcal{L}(\alpha) = \mathcal{L}(\phi(\alpha)), \quad \text{where } \phi(\alpha) = (1 - \alpha) \theta_a + \alpha \theta_b, \quad \alpha \in [0,1].
\)
So we treat $\mathcal{L}(\alpha)$ as a {distribution} over interpolation values $\alpha$, capturing the smoothness of the loss landscape between minima. The discrepancy between two graphs can then be quantified by the \textbf{Wasserstein-1 distance}~\citep{ramdas2017wasserstein} between their respective loss distributions:
\begin{equation}
    d_{\text{MC}}(G^1, G^2) = W_1\big(\mathcal{L}^{G^1}(\alpha), \mathcal{L}^{G^2}(\alpha)\big).
\end{equation}

{\bf Theoretical Justification: Upper Bounding Transferability Gap.}
\label{subsubsec:theory_transfer} We then theoretically show that $d_{\text{MC}}(G^1, G^2)$ provides an {upper bound} on the transferability performance gap between the two domains:
\begin{equation}
    \Delta_{da} \leq C \cdot O\bigg(d_{\text{MC}}(G^1, G^2)\bigg),
\end{equation}
for some constant $C$ dependent on model complexity and domain properties. Crucially, this result holds \textbf{independent of graph size}, indicating that a smaller mode connectivity distance implies {higher transferability}. The proof is shown in Appendix~\ref{the:final}.

\begin{figure}
    \centering
    \includegraphics[width=1.0\linewidth]{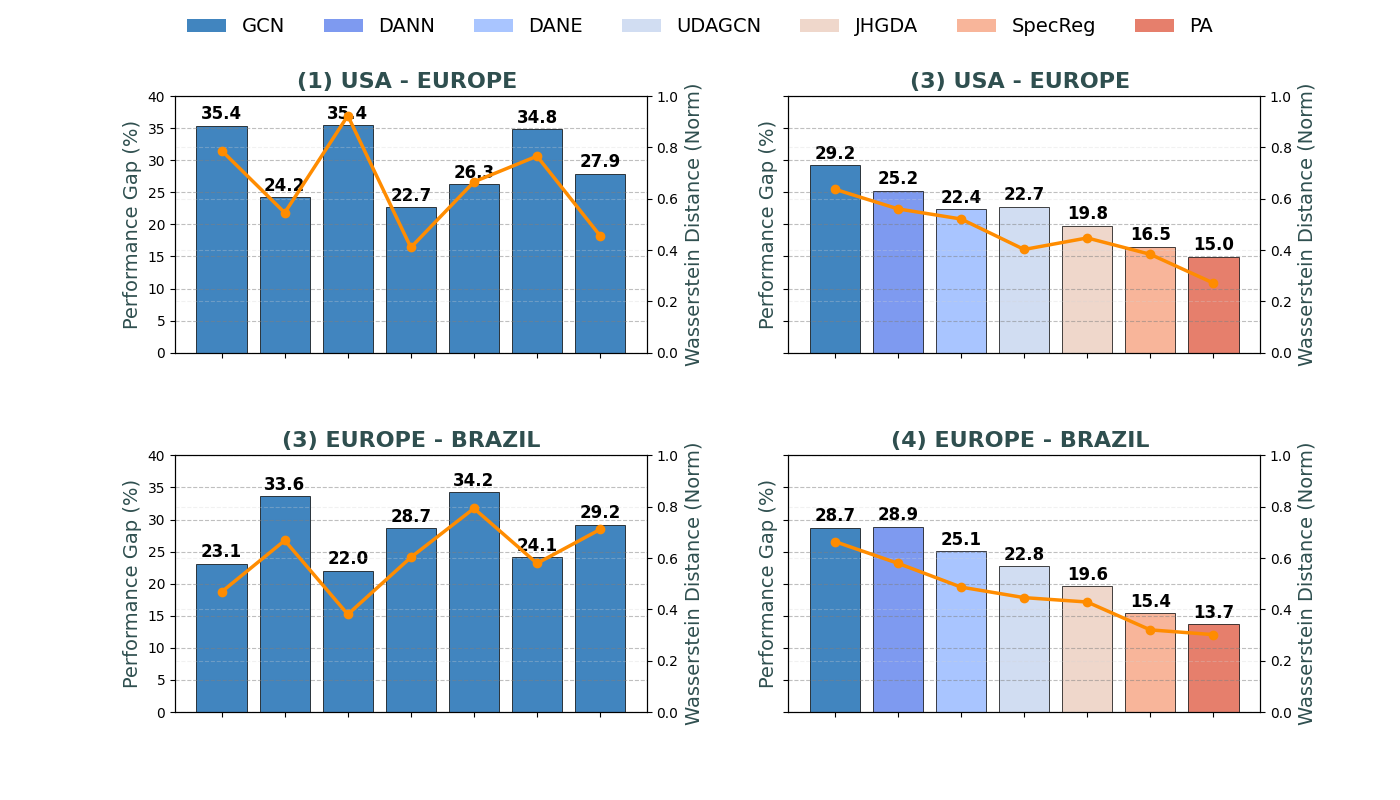}
\caption{Correlation between source/target performance gap and our proposed $d_{\text{MC}}(G^1, G^2)$. Cases 1 and 3 depict transfer learning across subgroups, while Cases 2 and 4 illustrate transfer learning across full graphs with different .}
    \label{fig:domainada}
    \vspace{-0.2in}
\end{figure}

{\bf Empirical Verification.}
\label{subsubsec:empirical_results} To empirically validate our theoretical insights, we investigate unsupervised graph domain adaptation~\citep{wu2020unsupervised}. Specifically, we train a GNN model on a source graph and then transfer it to a target graph. We evaluate several common graph domain adaptation methods, including vanilla transfer, DANN~\citep{ganin2016domain}, DANE~\citep{song2020domain}, UDAGCN~\citep{wu2020unsupervised}, JHGDA~\citep{shi2023improving}, and SpecReg~\citep{you2023graph}. We then analyze the correlation between the source/target performance gap and our proposed $d_{\text{MC}}$ metric. To generate statistically significant results for vanilla transfer, we consider transferring between $7$ subgroups of the source and target graphs. For other baseline methods, we directly consider transfer learning on the full graphs. The results are shown in Figure~\ref{fig:domainada}, and  key observations include:
\begin{itemize}[nosep, leftmargin=20pt]
    \item \textbf{Strong Correlation}: $d_{\text{MC}}(G^1, G^2)$ exhibits a \textbf{positive correlation} with the empirical performance gap across all scenarios. 
    \item \textbf{Effect of Domain Alignment}: Applying \textbf{domain alignment} techniques demonstrably lowers $d_{\text{MC}}$, visually demonstrating their effectiveness in enhancing knowledge transfer through the reshaping of the parameter space geometry.
\end{itemize}

These results underscore the {geometric perspective} of domain adaptation and suggest that {mode connectivity analysis} provides a principled approach to quantify and enhance {transfer learning across graph domains}.


\section{Related Work}
\label{sec: rw}

    Graph Neural Networks (GNNs) have been extensively studied from various theoretical perspectives, including expressive power~\citep{maron, xupowerful}, generalization properties~\citep{verma2019stability, aminian2024generalization}, and training stability~\citep{oono2019graph}. While these works provide valuable insights, they mainly focus on spectral properties and convergence behavior, leaving open questions about the structure of solutions found by different training runs. A thorough survey for relevant literature is in \citet{jegelka2022theory, huang2024foundations}. Another key research direction involves analyzing how GNNs propagate and aggregate information across graph structures~\citep{loukas2019graph, wu2019simplifying}. 
    However, a fundamental question remains unexplored: how do independently trained GNNs relate to each other in the parameter space? This study aims to bridge this gap.

    Mode connectivity refers to the existence of low-loss paths between different trained neural networks, demonstrating that solutions found by independent runs of training are not necessarily isolated~\citep{freeman2016topology, garipov2018loss, draxler2018essentially}. 
    A particularly important variant, linear mode connectivity (LMC), occurs when such connections are nearly linear~\citep{frankle2020linear, entezari2021role}. LMC has been widely observed in CNNs and transformers, with applications in model ensembling, adversarial robustness, and transfer learning~\citep{juneja2022linear, qin2022exploring, abdollahpourrostam2024unveiling}. Recent studies have also extended connectivity analysis to input space representations~\citep{vrabel2024input} and explored higher-dimensional connectivity structures~\citep{anonymous2024revisiting}. However, mode connectivity in GNNs remains largely unexplored. 

\section{Conclusion}
\label{sec: conclusion}
In this paper, we systematically study the mode connectivity behavior of GNNs and reveal that they exhibit distinct non-linear mode connectivity, primarily driven by graph structure rather than model architecture. This discovery deepens our understanding of GNN training dynamics and its link to improved generalization, paving the way for refined training methodologies and cross-domain adaptation techniques. Meanwhile, our research on mode connectivity has primarily focused on the node classification task. A promising future direction is to investigate the mode connectivity exhibited by GNNs on link prediction and graph-level tasks.

\bibliographystyle{ACM-Reference-Format}
\bibliography{main}

\appendix
\onecolumn
\section{More results about Mode connectivity in GNN}
\label{app: A}

\subsection{The performance of linear interpolations 
between two minima.}

\begin{figure*}[!ht]
    \centering
    \begin{subfigure}[b]{0.24\textwidth}
        \includegraphics[width=1.0\textwidth]{fig/init/init/cora_gcn_plot.png}
        \caption{Cora}
    \end{subfigure}
    \begin{subfigure}[b]{0.24\textwidth}
        \includegraphics[width=1.0\textwidth]{fig/init/init/citeseer_gcn_plot.png}
        \caption{CiteSeer}
    \end{subfigure}
    \begin{subfigure}[b]{0.24\textwidth}
        \includegraphics[width=1.0\textwidth]{fig/init/init/pubmed_gcn_plot.png}
        \caption{PubMed}
    \end{subfigure}
    \begin{subfigure}[b]{0.24\textwidth}
        \includegraphics[width=1.0\textwidth]{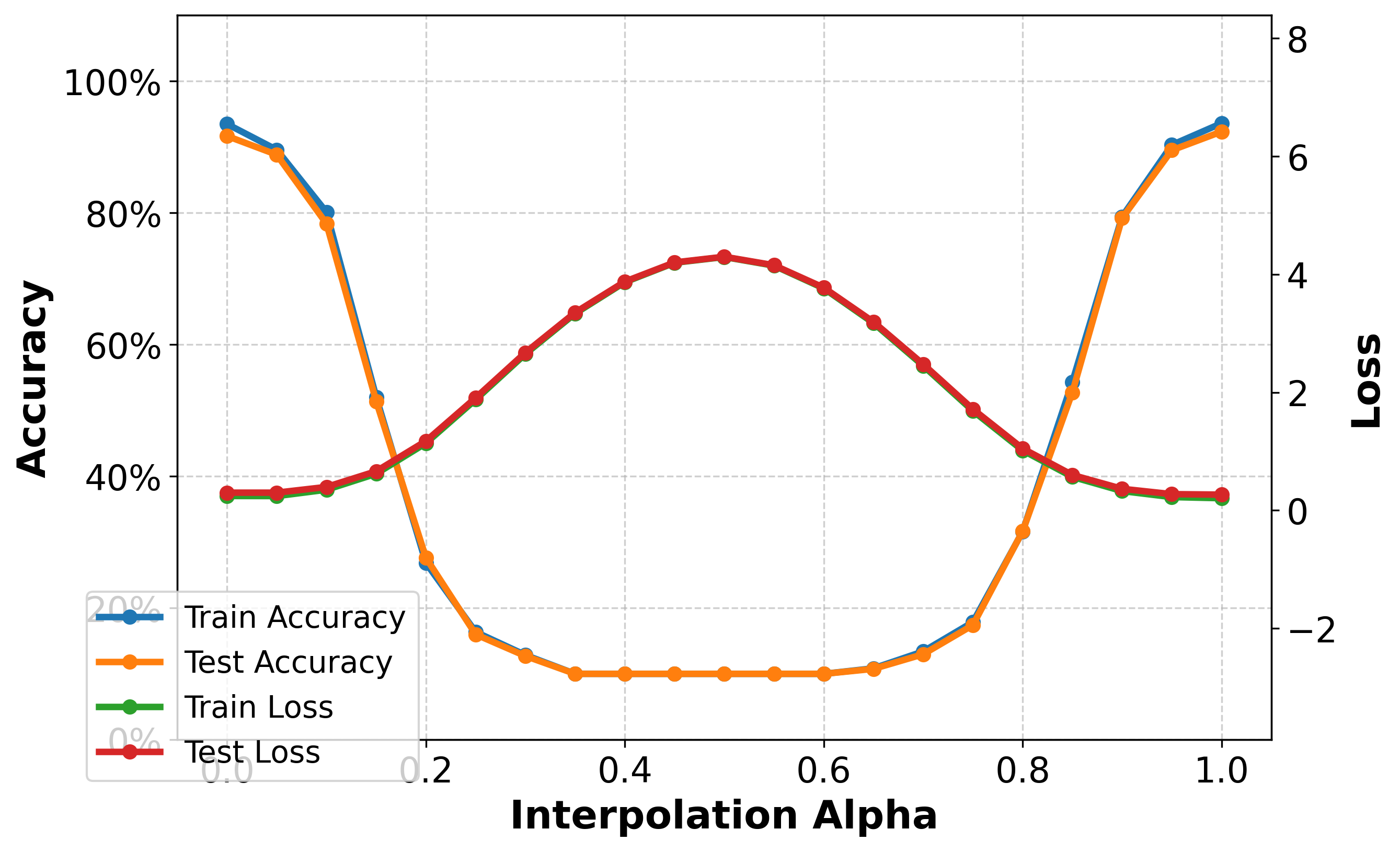}
        \caption{Amazon-Computer}
    \end{subfigure}
    \begin{subfigure}[b]{0.24\textwidth}
        \includegraphics[width=1.0\textwidth]{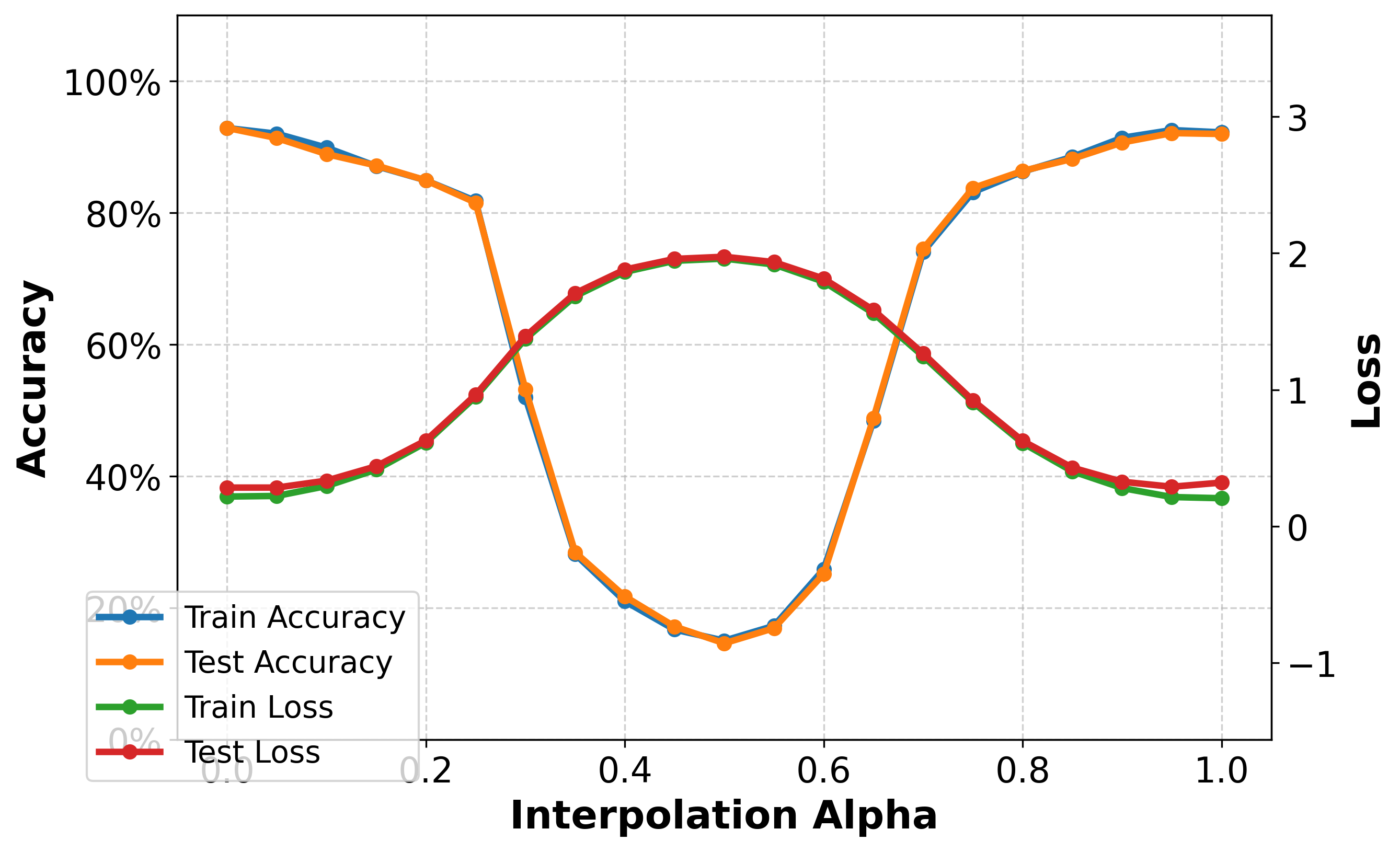}
        \caption{Amazon-Photo}
    \end{subfigure}
    \begin{subfigure}[b]{0.24\textwidth}
        \includegraphics[width=1.0\textwidth]{fig/init/init/coauthor-cs_gcn_plot.png}
        \caption{Coauthor-CS}
    \end{subfigure}
    \begin{subfigure}[b]{0.24\textwidth}
        \includegraphics[width=1.0\textwidth]{fig/init/init/coauthor-physics_gcn_plot.png}
        \caption{Coauthor-Physics}
    \end{subfigure}
    \begin{subfigure}[b]{0.24\textwidth}
        \includegraphics[width=1.0\textwidth]{fig/init/init/wikics_gcn_plot.png}
        \caption{WikiCS}
    \end{subfigure}
    \begin{subfigure}[b]{0.24\textwidth}
        \includegraphics[width=1.0\textwidth]{fig/init/init/squirrel_gcn_plot.png}
        \caption{Squirrel}
    \end{subfigure}
    \begin{subfigure}[b]{0.24\textwidth}
        \includegraphics[width=1.0\textwidth]{fig/init/init/chameleon_gcn_plot.png}
        \caption{Chameleon}
    \end{subfigure}
    \begin{subfigure}[b]{0.24\textwidth}
        \includegraphics[width=1.0\textwidth]{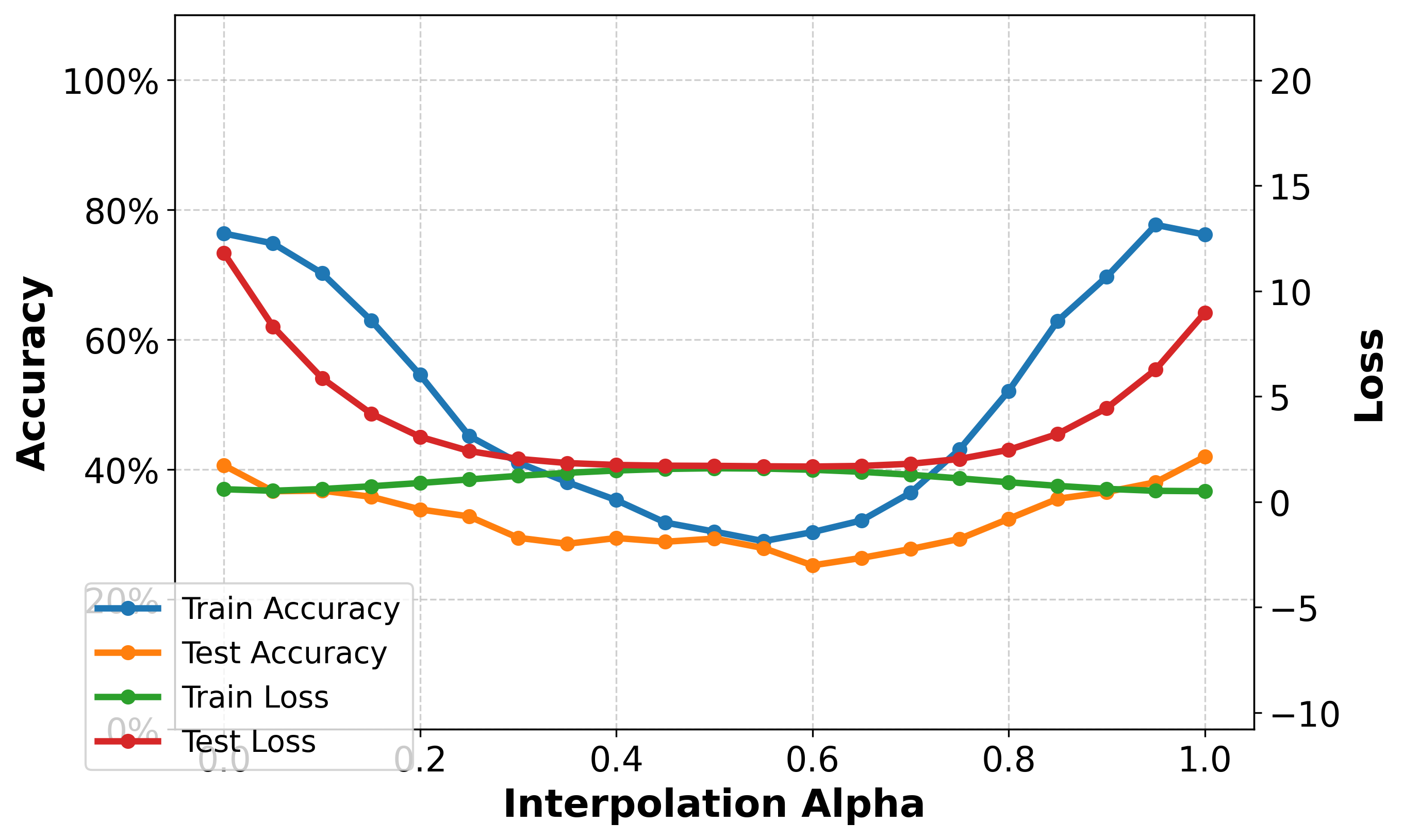}
        \caption{Roman-Empire}
    \end{subfigure}
    \begin{subfigure}[b]{0.24\textwidth}
        \includegraphics[width=1.0\textwidth]{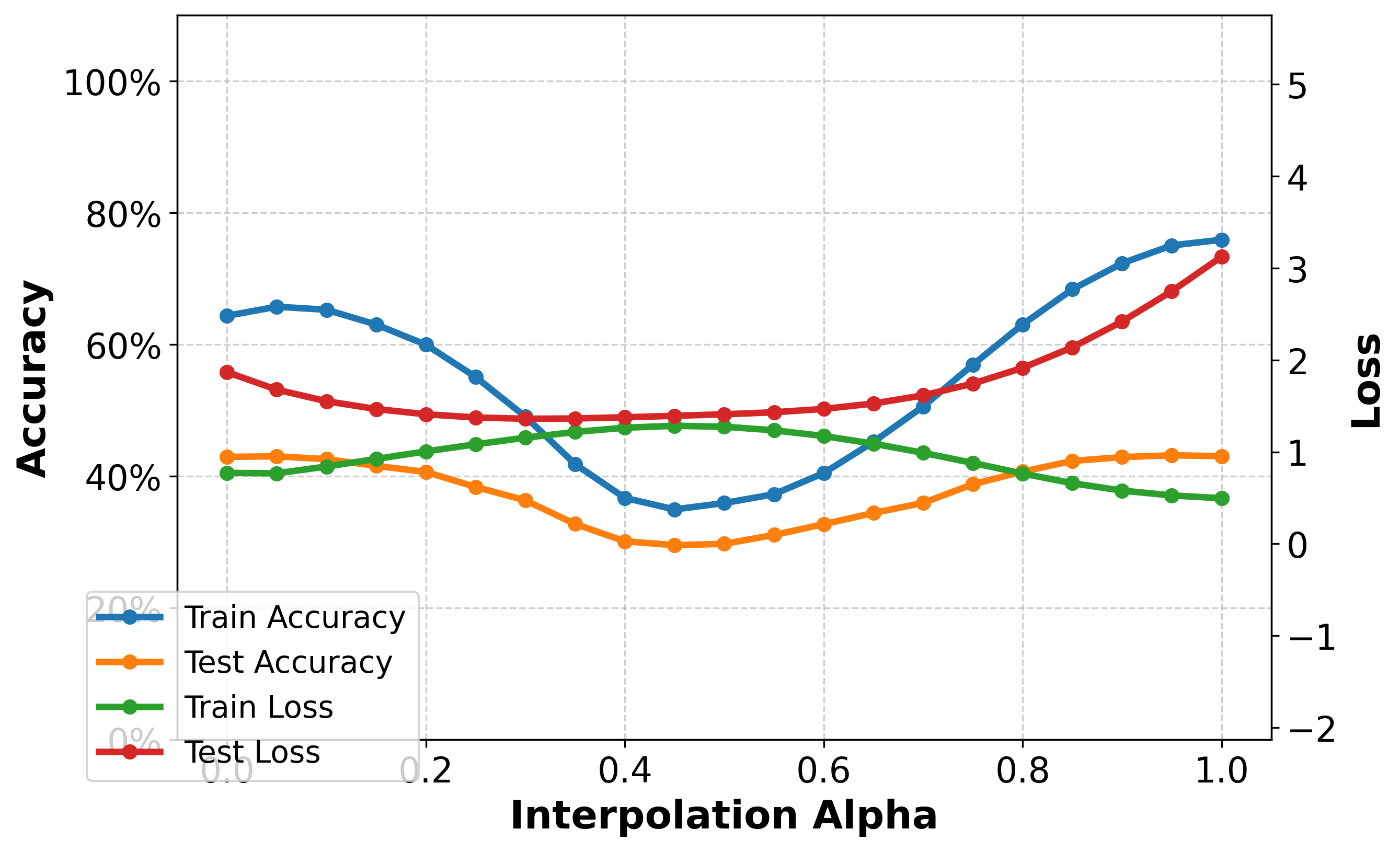}
        \caption{Amazon-Ratings }
    \end{subfigure}
        \begin{subfigure}[b]{0.24\textwidth}
        \includegraphics[width=1.0\textwidth]{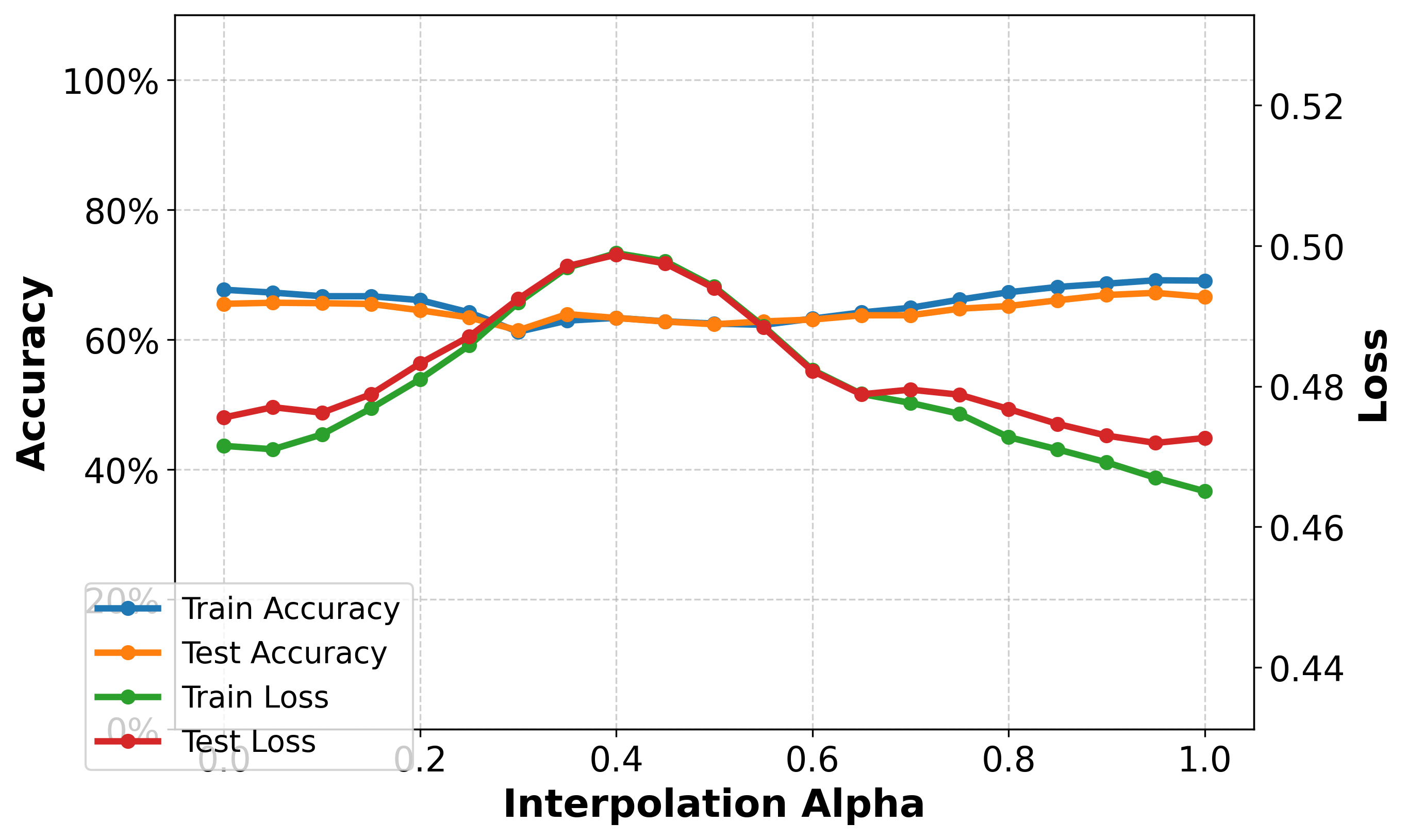}
        \caption{Minesweeper}
    \end{subfigure}
    \caption{The performance of interpolations along a non-linear path connecting two minima.}
    \label{fig:A1}
\end{figure*}

\clearpage

\subsection{ The performance of interpolations along quadratic Bézier curve connecting two minima}
\begin{figure*}[!ht]
    \centering
    \begin{subfigure}[b]{0.24\textwidth}
        \includegraphics[width=1.0\textwidth]{fig/Bezier/Bezier/cora_bezier_fitted_plot.png}
        \caption{Cora}
    \end{subfigure}
    \begin{subfigure}[b]{0.24\textwidth}
        \includegraphics[width=1.0\textwidth]{fig/Bezier/Bezier/citeseer_bezier_fitted_plot.png}
        \caption{CiteSeer}
    \end{subfigure}
    \begin{subfigure}[b]{0.24\textwidth}
        \includegraphics[width=1.0\textwidth]{fig/Bezier/Bezier/pubmed_bezier_fitted_plot.png}
        \caption{PubMed}
    \end{subfigure}
    \begin{subfigure}[b]{0.24\textwidth}
        \includegraphics[width=1.0\textwidth]{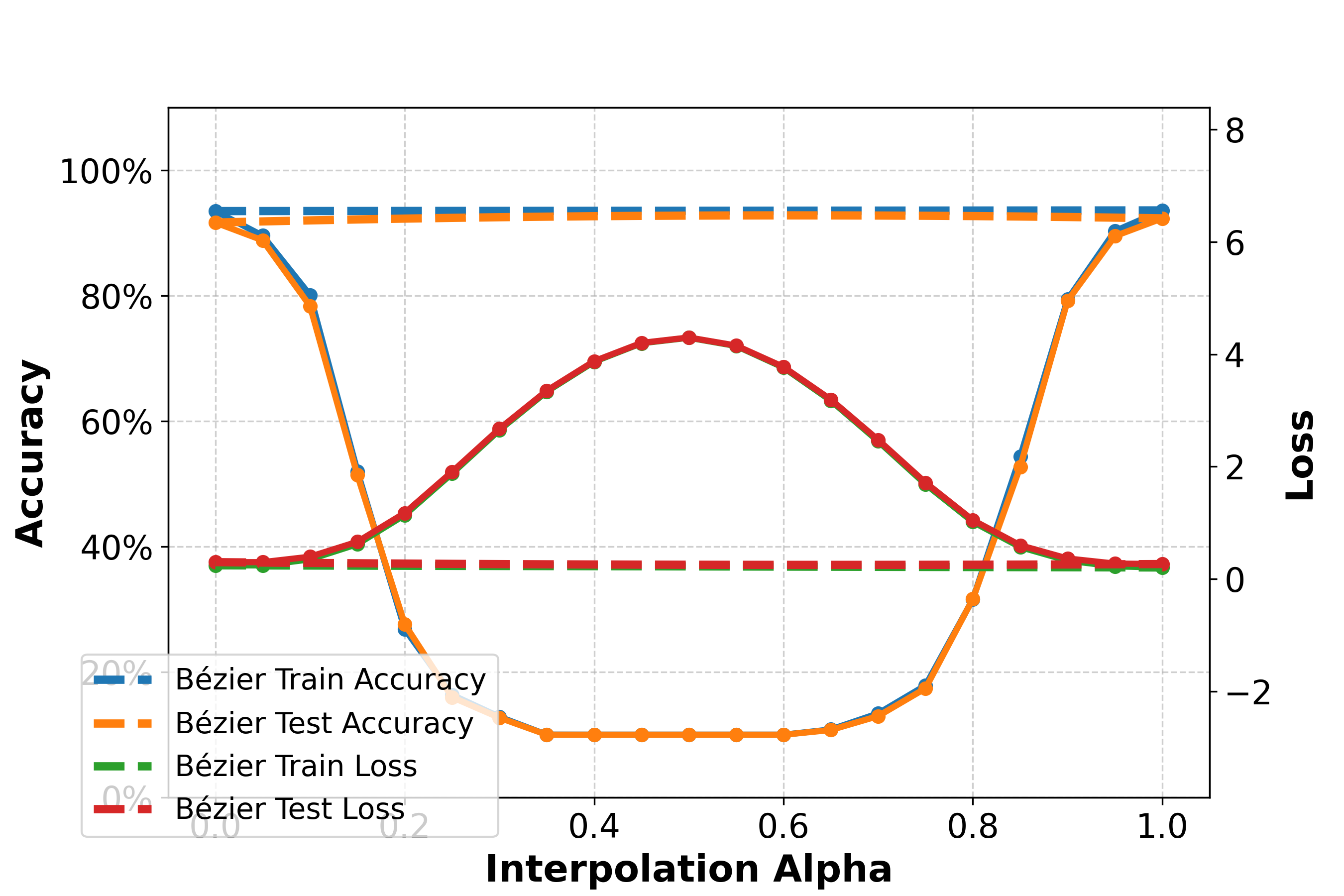}
        \caption{Amazon-Computer}
    \end{subfigure}
    \begin{subfigure}[b]{0.24\textwidth}
        \includegraphics[width=1.0\textwidth]{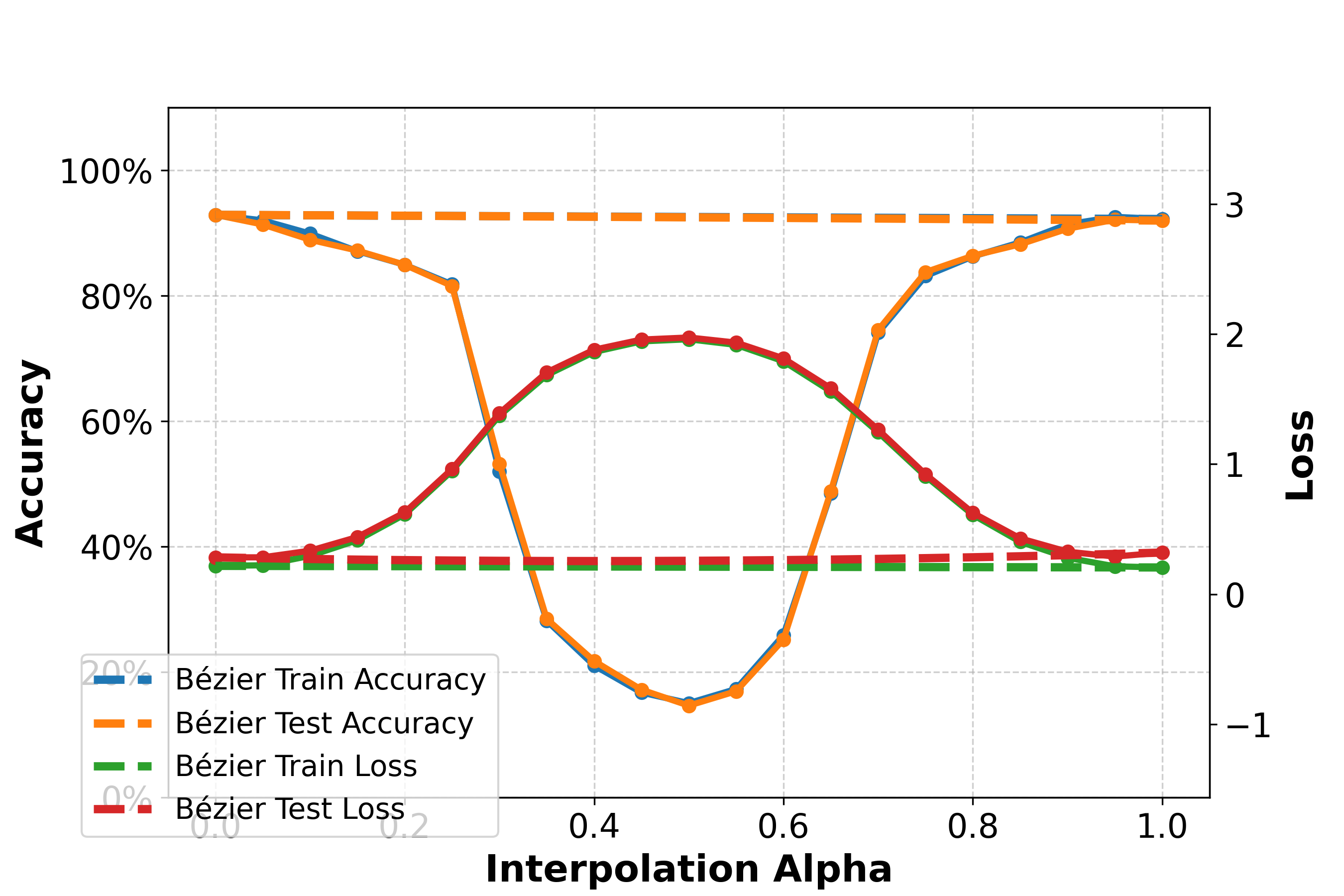}
        \caption{Amazon-Photo}
    \end{subfigure}
    \begin{subfigure}[b]{0.24\textwidth}
        \includegraphics[width=1.0\textwidth]{fig/Bezier/Bezier/coauthor-cs_bezier_fitted_plot.png}
        \caption{Coauthor-CS}
    \end{subfigure}
    \begin{subfigure}[b]{0.24\textwidth}
        \includegraphics[width=1.0\textwidth]{fig/Bezier/Bezier/coauthor-physics_bezier_fitted_plot.png}
        \caption{Coauthor-Physics}
    \end{subfigure}
    \begin{subfigure}[b]{0.24\textwidth}
        \includegraphics[width=1.0\textwidth]{fig/Bezier/Bezier/wikics_bezier_fitted_plot.png}
        \caption{WikiCS}
    \end{subfigure}
    \begin{subfigure}[b]{0.24\textwidth}
        \includegraphics[width=1.0\textwidth]{fig/Bezier/Bezier/squirrel_bezier_fitted_plot.png}
        \caption{Squirrel}
    \end{subfigure}
    \begin{subfigure}[b]{0.24\textwidth}
        \includegraphics[width=1.0\textwidth]{fig/Bezier/Bezier/chameleon_bezier_fitted_plot.png}
        \caption{Chameleon}
    \end{subfigure}
    \begin{subfigure}[b]{0.24\textwidth}
        \includegraphics[width=1.0\textwidth]{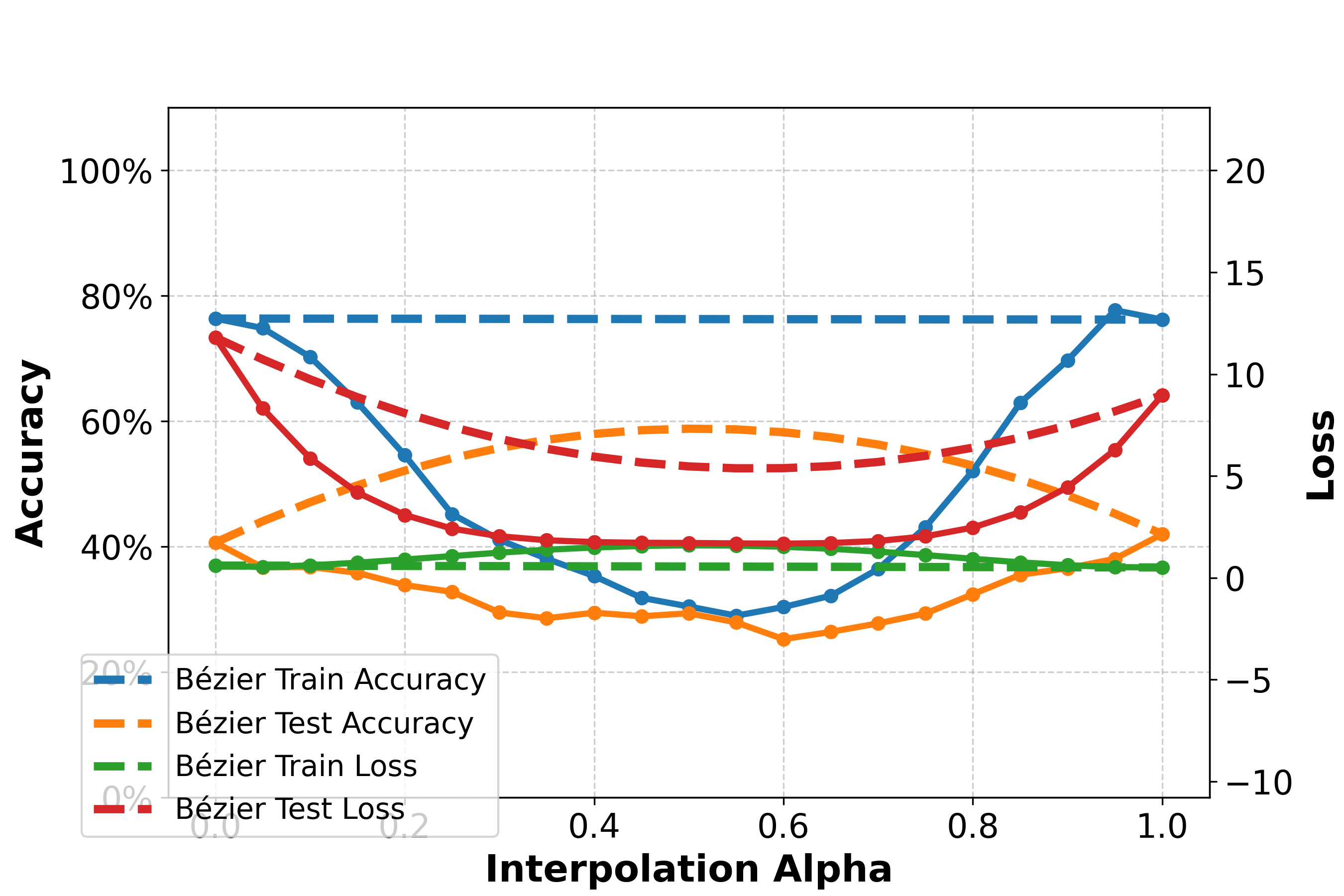}
        \caption{Roman-Empire}
    \end{subfigure}
    \begin{subfigure}[b]{0.24\textwidth}
        \includegraphics[width=1.0\textwidth]{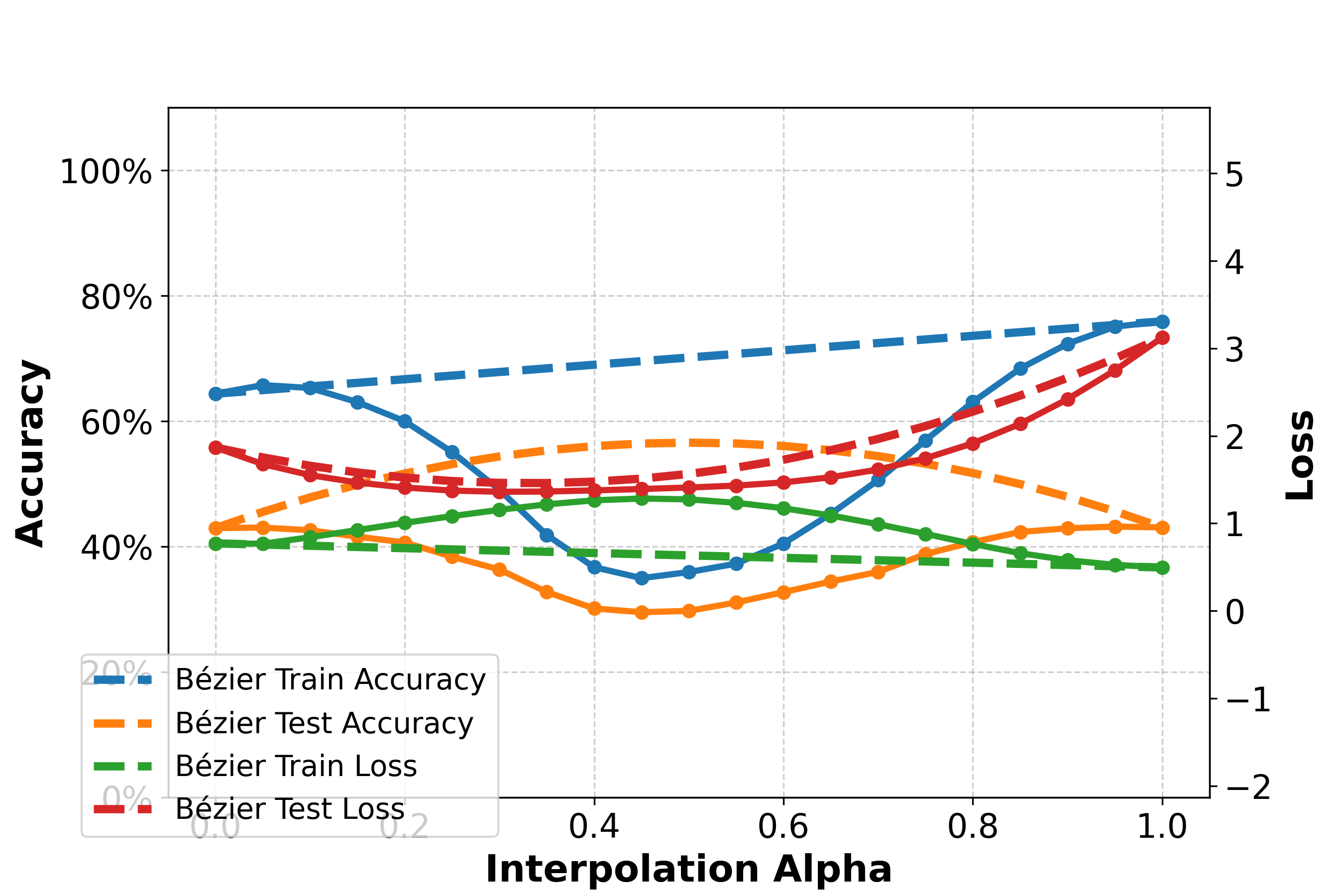}
        \caption{Amazon-Ratings }
    \end{subfigure}
        \begin{subfigure}[b]{0.24\textwidth}
        \includegraphics[width=1.0\textwidth]{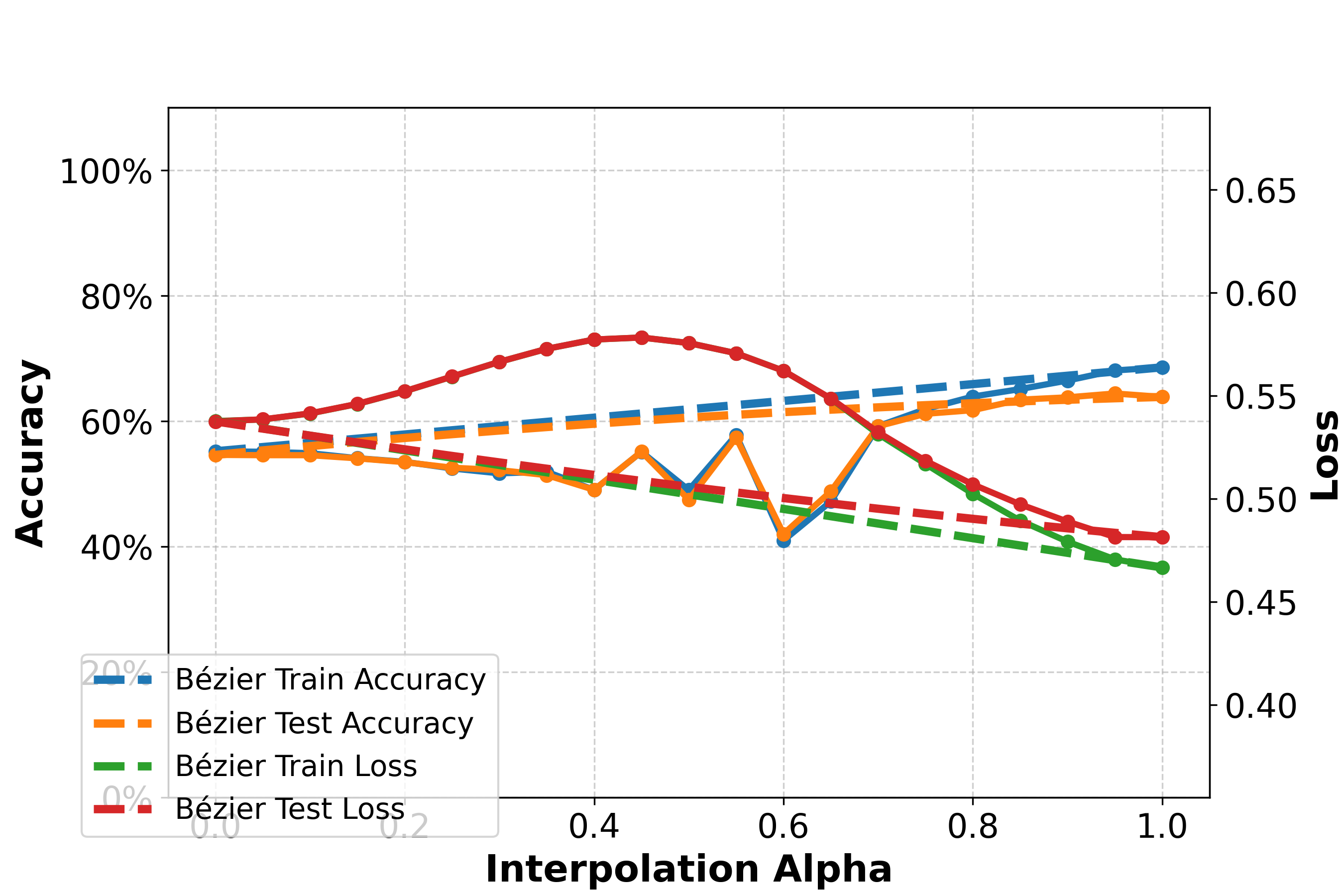}
        \caption{Minesweeper}
    \end{subfigure}
    \caption{The performance of interpolations along a non-linear path connecting two minima.}
    \label{fig:A2}
\end{figure*}
\clearpage

\subsection{Effect of convolution mechanism on mode connectivity}
\label{app: conv}
\begin{figure*}[!ht]
    \centering
    \begin{subfigure}[b]{0.24\textwidth}
        \includegraphics[width=1.0\textwidth]{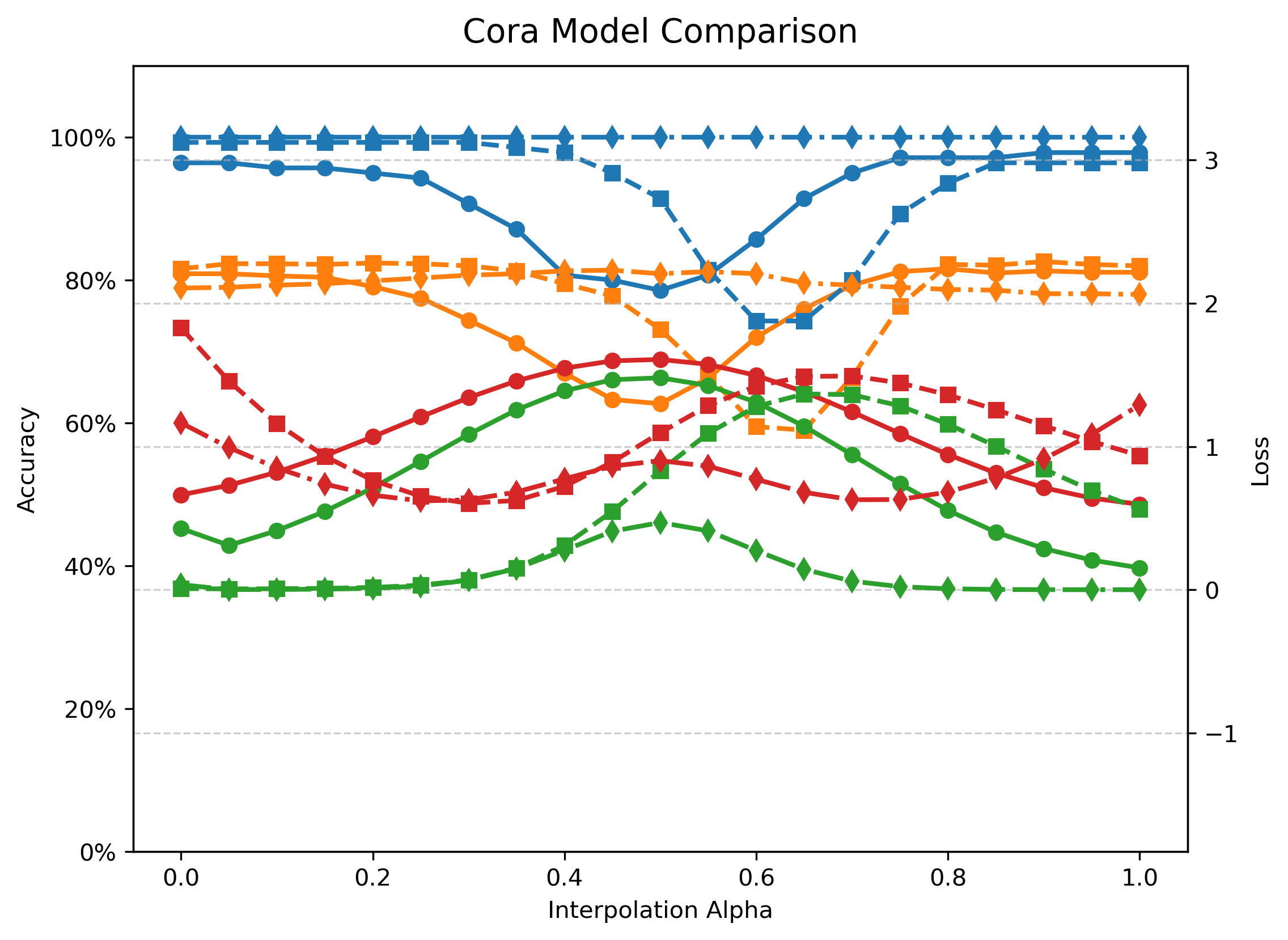}
        \caption{Cora}
    \end{subfigure}
    \begin{subfigure}[b]{0.24\textwidth}
        \includegraphics[width=1.0\textwidth]{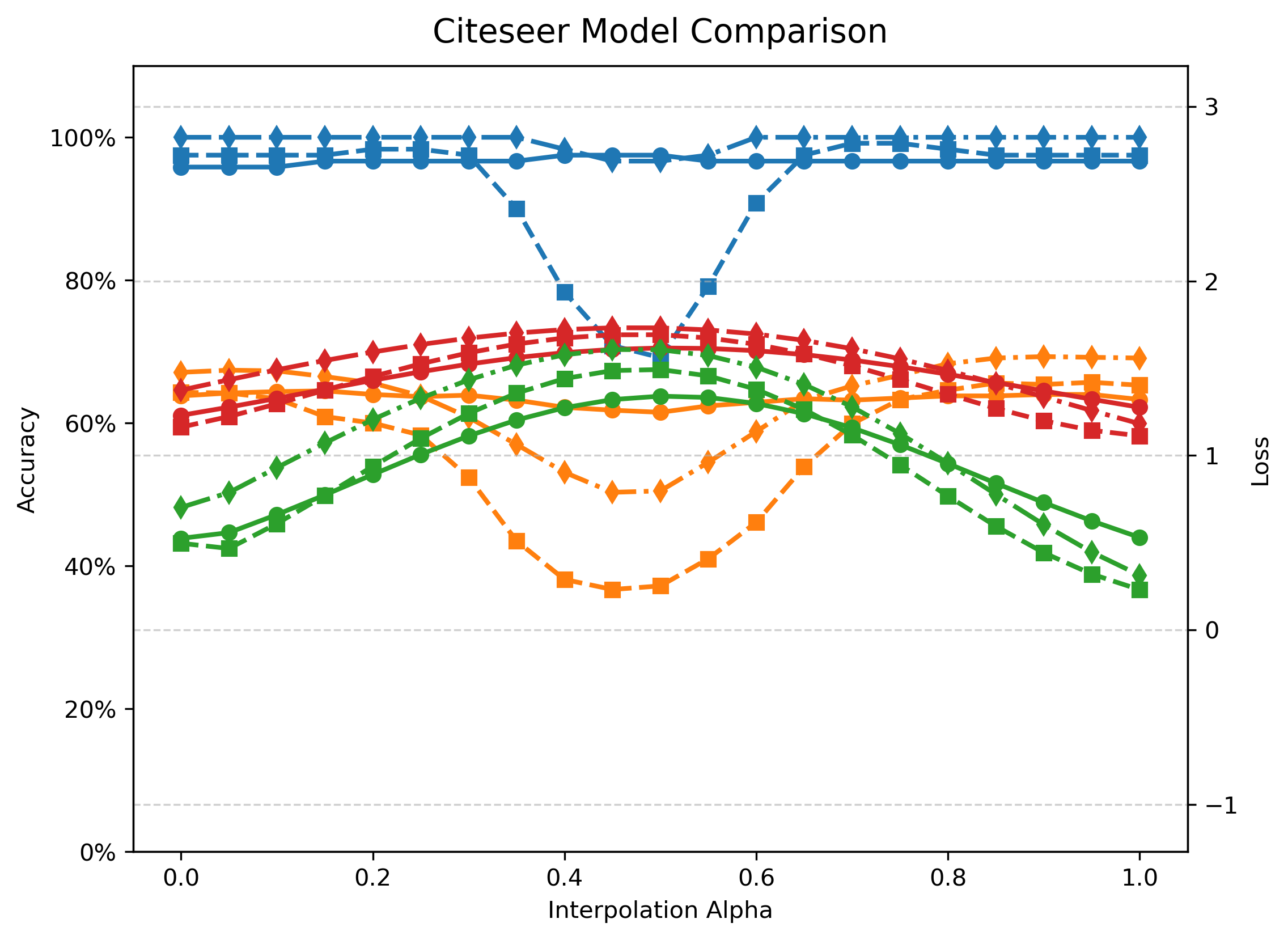}
        \caption{CiteSeer}
    \end{subfigure}
    \begin{subfigure}[b]{0.24\textwidth}
        \includegraphics[width=1.0\textwidth]{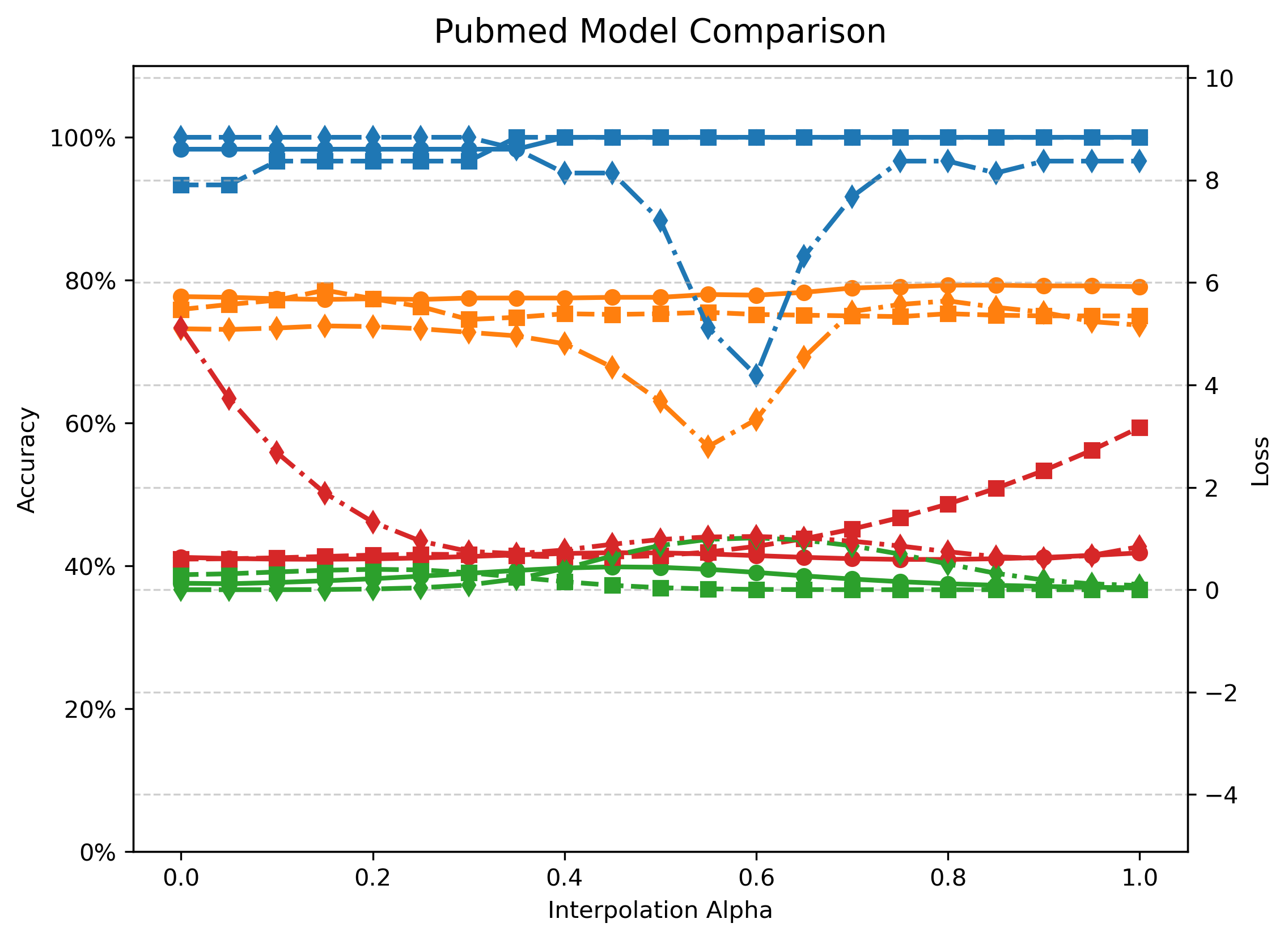}
        \caption{PubMed}
    \end{subfigure}
    \begin{subfigure}[b]{0.24\textwidth}
        \includegraphics[width=1.0\textwidth]{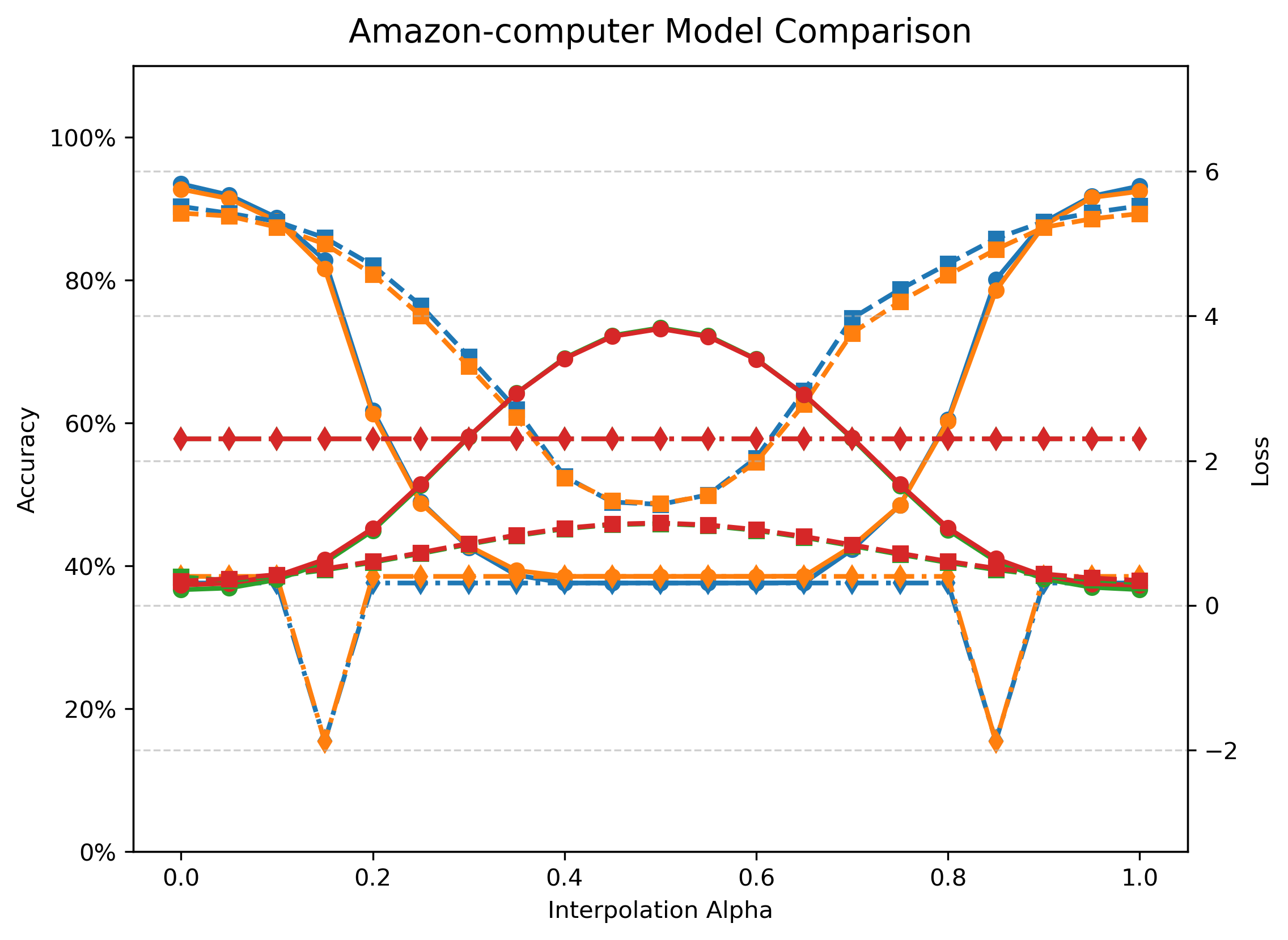}
        \caption{Amazon-Computer}
    \end{subfigure}
    \begin{subfigure}[b]{0.24\textwidth}
        \includegraphics[width=1.0\textwidth]{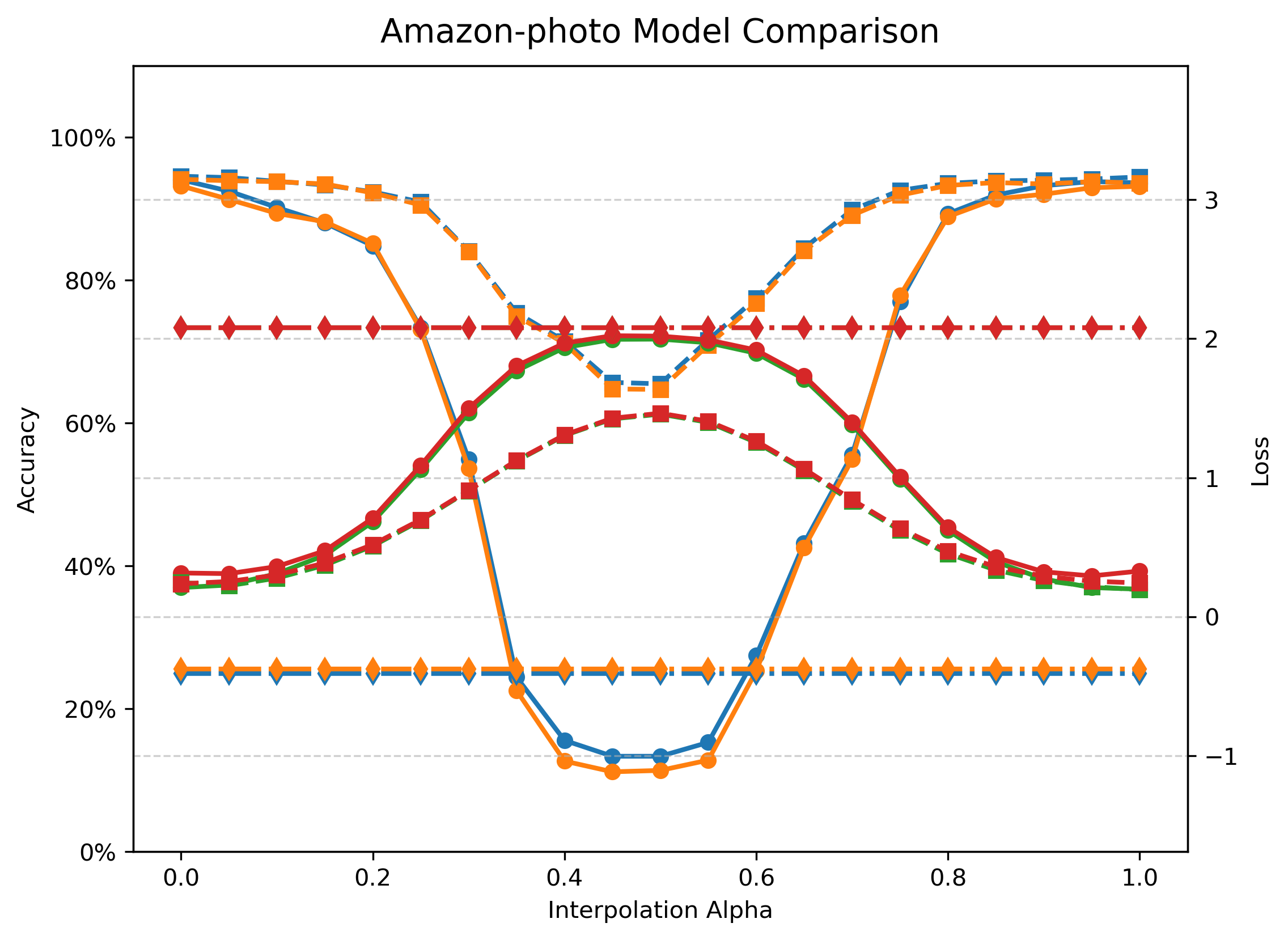}
        \caption{Amazon-Photo}
    \end{subfigure}
    \begin{subfigure}[b]{0.24\textwidth}
        \includegraphics[width=1.0\textwidth]{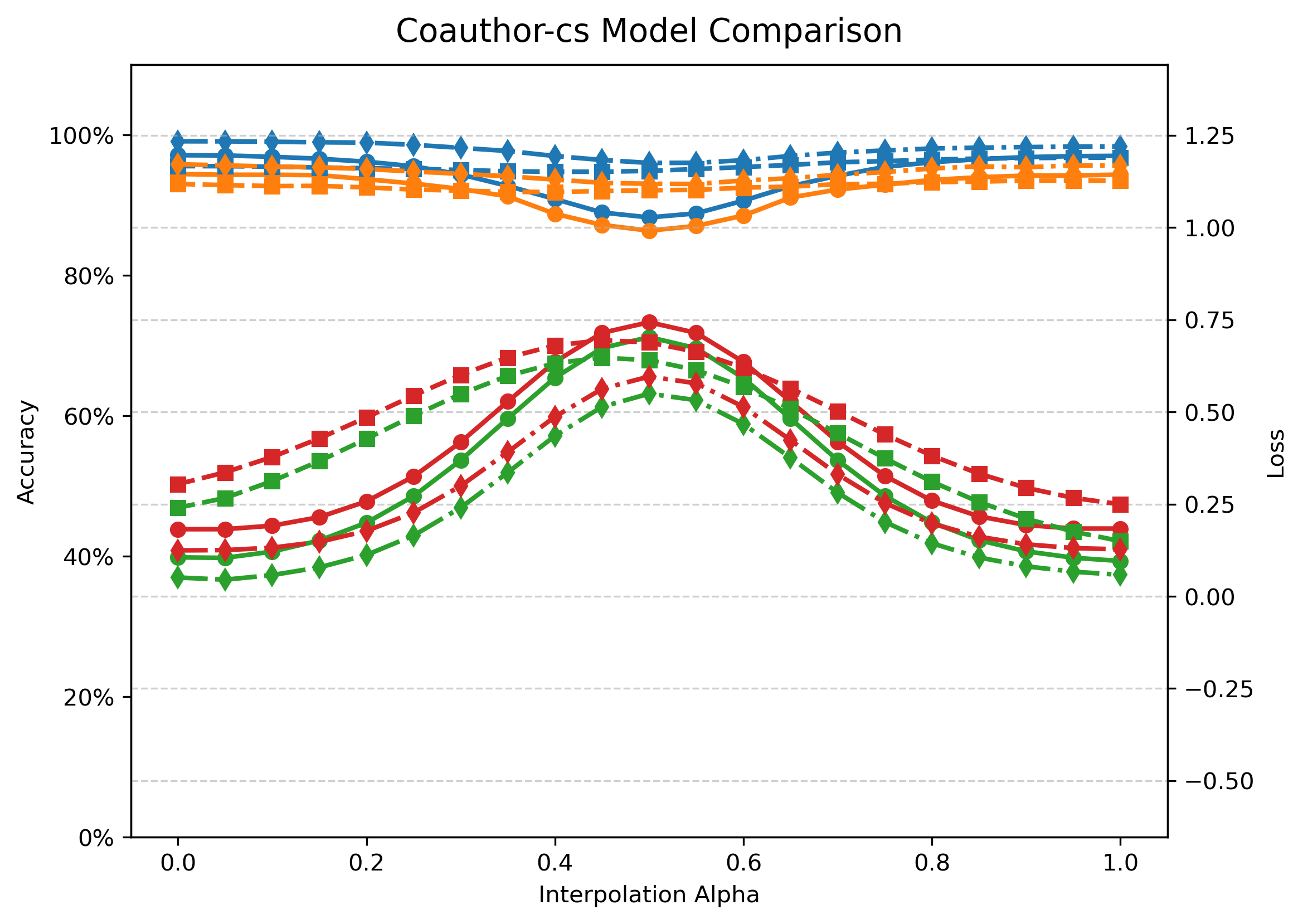}
        \caption{Coauthor-CS}
    \end{subfigure}
    \begin{subfigure}[b]{0.24\textwidth}
        \includegraphics[width=1.0\textwidth]{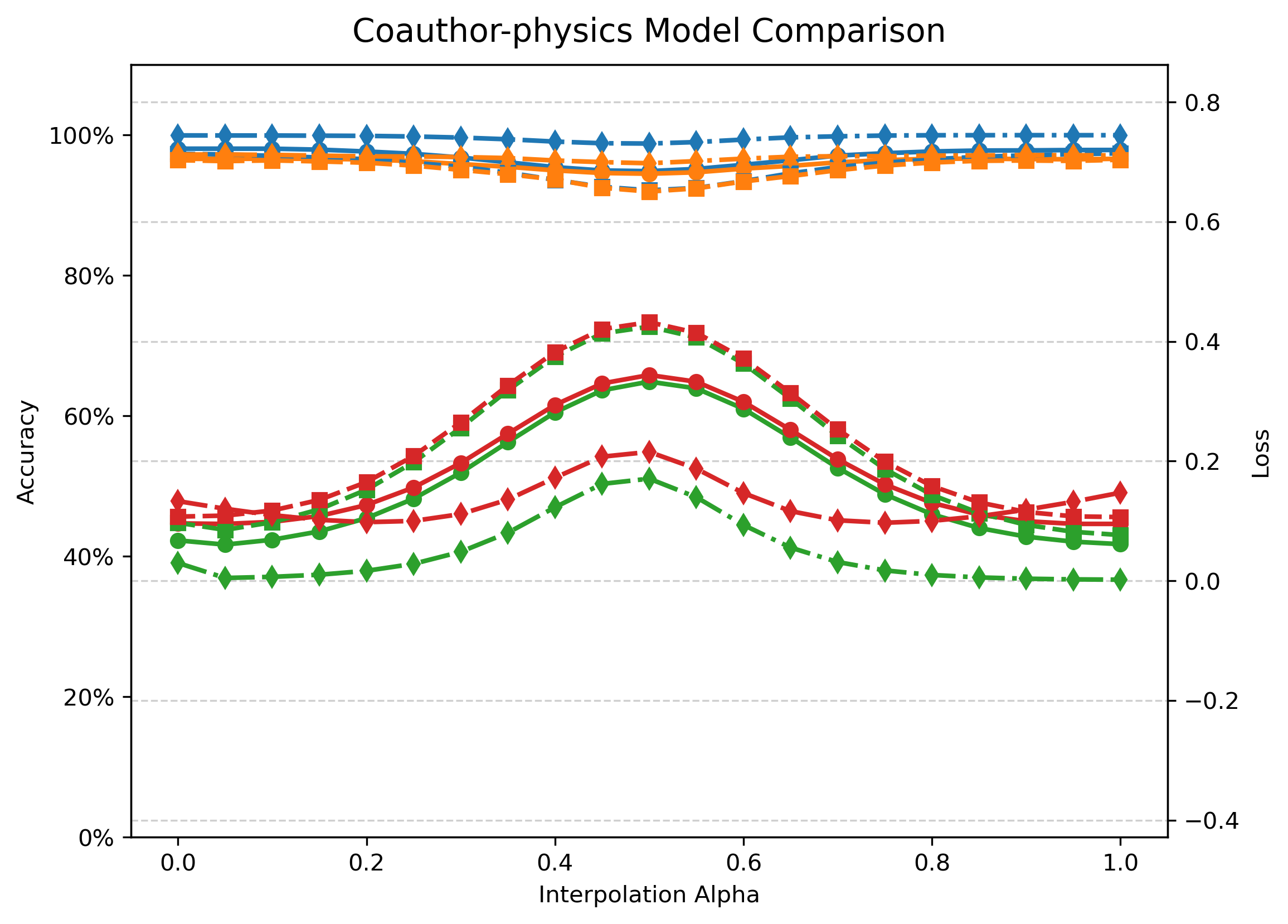}
        \caption{Coauthor-Physics}
    \end{subfigure}
    \begin{subfigure}[b]{0.24\textwidth}
        \includegraphics[width=1.0\textwidth]{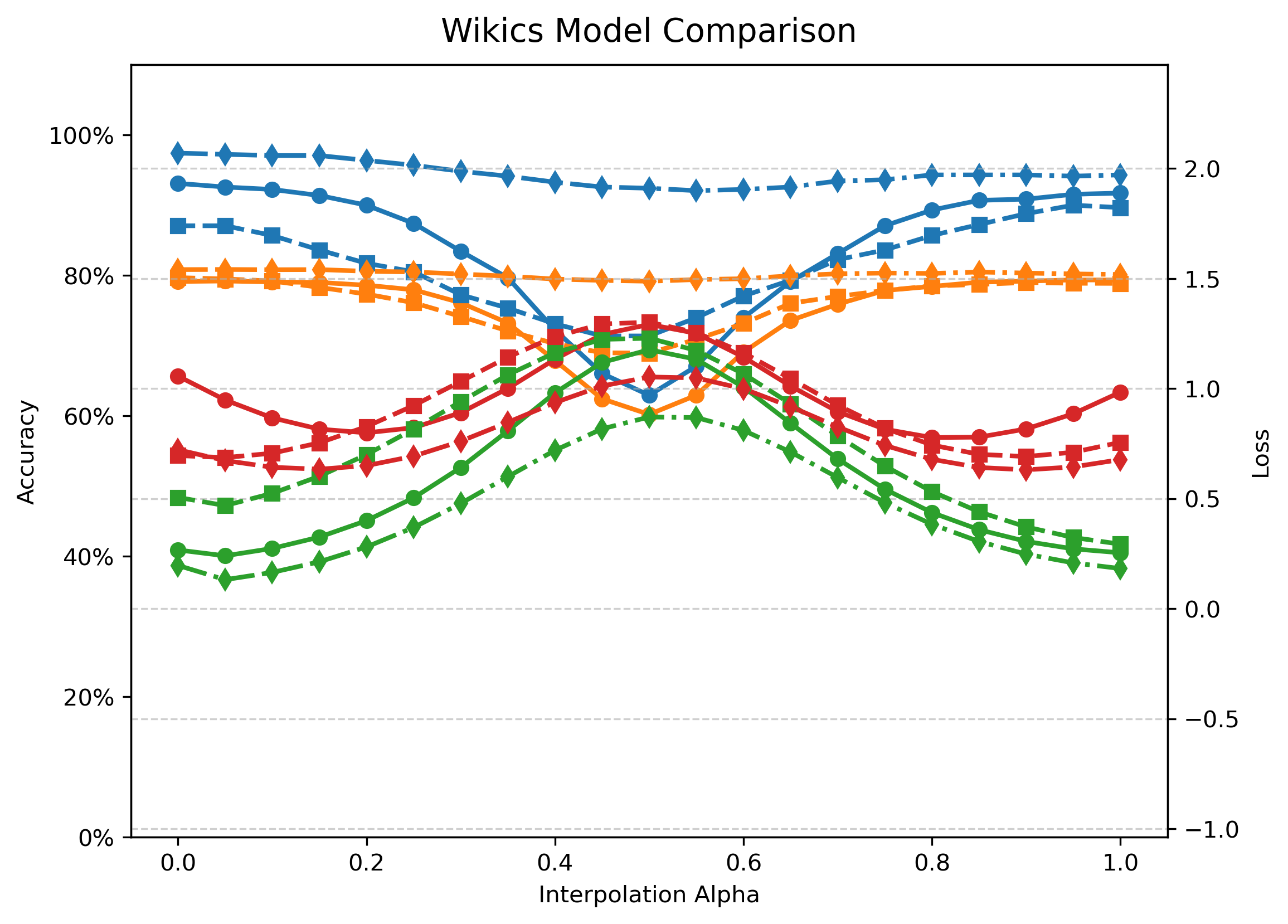}
        \caption{WikiCS}
    \end{subfigure}
    \begin{subfigure}[b]{0.24\textwidth}
        \includegraphics[width=1.0\textwidth]{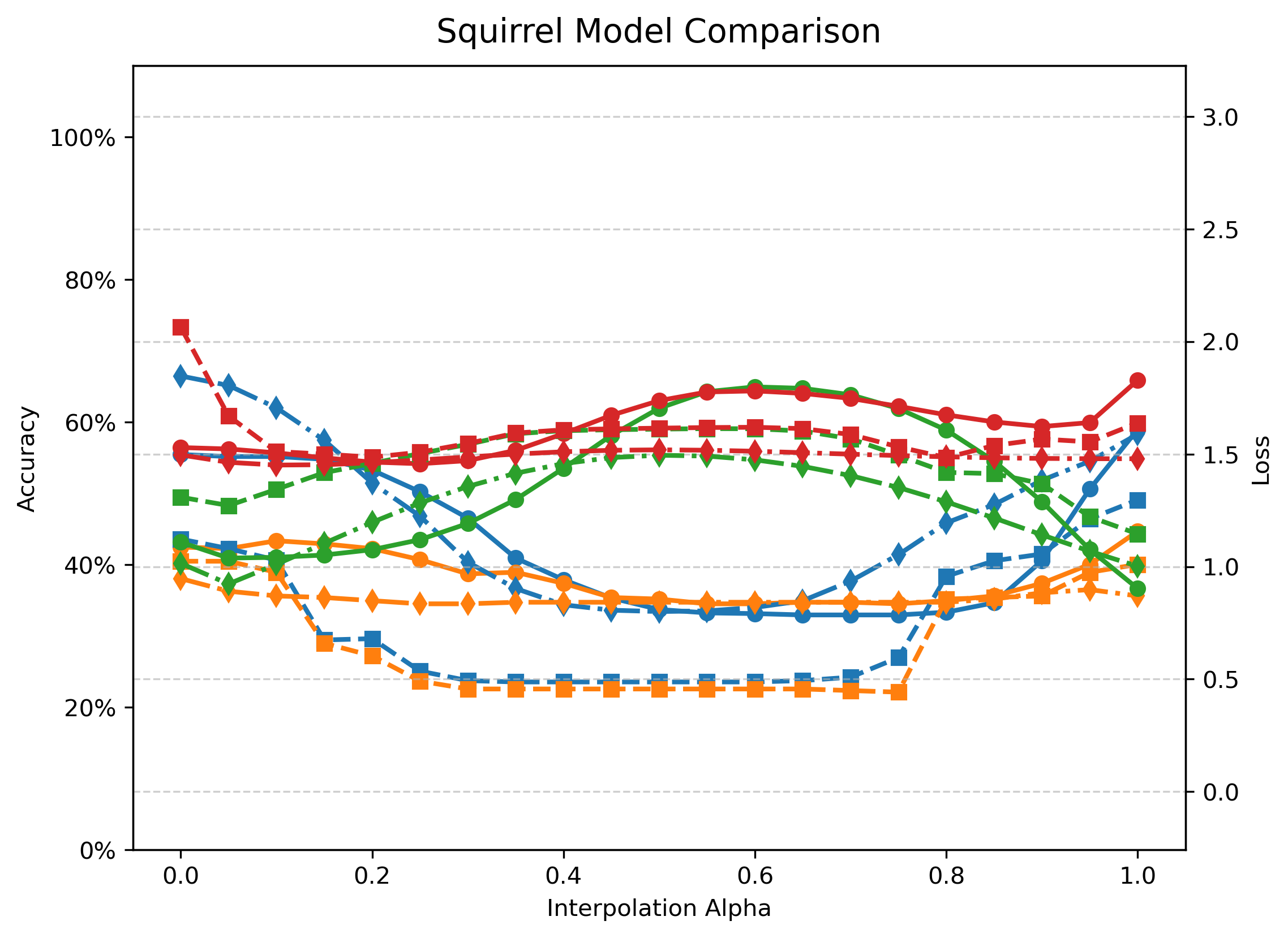}
        \caption{Squirrel}
    \end{subfigure}
    \begin{subfigure}[b]{0.24\textwidth}
        \includegraphics[width=1.0\textwidth]{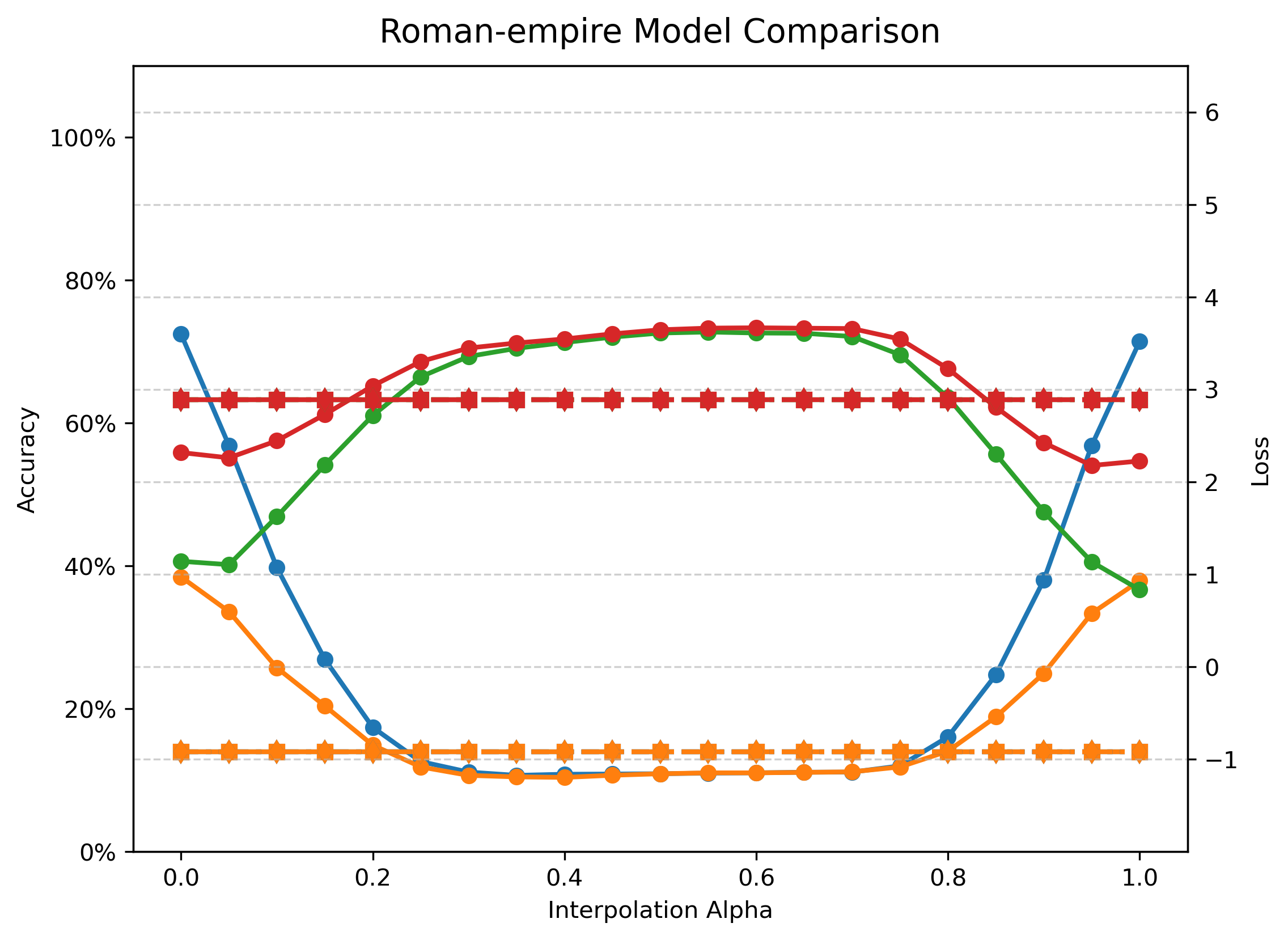}
        \caption{Chameleon}
    \end{subfigure}
    \begin{subfigure}[b]{0.24\textwidth}
        \includegraphics[width=1.0\textwidth]{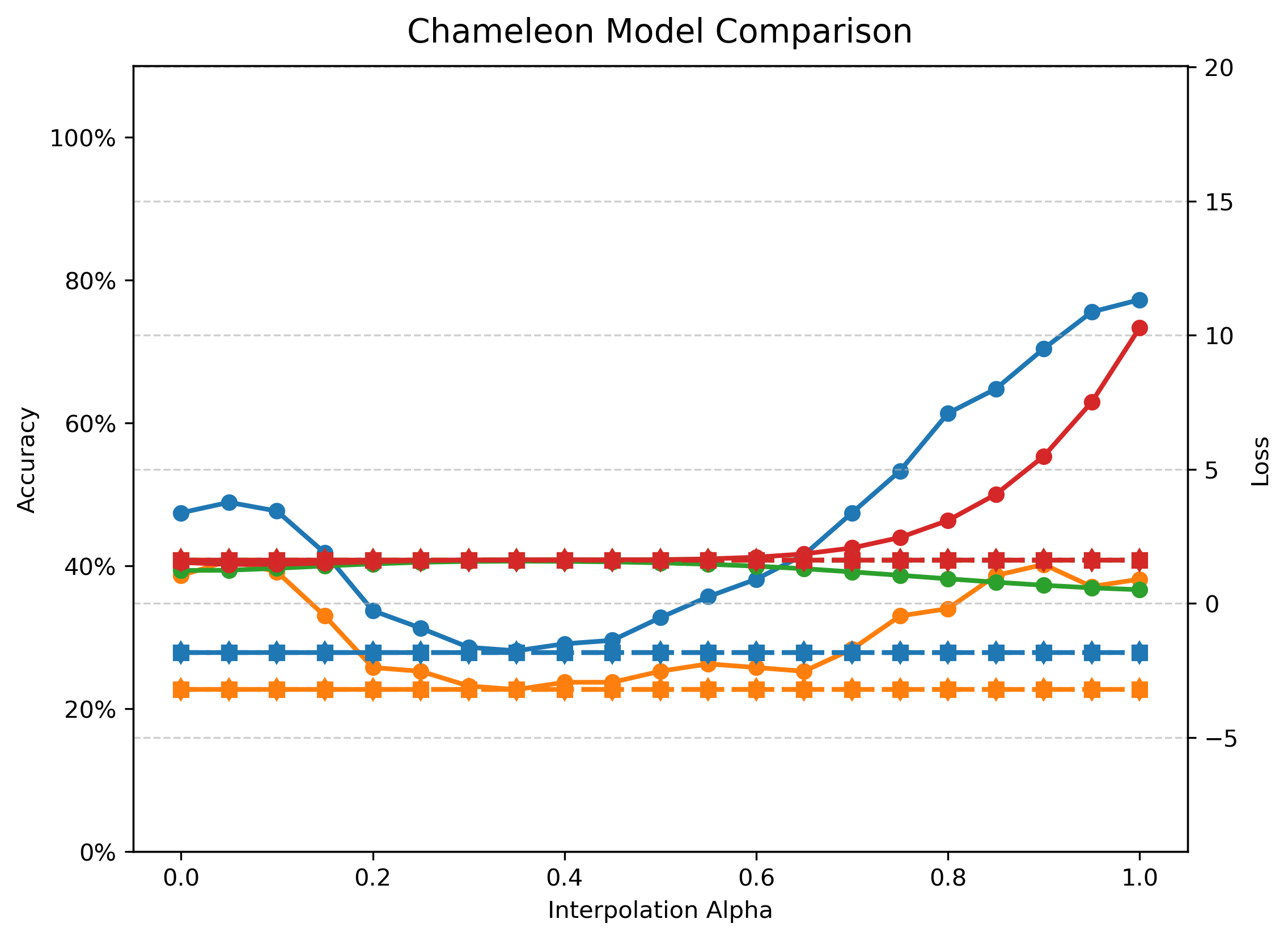}
        \caption{Roman-Empire}
    \end{subfigure}
    \begin{subfigure}[b]{0.24\textwidth}
        \includegraphics[width=1.0\textwidth]{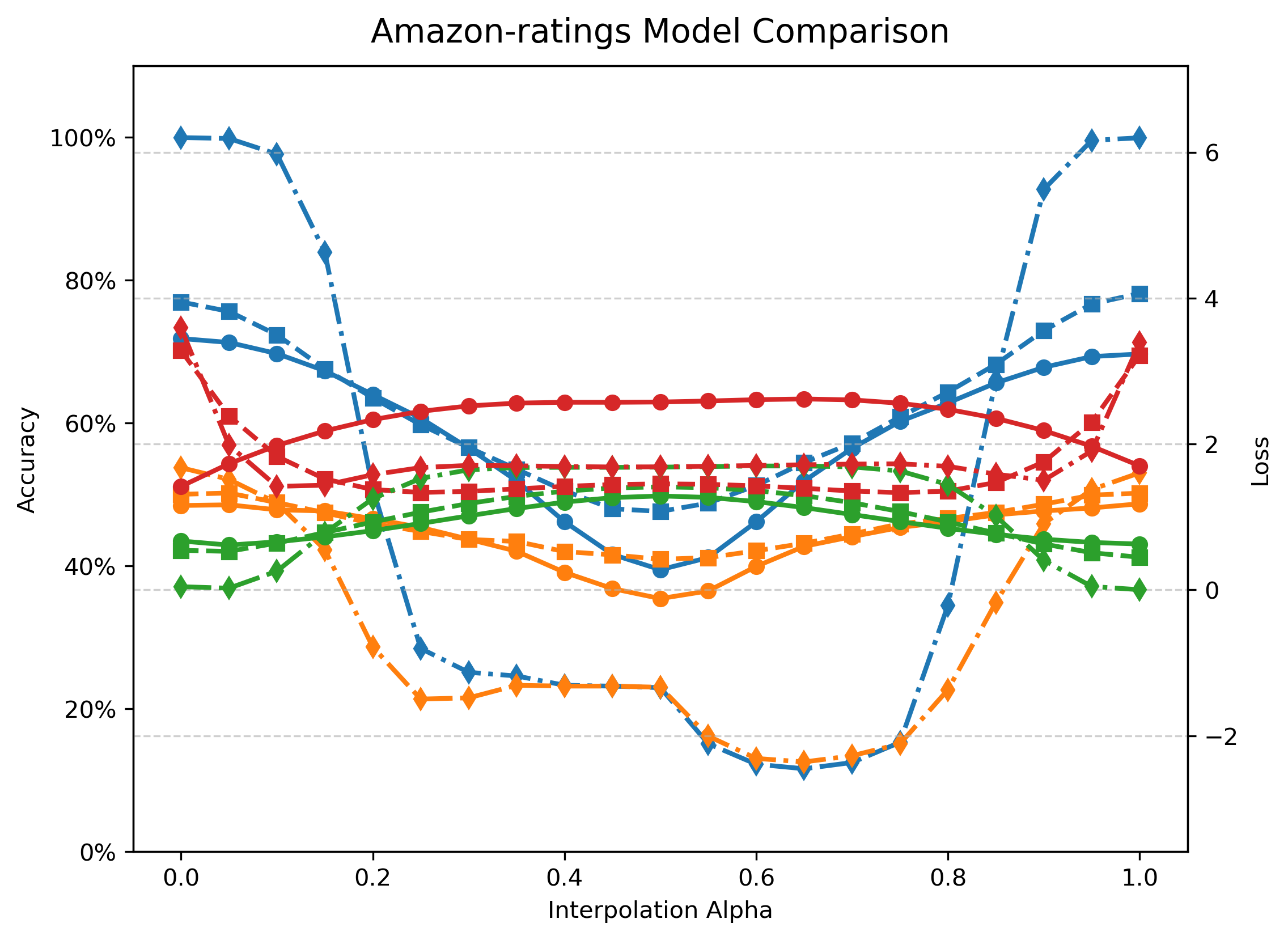}
        \caption{Amazon-Ratings }
    \end{subfigure}
        \begin{subfigure}[b]{0.24\textwidth}
        \includegraphics[width=1.0\textwidth]{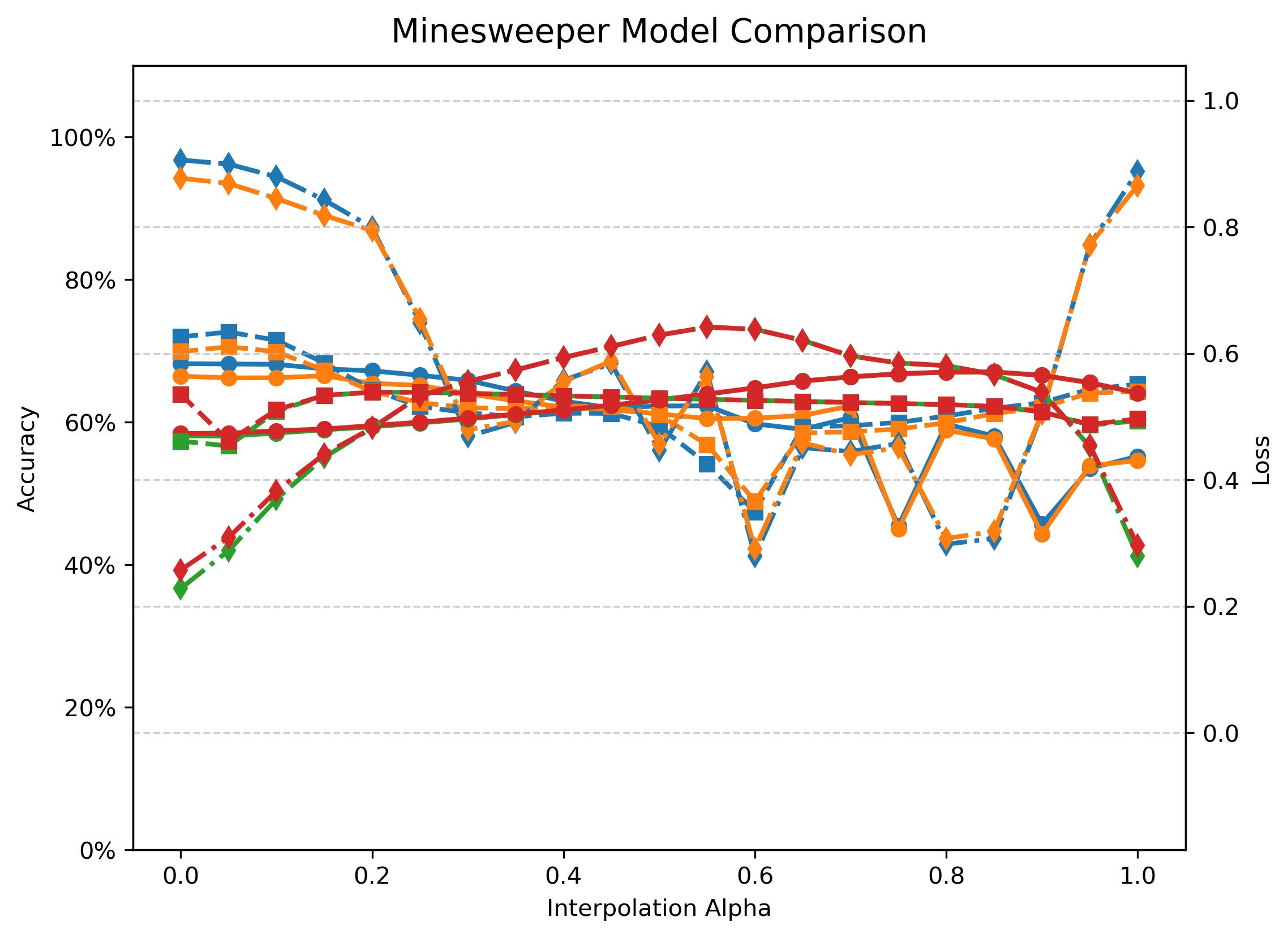}
        \caption{Minesweeper}
    \end{subfigure}
            \begin{subfigure}[b]{0.24\textwidth}
        \includegraphics[width=1.0\textwidth]{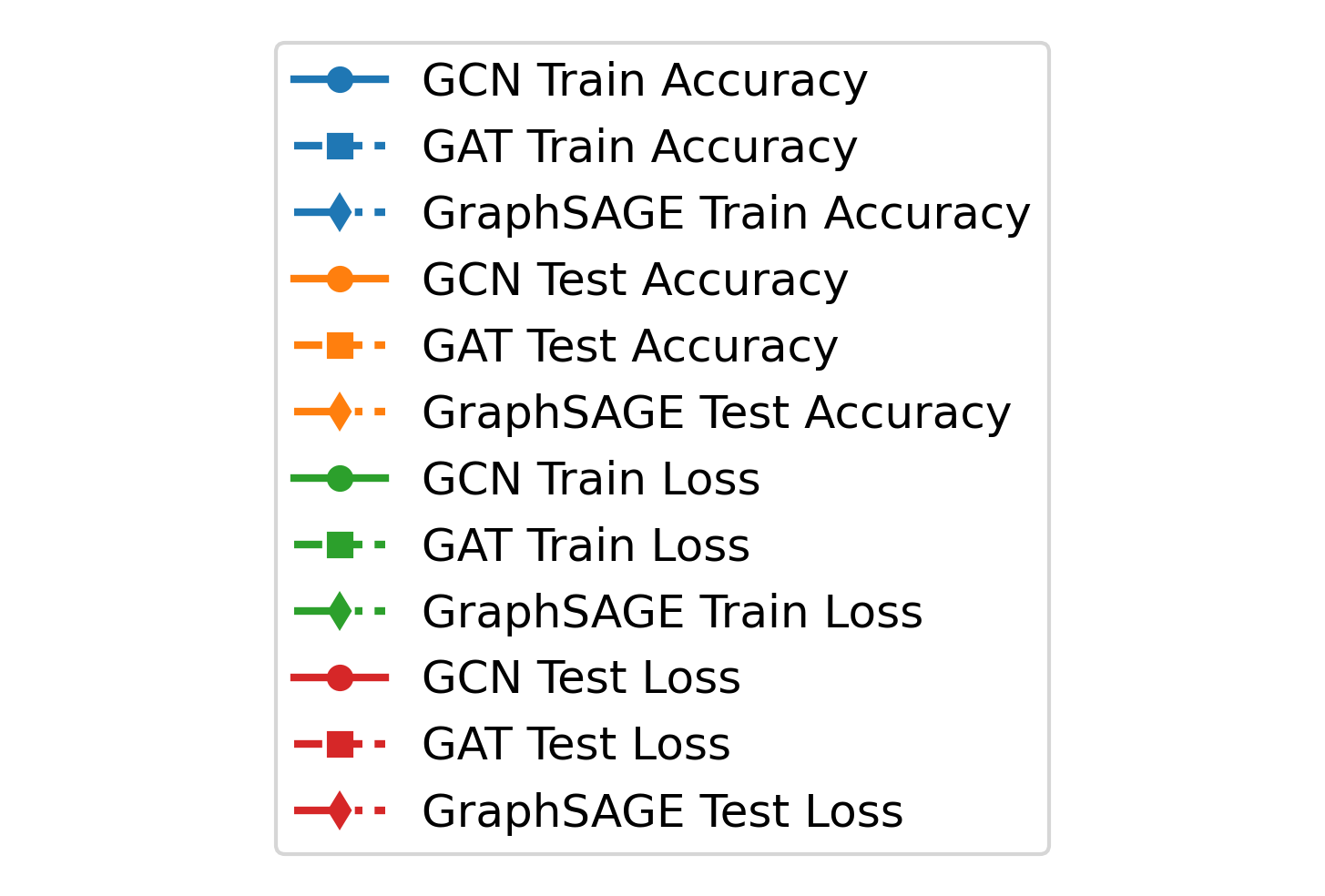}
        \caption{Legend}
    \end{subfigure}
    \caption{Performance of mode connectivity on different convolution mechanisms.}
    \label{fig:A3}
\end{figure*}

\clearpage
\subsection{Visualization of Loss basin and minimas }
\begin{figure*}[!ht]
    \centering
    \begin{subfigure}[b]{0.24\textwidth}
        \includegraphics[width=1.0\textwidth]{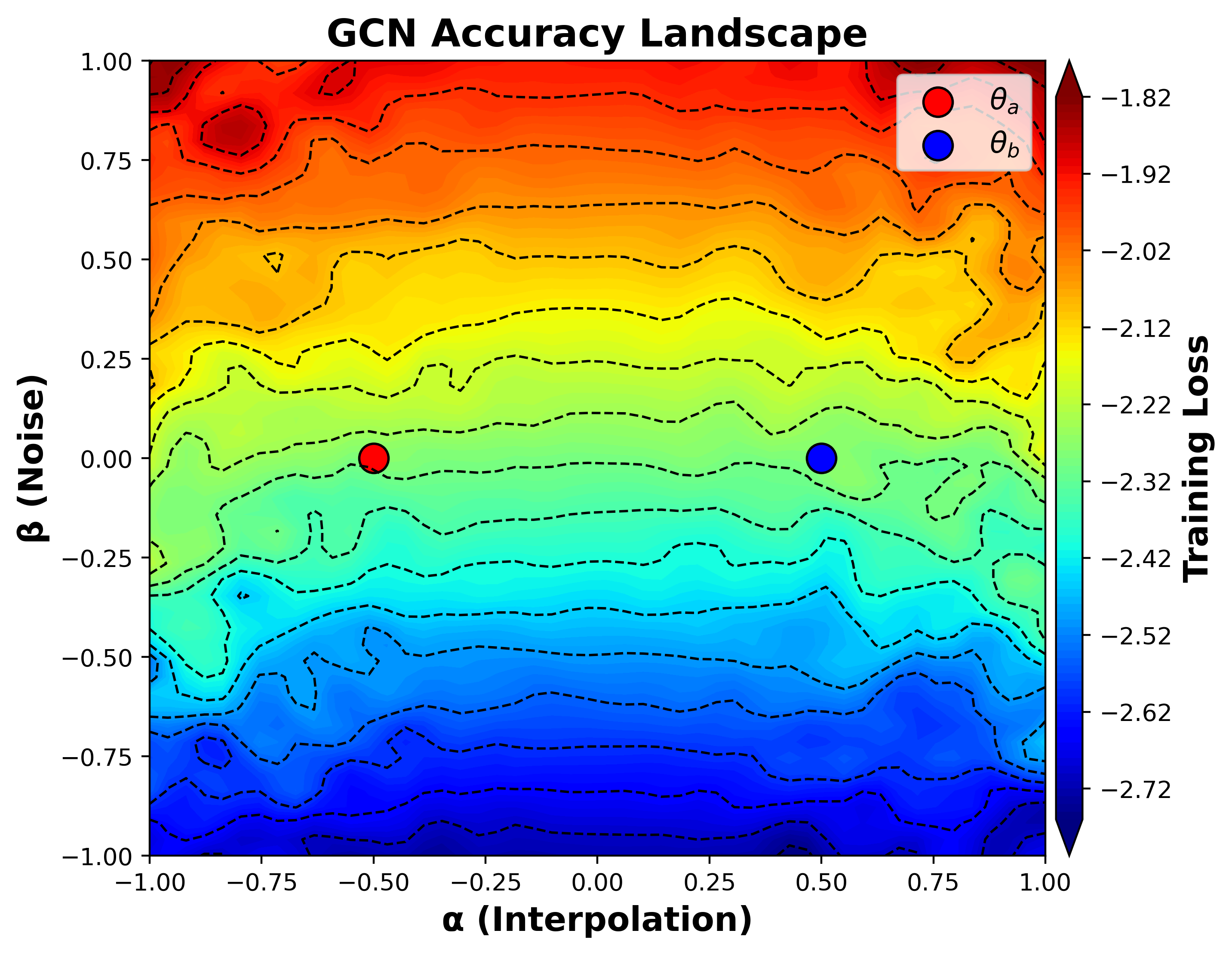}
        \caption{Cora}
    \end{subfigure}
    \begin{subfigure}[b]{0.24\textwidth}
        \includegraphics[width=1.0\textwidth]{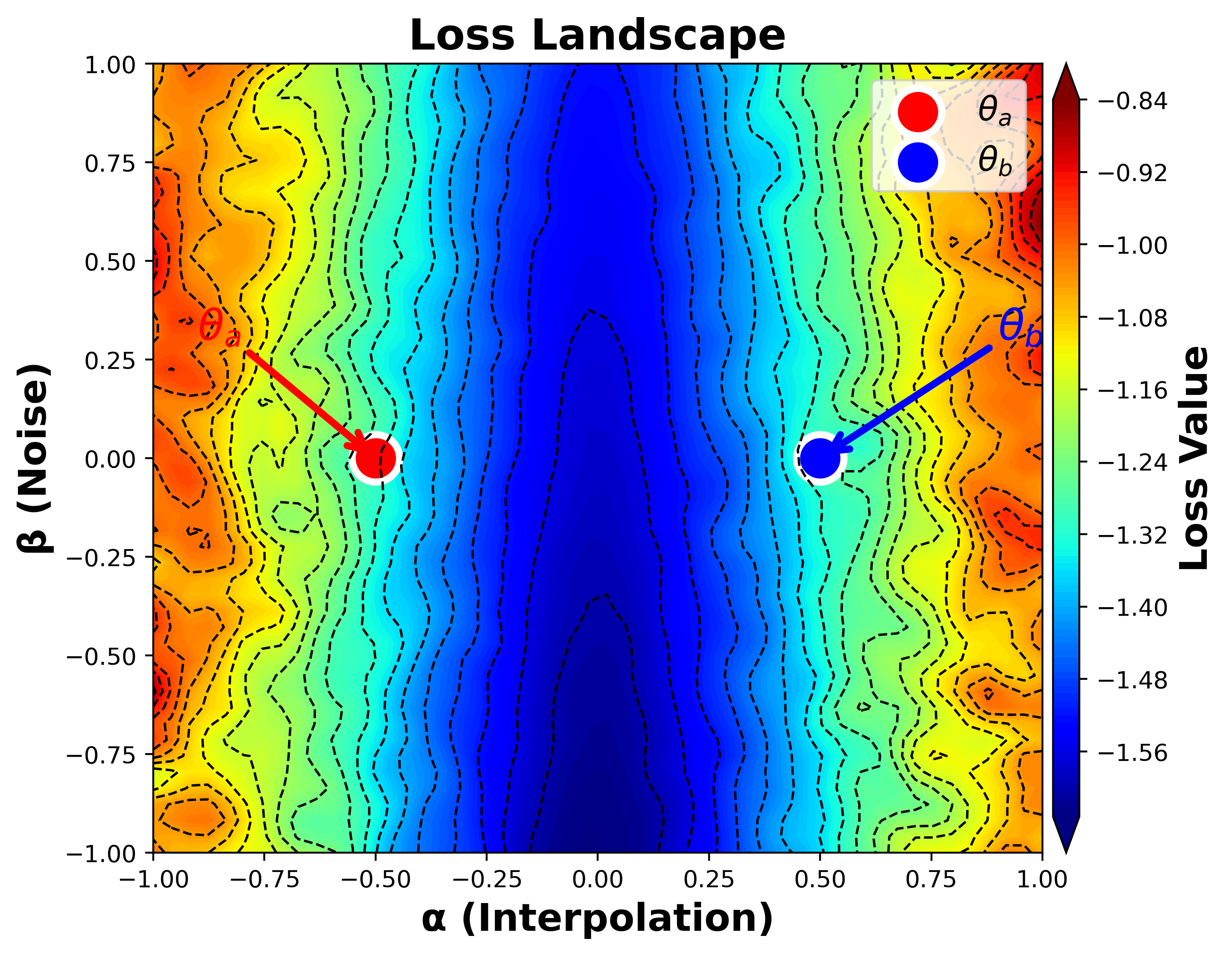}
        \caption{CiteSeer}
    \end{subfigure}
    \begin{subfigure}[b]{0.24\textwidth}
        \includegraphics[width=1.0\textwidth]{fig/basin/Basin/loss_landscape_fixed_pubmed.png}
        \caption{PubMed}
    \end{subfigure}
    \begin{subfigure}[b]{0.24\textwidth}
        \includegraphics[width=1.0\textwidth]{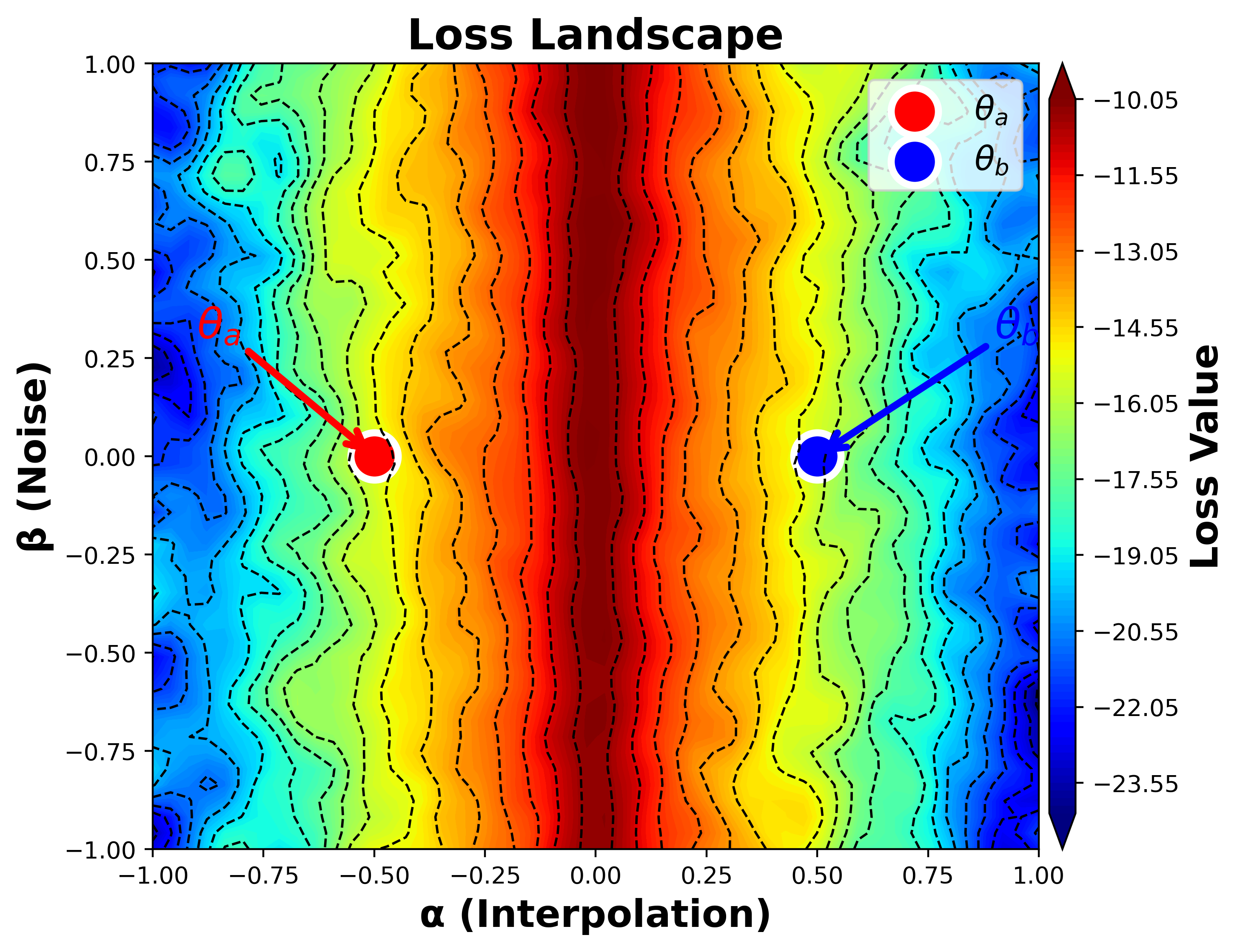}
        \caption{Amazon-Computer}
    \end{subfigure}
    \begin{subfigure}[b]{0.24\textwidth}
        \includegraphics[width=1.0\textwidth]{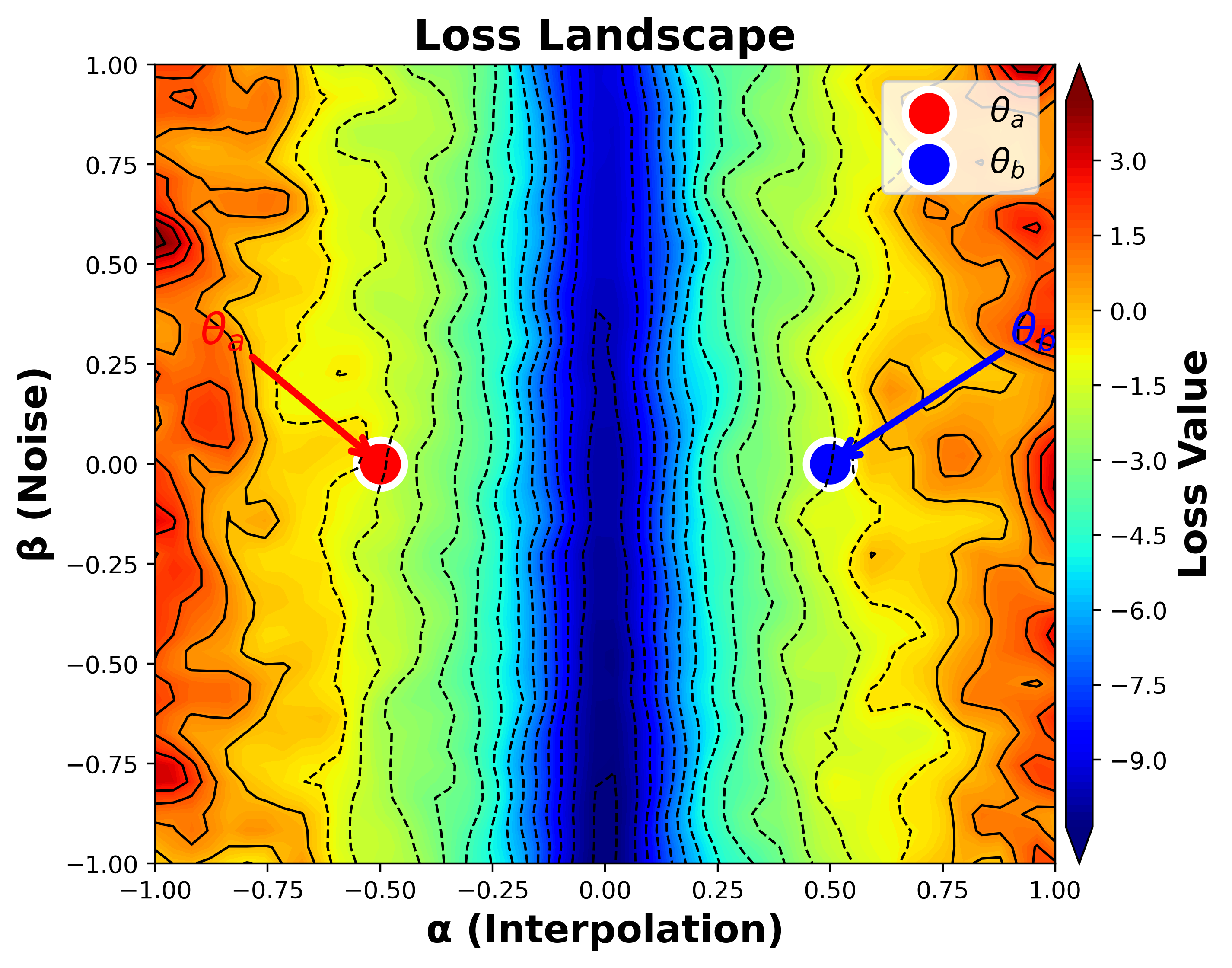}
        \caption{Amazon-Photo}
    \end{subfigure}
    \begin{subfigure}[b]{0.24\textwidth}
        \includegraphics[width=1.0\textwidth]{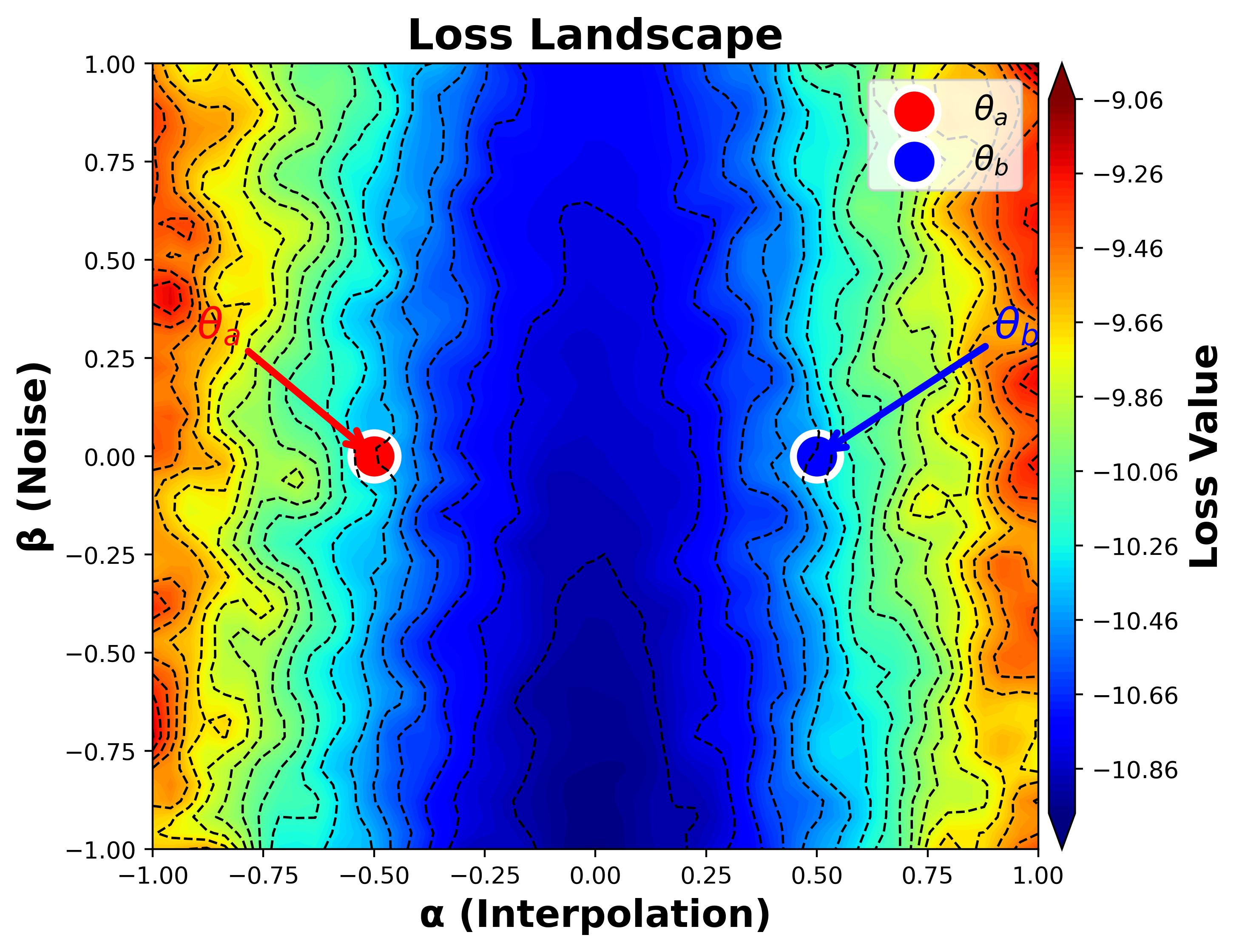}
        \caption{Coauthor-CS}
    \end{subfigure}
    \begin{subfigure}[b]{0.24\textwidth}
        \includegraphics[width=1.0\textwidth]{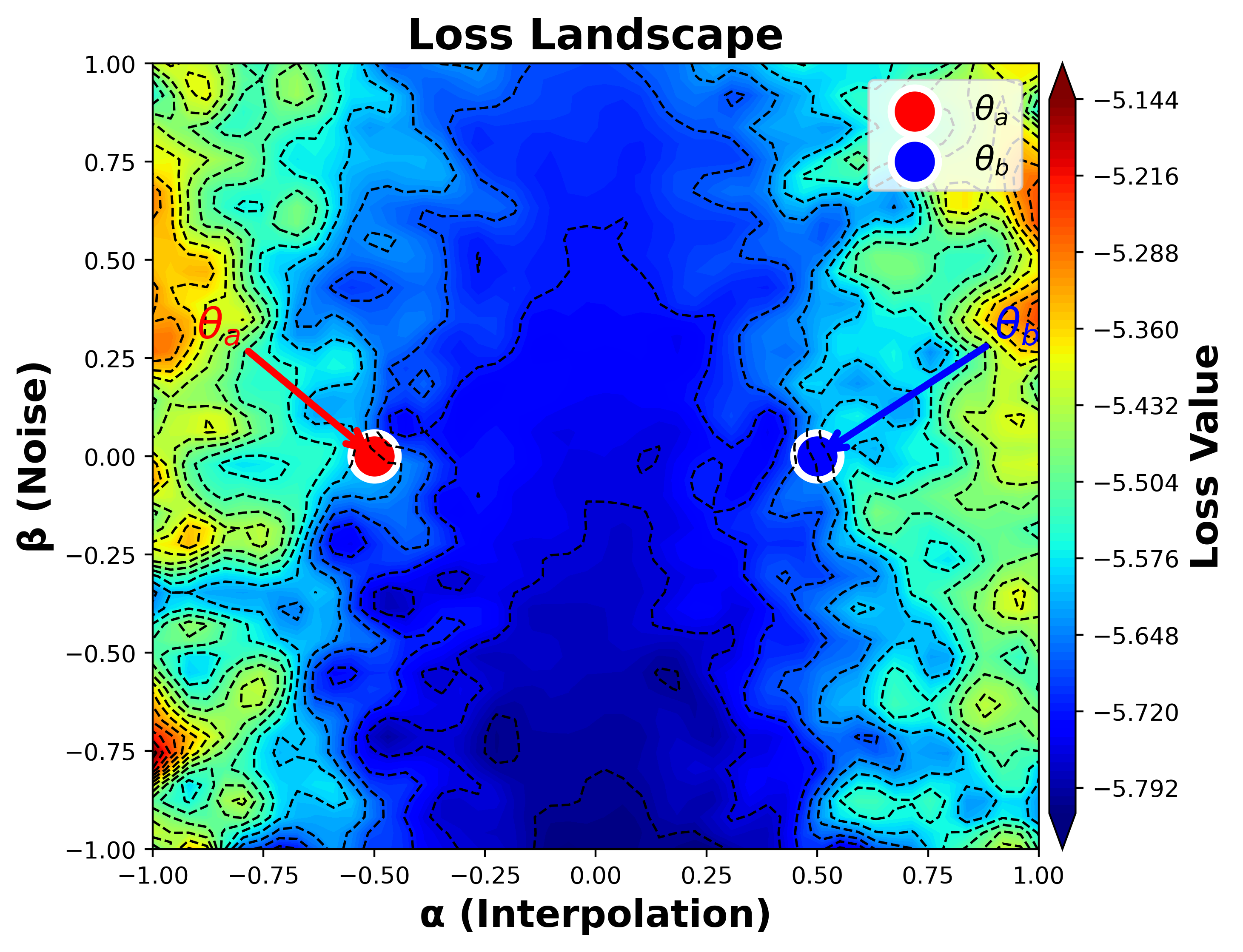}
        \caption{Coauthor-Physics}
    \end{subfigure}
    \begin{subfigure}[b]{0.24\textwidth}
        \includegraphics[width=1.0\textwidth]{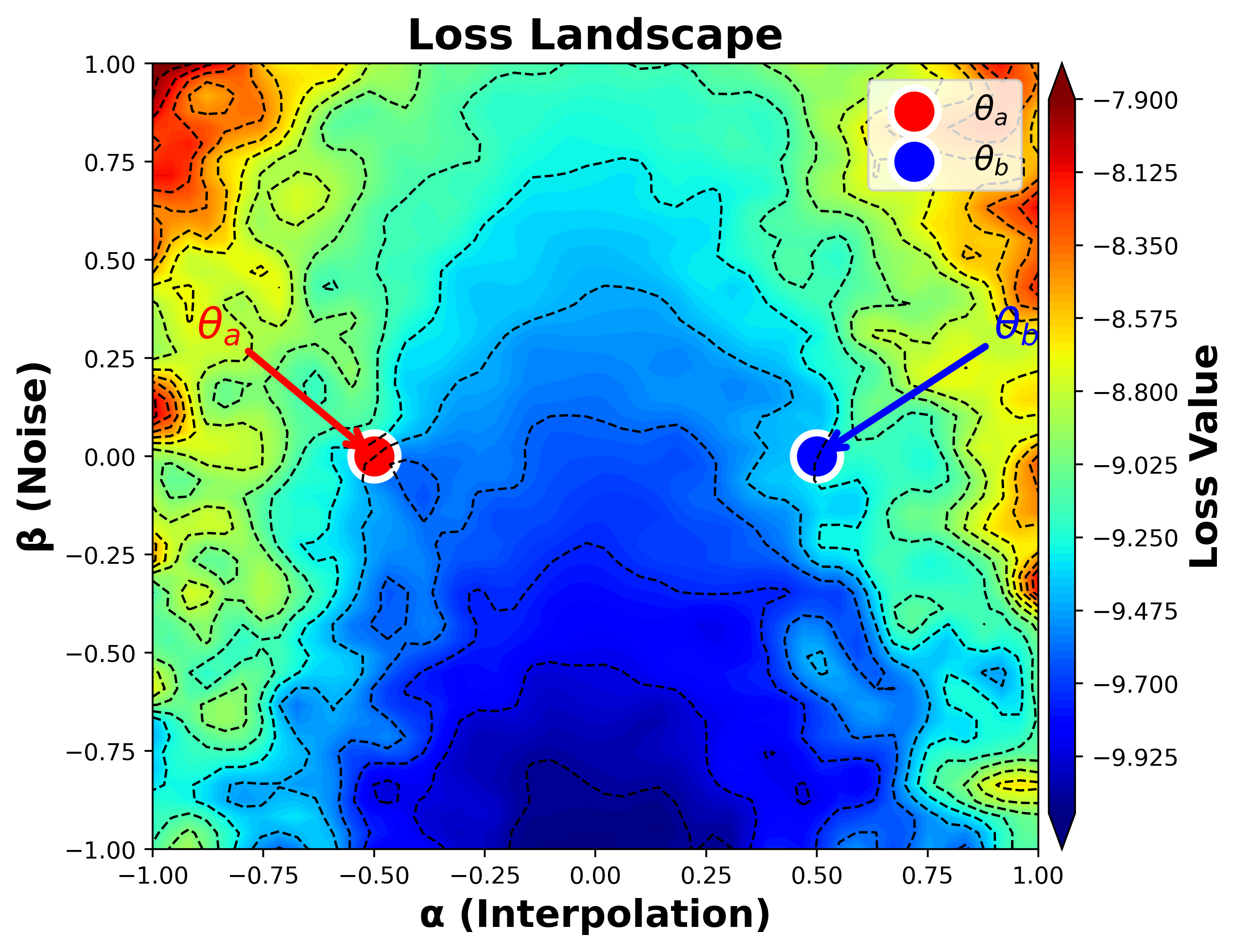}
        \caption{WikiCS}
    \end{subfigure}
    \begin{subfigure}[b]{0.24\textwidth}
        \includegraphics[width=1.0\textwidth]{fig/basin/Basin/loss_landscape_fixed_squirrel.png}
        \caption{Squirrel}
    \end{subfigure}
    \begin{subfigure}[b]{0.24\textwidth}
        \includegraphics[width=1.0\textwidth]{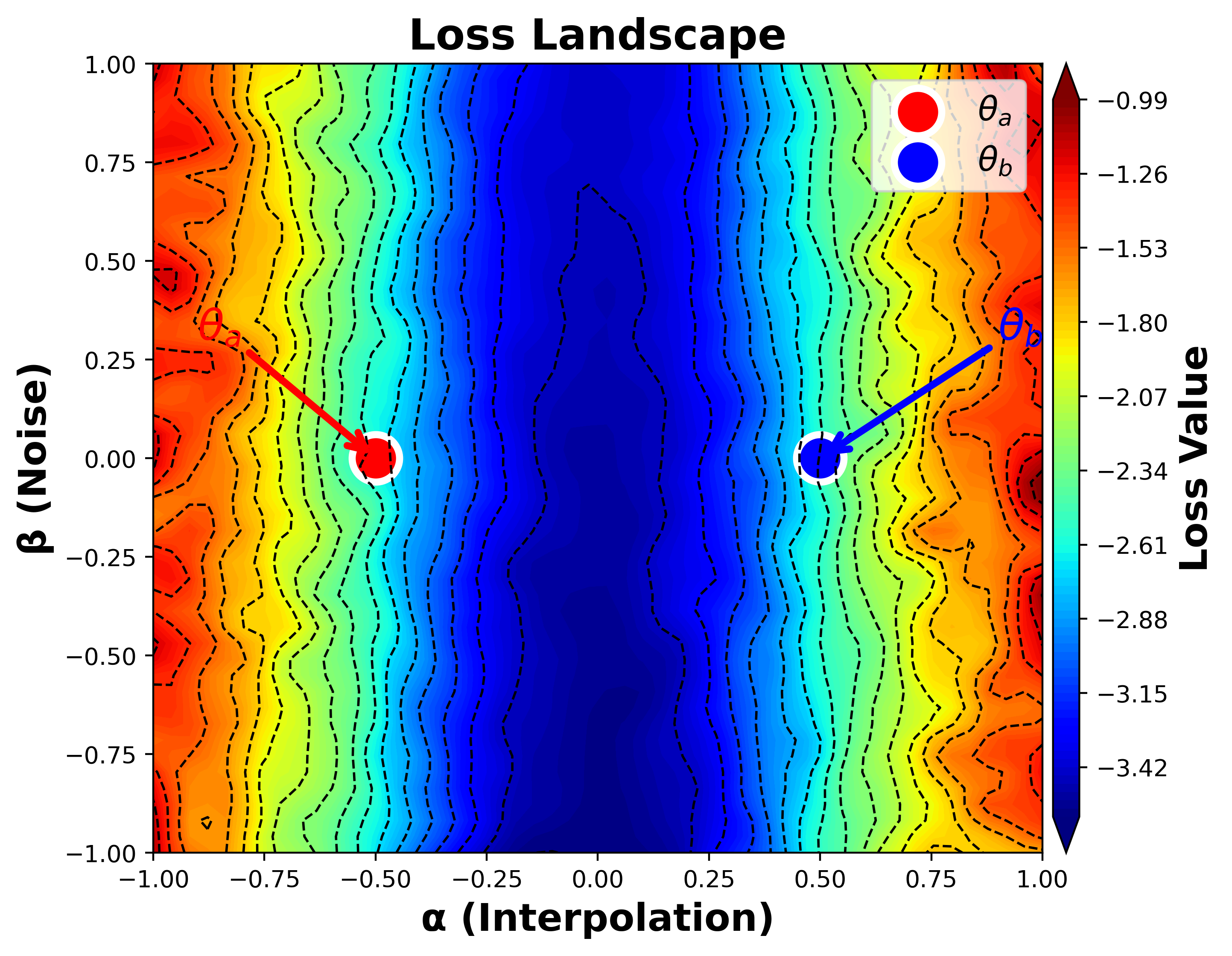}
        \caption{Chameleon}
    \end{subfigure}
    \begin{subfigure}[b]{0.24\textwidth}
        \includegraphics[width=1.0\textwidth]{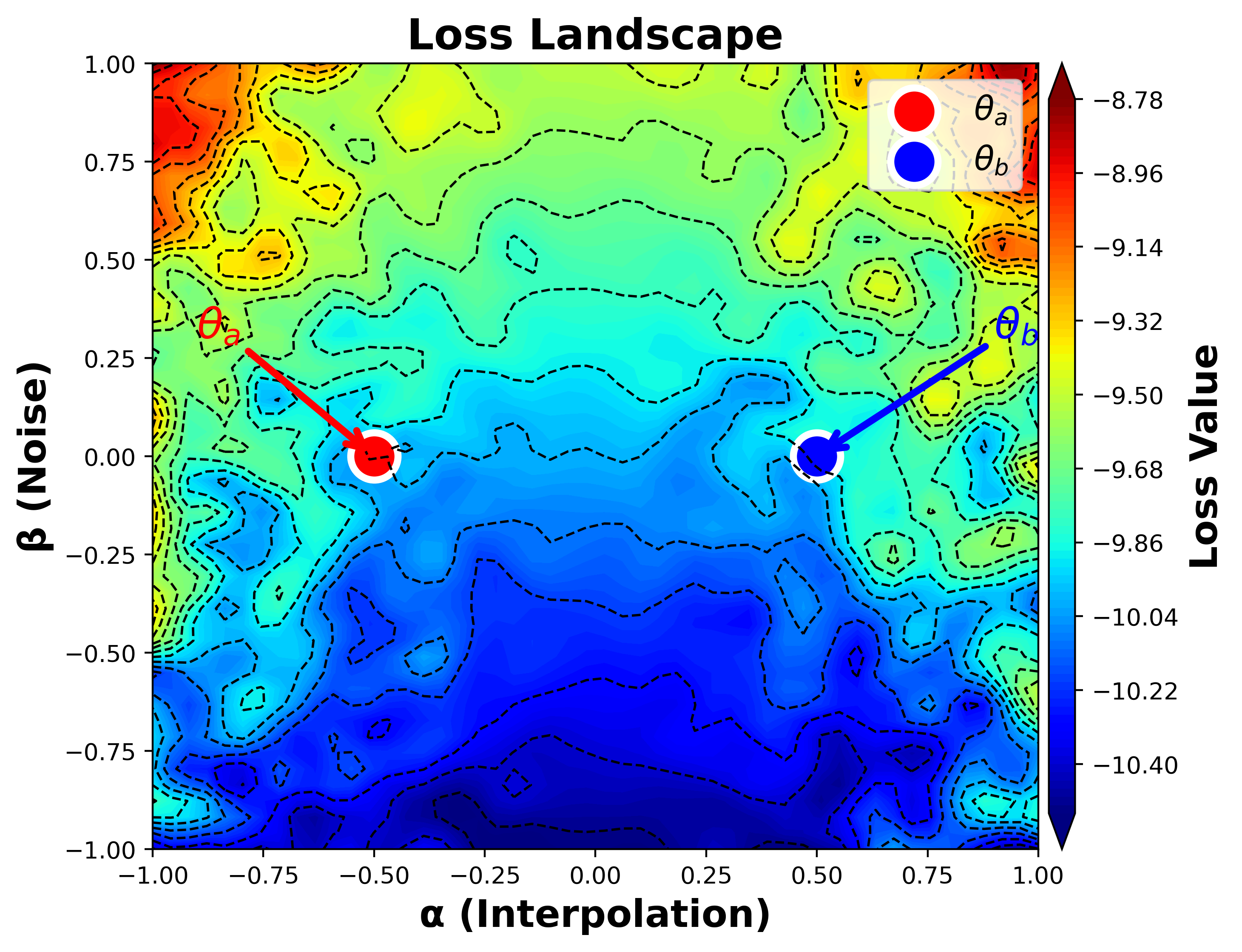}
        \caption{Roman-Empire}
    \end{subfigure}
    \begin{subfigure}[b]{0.24\textwidth}
        \includegraphics[width=1.0\textwidth]{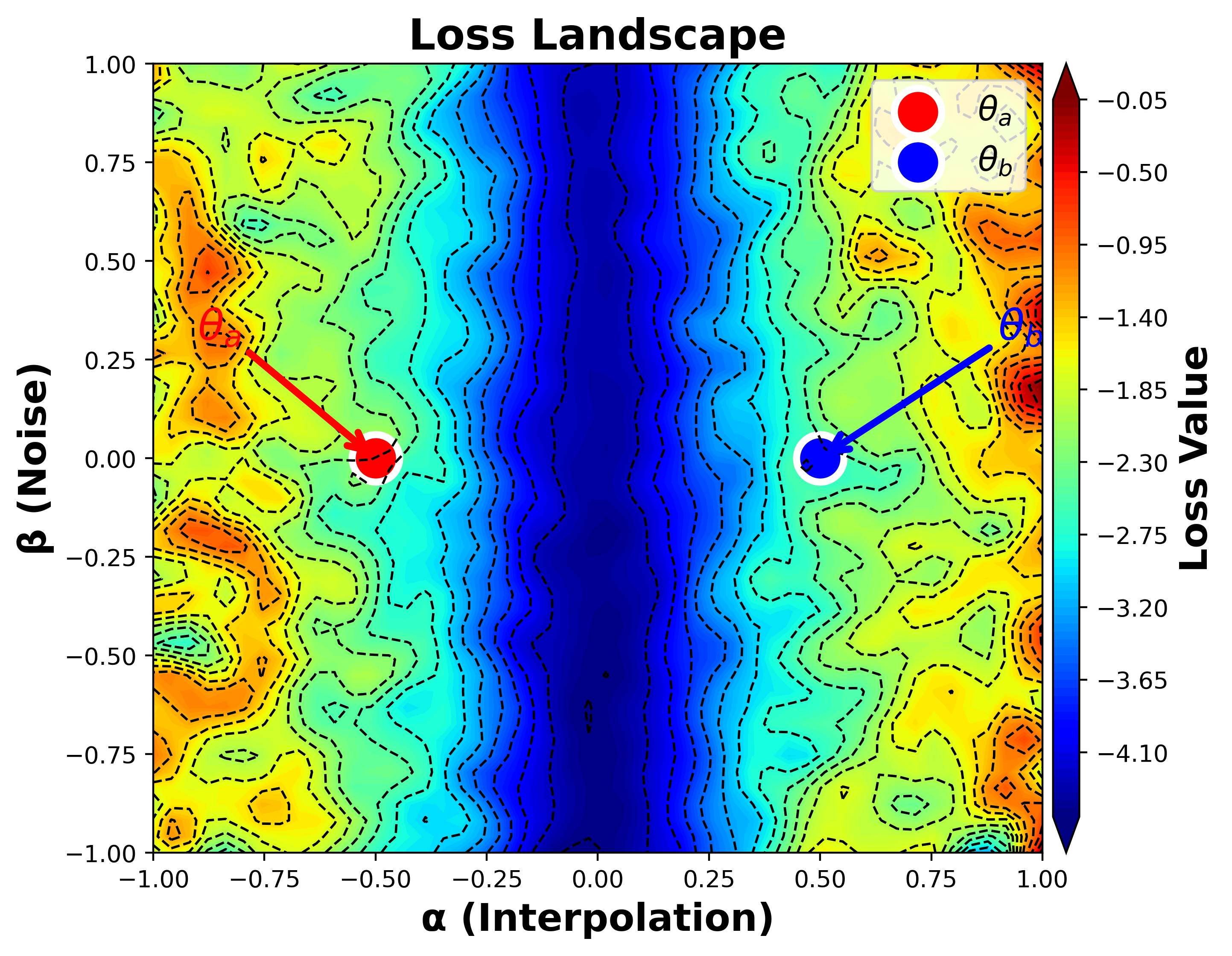}
        \caption{Amazon-Ratings }
    \end{subfigure}
        \begin{subfigure}[b]{0.24\textwidth}
        \includegraphics[width=1.0\textwidth]{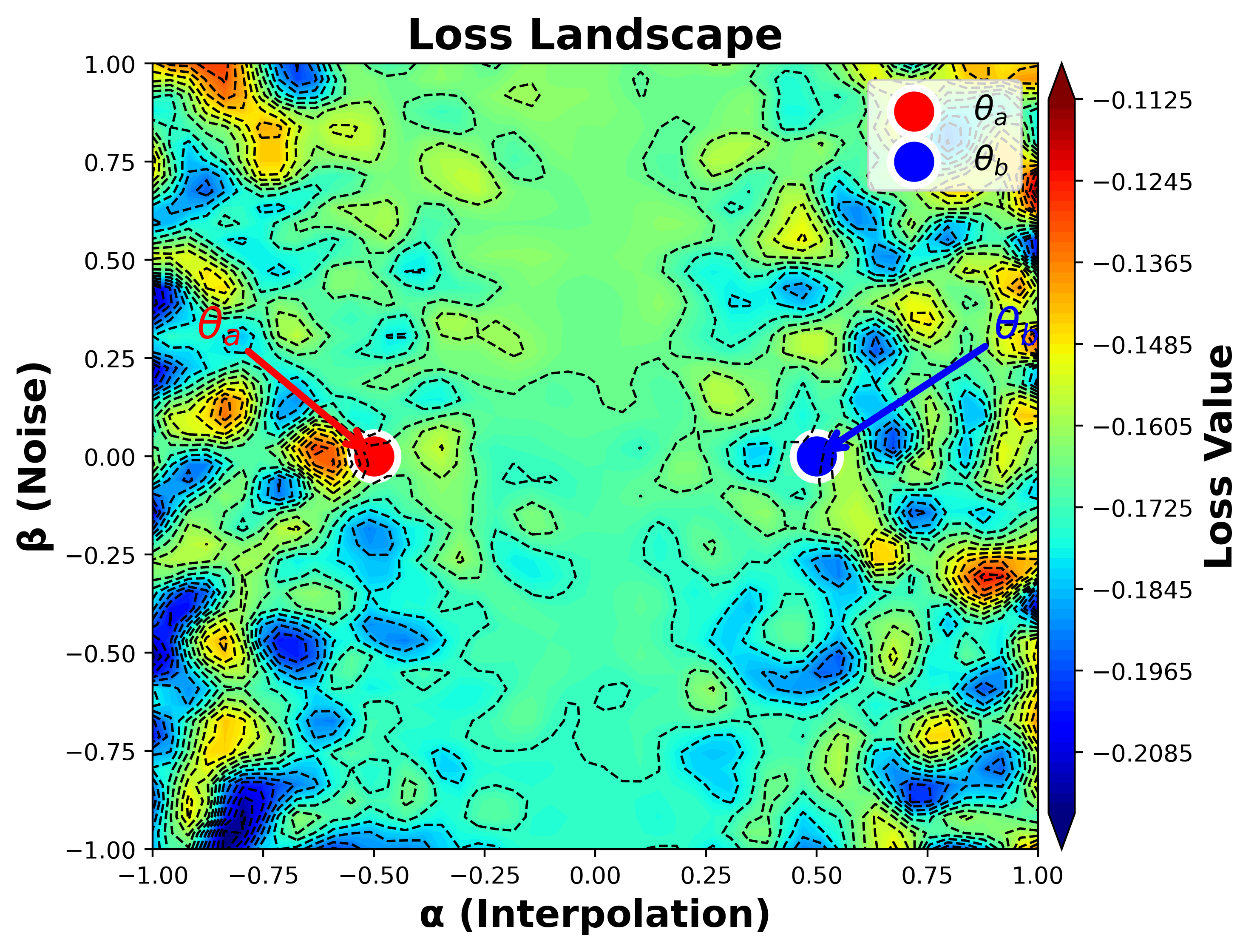}
        \caption{Minesweeper}
    \end{subfigure}
    \caption{Performance of mode connectivity on different convolution mechanisms.}
    \label{fig:A4}
\end{figure*}

\section{Experimental Settings \label{app:implementation}}

We follow the model architecture design and hyperparameter settings of \citet{luo2024classic}. In order to accommodate the computational requirements for our extensive experiments, we harness a variety of high-capacity GPU resources. This includes: Tesla V100 32Gb, NVIDIA RTX A6000
48Gb, NVIDIA RTX A5000 24Gb, and Quadro RTX 8000 48Gb.

\section{Datasets}

In this paper, we adopt $12$ datasets from different domains.

\noindent{\small$\bullet$} \textbf{[Citation network]}. \emph{Cora}, \emph{Citeseer}, and \emph{Pubmed}~\citep{Yang2016RevisitingSL} are citation graphs where each node corresponds to a scientific paper. In these graphs, nodes are characterized by bag-of-words feature vectors, and each one is assigned a label that indicates its research field. It is important to note that all three datasets are examples of homophilous graphs.

\noindent{\small$\bullet$} \textbf{[Amazon network]}. \citet{Shchur2018PitfallsOG} In this network, products are nodes, and an edge signifies that two products are often bought together.  Each product has associated reviews, which are treated as a bag of words.  The task is to determine the product category for each item in the network.

\noindent{\small$\bullet$} \textbf{[Coauthor network]}. \citet{Shchur2018PitfallsOG} The network represents authors connected by co-authorship.  Using the keywords from their published papers, we aim to classify each author according to their research field.

\noindent{\small$\bullet$} \textbf{[Wikics network]}. WikiCS~\citep{mernyei2020wiki} is a hyperlink network in the field of computer science on Wikipedia. The categories correspond to different research directions in computer science, such as artificial intelligence, computer vision, network security, etc.

\noindent{\small$\bullet$} \textbf{[Wikipedia network]}.  Squirrel and Chameleon~\citep{platonov2023a} represent two distinct portions of the Wikipedia web.  The objective is to categorize each individual webpage (node) within these portions into one of five traffic-based classifications, determined by their respective average monthly page views.

\noindent{\small$\bullet$} \textbf{[Heterophilous network]}. These networks are from \citet{platonov2023a} Amazon Ratings, A co-purchasing network of products with reviews used to predict product category. Minesweeper, a synthetic grid-based graph where nodes represent cells, and the task is to identify mines using information about neighboring mines. Roman Empire, a word dependency graph from a Wikipedia article, where nodes are words, and edges represent sequential or syntactic relationships, with the task of classifying words by their syntactic roles.

\end{document}